\PassOptionsToPackage{table}{xcolor}
\pdfoutput=1

\documentclass[11pt]{article}
\usepackage{float}
\usepackage{acl}
\usepackage[utf8]{inputenc}
\usepackage{pgfplots}
\usepackage{dsfont}

\DeclareUnicodeCharacter{2212}{−}
\usepgfplotslibrary{groupplots,dateplot}
\usetikzlibrary{patterns,shapes.arrows}
\pgfplotsset{compat=newest}

\usepackage{tikzscale} 
\usepackage[utf8]{inputenc}
\usepackage{amsmath}
\usepackage{bbm}
\newcommand{\probP}{\text{I\kern-0.15em P}}

\usepackage{colortbl} % Include this in the preamble
\usepackage{makecell}

\usepackage{graphicx}

\usepackage{multirow, colortbl}
\usepackage{color}
\usepackage{array}
\usepackage{multirow}
\usepackage{amssymb}
\usepackage{amsmath}
\usepackage{pifont}
\usepackage{CJKutf8}
\usepackage{tabularx,booktabs}
\usepgfplotslibrary{groupplots}
\usepackage{makecell}
\usepackage{enumitem}
\usepackage{placeins}
\usepackage[normalem]{ulem}
\useunder{\uline}{\ul}{}
\usepackage{graphicx}
\usepackage{booktabs}
\usepackage{pifont}
\usepackage{pgfplots}
\pgfplotsset{width=10cm,compat=1.9}
\usepackage[]{algpseudocode}
\usepackage[]{algorithm}
\algtext*{EndFor}%
\algtext*{EndProcedure}%
\usepackage{booktabs}
\usepackage[normalem]{ulem}
\useunder{\uline}{\ul}{}
\usepackage{times}
\usepackage{latexsym}
\usepackage{adjustbox}

\usepackage[T1]{fontenc}
\usepackage[english]{babel}

\usepackage{amsmath}
\usepackage{amssymb}
\usepackage{booktabs}
\usepackage{multirow}
\usepackage{scalerel,xparse}

\usepackage{cleveref}
\usepackage{microtype}
\usepackage{enumitem}
\usepackage{pgfplots}
\usepackage{dsfont}

\usepackage{multirow}
\usepackage[normalem]{ulem}
\useunder{\uline}{\ul}{}
\usepackage{fontawesome5}
\usepackage{booktabs}
\usepackage{color}
\usepackage{xspace}
\usepackage{todonotes}
\usepackage{enumitem}
\usepackage[most]{tcolorbox}
\usepackage{array}
\usepackage{cleveref}

\usepackage{graphicx}
\usepackage{subfigure}

\definecolor{bggray}{rgb}{0.95, 0.95, 0.95}
\usepackage[%
    framemethod=tikz,
    skipbelow=\topskip,
    skipabove=\topskip
]{mdframed}
\mdfsetup{%
    leftmargin=0pt,
    rightmargin=0pt,
    backgroundcolor=bggray,
    middlelinecolor=black,
    roundcorner=3
}
\newtcolorbox[list inside=prompt,auto counter,number within=section]{prompt}[1][]{
    colbacktitle=black!60,
    fonttitle=\small,
    coltitle=white,
    fontupper=\footnotesize,
    boxsep=4pt,
    left=0pt,
    top=0pt,
    bottom=0pt,
    boxrule=1pt,
    #1,
}

\newtcolorbox[list inside=casestudycon,auto counter,number within=section]{casestudycon}[1][]{
    colbacktitle=red!70,
    fonttitle=\small,
    coltitle=white,
    fontupper=\footnotesize,
    boxsep=4pt,
    left=0pt,
    right=0pt, % Adjust the right margin if needed
    top=0pt,
    bottom=0pt,
    boxrule=1pt,
    width=\textwidth, % Make the box span the text width
    #1,
}

\newtcolorbox[list inside=casestudypro,auto counter,number within=section]{casestudypro}[1][]{
    colbacktitle=green!90,
    fonttitle=\small,
    coltitle=white,
    fontupper=\footnotesize,
    boxsep=4pt,
    left=0pt,
    right=0pt, % Adjust the right margin if needed
    top=0pt,
    bottom=0pt,
    boxrule=1pt,
    width=\textwidth, % Make the box span the text width
    #1,
}

\newtcolorbox[list inside=casestudyconmain,auto counter,number within=section]{casestudyconmain}[1][]{
    colbacktitle=red!70,
    fonttitle=\small,
    coltitle=white,
    fontupper=\footnotesize,
    boxsep=4pt,
    left=0pt,
    right=0pt, % Adjust the right margin if needed
    top=0pt,
    bottom=0pt,
    boxrule=1pt,
    #1,
}

\newtcolorbox[list inside=casestudypromain,auto counter,number within=section]{casestudypromain}[1][]{
    colbacktitle=green!90,
    fonttitle=\small,
    coltitle=white,
    fontupper=\footnotesize,
    boxsep=4pt,
    left=0pt,
    right=0pt, % Adjust the right margin if needed
    top=0pt,
    bottom=0pt,
    boxrule=1pt,
    #1,
}

\newtcolorbox[list inside=casestudymix,auto counter,number within=section]{casestudymix}[1][]{
    colbacktitle=black!90,
    fonttitle=\small,
    coltitle=white,
    fontupper=\footnotesize,
    boxsep=4pt,
    left=0pt,
    right=0pt, % Adjust the right margin if needed
    top=0pt,
    bottom=0pt,
    boxrule=1pt,
    width=\textwidth, % Make the box span the text width
    #1,
}

\DeclareUnicodeCharacter{2212}{−}
\usepgfplotslibrary{groupplots,dateplot}
\usetikzlibrary{patterns,shapes.arrows}
\pgfplotsset{compat=newest}
\definecolor{zoey green}{rgb}{0.684,0.836,0.227}
\newcommand{\mycomment}[3]{}

\newcommand{\ignore}[1]{}

\title{Multiple LLM Agents Debate for Equitable Cultural Alignment}
\author{
Dayeon Ki \quad 
Rachel Rudinger \quad 
Tianyi Zhou \quad 
Marine Carpuat \\
University of Maryland \\
\texttt{\{dayeonki,rudinger,tianyi,marine\}@umd.edu}
}

\begin{document}
\maketitle

\begin{abstract}
% TL;DR for Openreview submission
% We present a Multi-Agent Debate framework to improve equitable cultural alignment in LLMs.

% Research Area/Track for Openreview submission (https://aclrollingreview.org/areas)
% Option 1. Computational Social Science and Cultural Analytics
% keywords: language/cultural bias analysis

% Option 2. Human-Centered NLP
% values and culture -> but this seems quite odd since we don't have any humans involved

% New Abstract
%As large language models (LLMs) advance, their potential to serve multicultural communities underscores the critical need for equitable cultural alignment. 
Large Language Models (LLMs) need to adapt their predictions to diverse cultural contexts to benefit diverse communities across the world. While previous efforts have focused on single-LLM, single-turn approaches, we propose to exploit the complementary strengths of multiple LLMs to promote cultural adaptability. We introduce a Multi-Agent Debate framework, where two LLM-based agents debate over a cultural scenario and collaboratively reach a final decision. We propose two variants: one where either LLM agents exclusively debate and another where they dynamically choose between self-reflection and debate during their turns. We evaluate these approaches on 7 open-weight LLMs (and 21 LLM combinations)  using the \textsc{NormAd-eti} benchmark for social etiquette norms in 75 countries. Experiments show that debate improves both overall accuracy and cultural group parity over single-LLM baselines. Notably, multi-agent debate enables relatively small LLMs (7-9\textsc{B}) to achieve accuracies comparable to that of a much larger model (27\textsc{B} parameters).\footnote{We release our code and dataset at \url{https://github.com/dayeonki/cultural_debate}.}

\end{abstract}

\section{Introduction}

\begin{figure*}
    \centering
    \includegraphics[width=\linewidth]{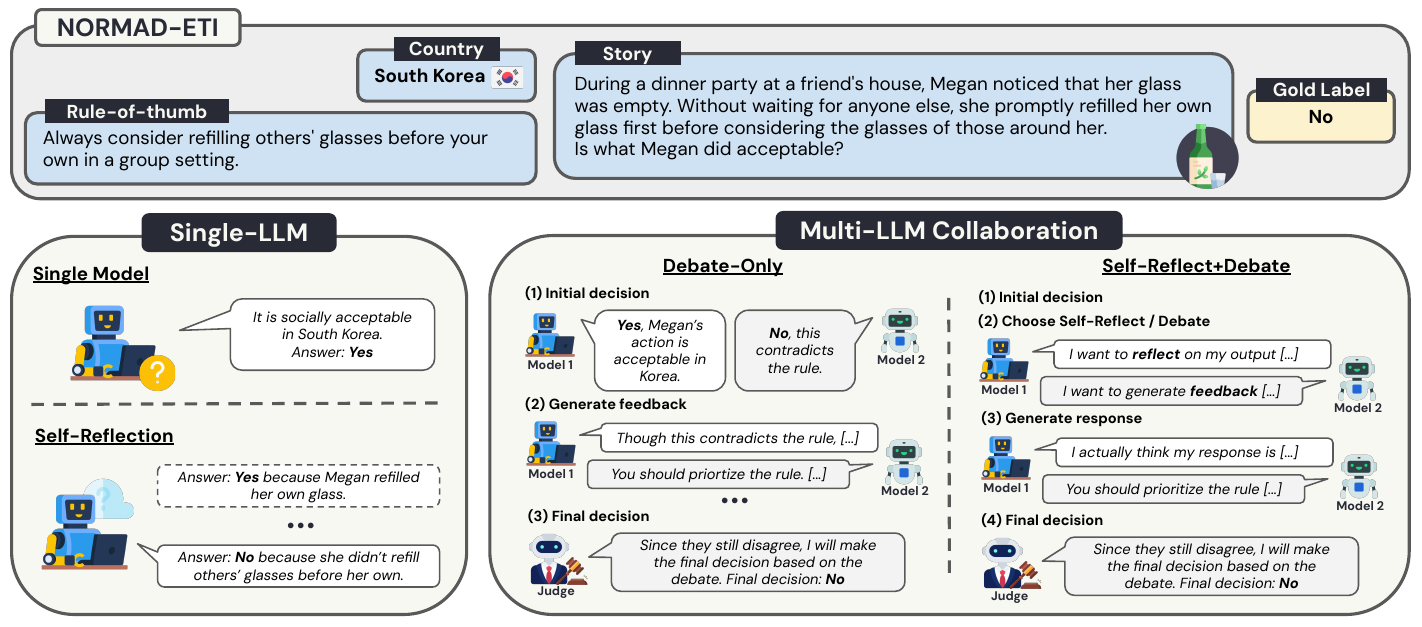}
    \caption{Given a country, rule-of-thumb, and story from the \textsc{NormAd-eti} dataset, each method predicts a ternary label (Yes, No, Neither). \textbf{Single-LLM:} \textbf{1)} Single Model: A single LLM predicts the label, with or without rule-of-thumb information. \textbf{2)} Self-Reflection: LLM reflects on its initial outputs and revises its prediction. \textbf{Multi-LLM Collaboration:} \textbf{3)} Debate-Only: Two LLM-based agents engage in a debate, generating initial decisions, providing feedback to each other, and making final decisions based on the exchanged feedback. If their final decisions differ, a judge LLM resolves the disagreement based on the debate history. \textbf{4)} Self-Reflect+Debate: Each LLM can choose to self-reflect or debate during their turn.}
    \label{fig:main_figure}
\end{figure*}

% New Introduction
%As large language models (LLMs) continue to advance, their potential to serve multicultural and global communities becomes increasingly apparent. This highlights the growing importance of ensuring that LLMs are accessible and effective across diverse cultural contexts to support equitable deployment. 

Large Language Models (LLMs) should be able to adapt their predictions to distinct cultural contexts to benefit diverse communities across the world. Recent efforts in this direction have focused on aligning LLMs to reflect diverse human values and norms across cultures \cite{modular, choenni-etal-2024-echoes, culturellm, selfalignment}. However, these efforts have focused on single-LLM, single-turn generation approaches \cite{cao-etal-2023-assessing, investigating, xu2024selfpluralisingculturealignmentlarge}. Training specialized LLMs \cite{culturellm} and prompt design \cite{shen-etal-2024-understanding} have been shown to improve overall cultural alignment. Yet it can be expected that data selection and model design decisions make it challenging for any single model to support all cultures equally well. 

Meanwhile, a wealth of open-weight LLMs have been released, with varying training data distributions, alignment processes, and language coverage \cite{yu2023kola, du2023improving, bansal2024llm}, thus offering the potential for complementary  perspectives and reasoning paths~\cite{farperspective, liang-etal-2024-encouraging}. Interactions between such models may lead to more accurate understanding and evaluation of culturally sensitive scenarios, which motivates us to explore multi-LLM collaboration-based approaches. 

How can multiple LLMs collaborate toward equitable alignment across cultures? We investigate a common form of multi-LLM collaboration: debate \cite{irving2018ai, khan2024debatingpersuasivellmsleads, kenton2024scalableoversightweakllms}. We propose a \textbf{Multi-Agent Debate} framework, where two LLM agents debate over the given scenario and collaboratively arrive at a final decision with a judge LLM. We introduce two key variants as illustrated in Figure \ref{fig:main_figure}: \textbf{1) Debate-Only}, where multiple LLM agents exclusively engage in debate with a discussant, and \textbf{2) Self-Reflect+Debate}, where each LLM agent dynamically choose between self-reflection and debating during its turn. To conduct a comprehensive comparison study, we investigate two additional strategies based on single-LLM in Figure \ref{fig:main_figure}: 3) Single Model, where a single LLM generates outputs, and 4) Self-Reflection, where an LLM generates verbal self-reflections on its own outputs and incorporate them in subsequent iterations.

Using the \textsc{NormAd-eti} benchmark \cite{normad}, which probes social-etiquette related social norms across 75 countries, we evaluate the four above strategies across 7 LLMs (and 21 LLM combinations). We begin by improving single-LLM performance through cultural contextualization and self-reflection (\S \ref{sec:single_model}), and show that the best-performing LLM often varies across cultures (\S \ref{sec:oracle}), which motivates the multi-agent debate setup (\S \ref{sec:debate_only}, \S \ref{sec:selfreflect_debate}). As we will see, using multi-turn multi-agent approaches improve over single-turn single-LLM accuracies.  We also analyze how predictions evolve throughout the debate stages: while the  LLM agents often disagree initially (\S \ref{sec:decision_dynamics}), debate effectively guides the agents to revise these decisions, with most revisions resulting in correct final outcomes. Moreover, we demonstrate that improvements are consistent across cultural groups, with the multi-agent debate setup achieving the highest parity in performance, particularly benefiting underrepresented cultural groups (\S \ref{sec:cultural_group_parity}).

\section{Background and Related Work}

\paragraph{LLM Cultural Alignment.}
The accuracy of LLM predictions across diverse cultural contexts has received increased attention. One line of work examines the cross-cultural difference in moral values and beliefs encoded in LLMs and their alignment with diverse human perspectives \cite{moralstories, moralencoded, jinnai-2024-cross, cao-etal-2023-assessing, arora-etal-2023-probing, trollymultilingual}. Another explores how LLMs interpret social etiquette and norms across varying cultural contexts \cite{farperspective, normdial, renovi, culturebank, normad, normbank}. Building on these works, our study focuses specifically on the cultural alignment of LLMs with social norms, using \textsc{NormAd-eti} \cite{normad} benchmark as a testbed.

% Previous works have primarily evaluated cultural alignment by comparing LM responses to sociological survey data from diverse populations, such as Hofstede’s cultural dimensions \cite{hofstede1983review, masoud2024culturalalignmentlargelanguage, kharchenko2024llmsrepresentvaluescultures} and the World Values Survey (WVS) \cite{haerpfer2021world, selfalignment, arora-etal-2023-probing, investigating, wang-etal-2024-countries}. Alternatively, LLMs have been probed using cloze-style or survey-based questions in QA setups \cite{culturegen, choenni-etal-2024-echoes, culturalbench}. 

% \mc{weird to introduce acronym here if the section is at the end. Maybe it should be background instead? also why not just use LLMs?}

% \paragraph{Enhancing Cultural Alignment.}
Extensive efforts have also been made to enhance cultural alignment in LLMs, encompassing advancements in pre-training \cite{acegpt, seallm, wang-etal-2024-countries}, alignment training \cite{choenni-etal-2024-echoes, culturellm, mukherjee-etal-2024-global}, and inference strategies, such as effective prompt design \cite{investigating, cao-etal-2023-assessing, shen-etal-2024-understanding} and in-context learning \cite{selfalignment, lahoti-etal-2023-improving}. However, prior work has primarily focused on single LLMs in single-turn interactions, with the use of multiple LLMs under-explored. The closest work to ours is CulturePark \cite{culturepark}, which leverages an LLM-powered multi-agent communication framework for cultural data collection. We depart from this work by adapting a multi-agent debate framework specifically to enhance cultural alignment, presenting a new perspective within model inference strategies.

\paragraph{Multi-Agent Debate.}
The diverse perspectives of LLMs arise from differences in training data and alignment processes, making them well-suited for pluralistic alignment tasks \cite{modular}. Inspired by human reasoning, prior works have proposed frameworks where multiple LLMs generate responses, process others' opinions, and engage in debates to reach consensus \cite{estornell2024multillm}. This collaborative approach has shown potential in problem-solving by combining knowledge and extracting diverse perspectives \cite{farperspective}. Most works focus on improving factuality and reasoning capabilities, where they show that debate reduce the tendency of LLMs to hallucinate incorrect facts through ``\textit{society of minds}'' interaction \cite{du2023improvingfactualityreasoninglanguage} or encourage divergent thinking \cite{liang-etal-2024-encouraging} in arithmetic reasoning, question-answering, and translation tasks. Other studies used multi-agent debate frameworks to evaluate LLM-generated responses \cite{chateval} or to represent different nationalities for constructing culturally specific debate datasets \cite{culturepark}. However, no prior work has specifically employed a multi-agent debate framework to enhance cultural alignment in LLMs, and our work aims to fill this gap.

% \mc{I think this section would work well as background/motivation just after intro.}
% \mc{Can you name the tasks or kinds of tasks that have been tackled in prior work? I think it would help to make the difference with this work more concrete.} 
\section{Steering LLMs for Cultural Alignment}
\label{sec:methods}

We present an overview of our tested strategies for improving cultural alignment in Figure \ref{fig:main_figure}. We first adapt existing approaches as \textbf{1) Single-LLM} (\S \ref{method:single_llm}) and propose \textbf{2) Multi-LLM Collaboration}-based approaches (\S \ref{method:multi_llm}). All prompts are in Appendix \ref{appendix:prompts}.

\subsection{Single-LLM}
\label{method:single_llm}
% \paragraph{Single Model.}
% We evaluate LLMs with two levels of cultural contextualization: with and without the relevant social norm to the story (\textit{rule-of-thumb}) in the prompt.

\paragraph{Self-Reflection.}
Previous studies have shown that LLMs can evaluate their outputs and learn from their own feedback \cite{kadavath2022language, reflexion, pan2023automatically, renze2024selfreflectionllmagentseffects}. Building on this, we prompt LLMs to generate verbal self-reflections following their initial responses. Specifically, given the country, rule-of-thumb, and story as input, we prompt an LLM $\mathcal{M}$ to 1) generate an initial decision $\hat{y}_0^\mathcal{M}$ by evaluating the cultural adaptability of the given scenario, 2) generate a rationale for their decision $f^\mathcal{M}$, and 3) incorporate this as additional context to make a final decision $\hat{y}_f^\mathcal{M}$.

% \mc{Use notations consistent with those used for debate throughout to formalize the process.}

% \subsection{Selective Model}
% \label{method:selective model}
% Stories from different cultures may not benefit uniformly from a single model, as the best-performing model often varies across cultures. To explore the potential of \textbf{multi-LLM collaboration}, we theoretically evaluate an approach that combines the strengths of multiple LLMs by routing queries based on ground truth labels. Specifically, given two LLMs, $\mathcal{M}_1$ and $\mathcal{M}_2$, predictions for each culture are assigned to the LLM with higher mean accuracy. For instance, if $\mathrm{Acc}(\mathcal{M}_1) > \mathrm{Acc}(\mathcal{M}_2)$ for entries related to South Korean culture $\mathcal{C}_{sk}$, then $\mathcal{M}_1$ handles all predictions for $\mathcal{C}_{sk}$. While this approach is not directly comparable to other methods, as it relies on access to gold labels, it provides a theoretical upper bound and motivates for other forms of multi-LLM collaboration strategies (\S \ref{method:multi-agent debate}).

\subsection{Multi-LLM Collaboration}
\label{method:multi_llm}
On one hand, adapting a single LLM to reflect on its own generations poses challenges such as confirmation bias and hallucination, which can lead to unreliable self-evaluation \cite{feng-etal-2024-dont, liang-etal-2024-encouraging}. On the other hand, LLMs often exhibit varying knowledge coverage, with the potential to complement each other due to differences in training data distributions and alignment processes \cite{yu2023kola, du2023improving, bansal2024llm}. We tap into this \textit{knowledge complementarity} through one common form of multi-LLM collaboration, debate \cite{irving2018ai}, where two LLM-based agents debate and collaboratively evaluate the given scenario.

\paragraph{Debate-Only.}
Given the same inputs as in \S \ref{method:single_llm}, two LLM agents, $\mathcal{M}_1$ and $\mathcal{M}_2$, independently generate their initial decisions, $\hat{y}_0^{\mathcal{M}_1}$ and $\hat{y}_0^{\mathcal{M}_2}$ respectively. Each agent then provides feedback, $f^{\mathcal{M}_1}$ and $f^{\mathcal{M}_2}$ to the discussant's initial decision. We exchange the feedback, and each agent incorporates the feedback along with their initial decisions to arrive at their final decisions, $\hat{y}_f^{\mathcal{M}_1}$ and $\hat{y}_f^{\mathcal{M}_2}$. For $i=1$ and $j=2$ (and vice versa):

\begin{equation}
\begin{split}
    \hat{y}_f^{\mathcal{M}_i} &= \mathcal{M}_i(\hat{y}_0^{\mathcal{M}_i}, \hat{y}_0^{\mathcal{M}_j}, f^{\mathcal{M}_i}, f^{\mathcal{M}_j}) \\
\end{split}
\end{equation}
If their final decisions are identical, we use it as the aggregated decision, regardless of its correctness. Otherwise, we employ a judge LLM, $\mathcal{M}_j$, to summarize and synthesize the debate history into a final decision \cite{feng-etal-2024-dont, li-etal-2024-llms-speak}.

%     \hat{y}_f^{\mathcal{M}_2} &= \mathcal{M}_2(\hat{y}_0^{\mathcal{M}_1}, \hat{y}_0^{\mathcal{M}_2}, f^{\mathcal{M}_1}, f^{\mathcal{M}_2})

\begin{equation}
\begin{split}
    \mathcal{M}_j(\hat{y}_0^{\mathcal{M}_1}, \hat{y}_0^{\mathcal{M}_2}, f^{\mathcal{M}_1}, f^{\mathcal{M}_2},
    \hat{y}_f^{\mathcal{M}_1}, \hat{y}_f^{\mathcal{M}_2}) \rightarrow \hat{y}_f^{\mathcal{M}_j}
\end{split}
\end{equation}

\paragraph{Self-Reflect+Debate.}
LLMs may exhibit varying preferences when responding to feedback. To account for this, we extend the Debate-Only setup by agents making choices during their turn: (A) self-reflection, where they generate reflection $r^{\mathcal{M}_i}$ on their previous output, or (B) debate, where they generate feedback $f^{\mathcal{M}_i}$ in response to the discussant's opinion. Formally, if $\mathcal{M}_1$ chooses (A) self-reflection and $\mathcal{M}_2$ chooses (B) debate, the updated decision for $i=1$ and $j=2$ (and vice versa) is:

\begin{equation}
\begin{split}
    \hat{y}_f^{\mathcal{M}_i} &= \mathcal{M}_i(\hat{y}_0^{\mathcal{M}_i}, \hat{y}_0^{\mathcal{M}_j}, r^{\mathcal{M}_1}, f^{\mathcal{M}_2}) \\
\end{split}
\end{equation}
We use the judge LLM if their final decisions differ. Note that in both multi-LLM setups, $\mathcal{M}_1$ and $\mathcal{M}_2$ are \textit{exchangeable}, as each agent has symmetrical access to the same information.

% We provide a broader action space to enable LLMs make more informed decisions by leveraging both their internal parametric knowledge and the external information presented in the prompt.\zk{Any citations?}

% These options are presented in a multiple-choice format. To mitigate position bias \cite{orderofoptions} and address potential inconsistencies in LLM choices, the process is repeated twice, alternating the order of the options.
\section{Experiment setup}

\subsection{Dataset}
We use \textsc{NormAd-eti} \cite{normad} for evaluation, a benchmark designed to assess the cultural adaptability of LLMs. The dataset contains 2.6K stories reflecting social and cultural norms from 75 countries, derived from the social-etiquette norms outlined in the Cultural Atlas.\footnote{\url{https://culturalatlas.sbs.com.au/}} Each story is associated with a country, a rule-of-thumb, and a ternary ground truth label $y \in$ \{Yes, No, Neither\} as shown in Figure \ref{fig:main_figure} (\textit{top}). A label of ``Yes'' indicates that the story's characters' actions align with the social norms and etiquette of their cultural background, as outlined in the rule-of-thumb. ``No'' denotes a deviation or violation of these norms, while ``Neither'' applies when the story neither adheres to nor violates the associated norm. Detailed statistics for each cultural bin categorized according to the Inglehart-Welzel cultural map are provided in Appendix Table \ref{tab:normad_stats}.

\subsection{Models}
We use the 7-9B variants of seven open-weight LLMs with varying levels of multilinguality. We posit that differences in training data distributions and language coverage by each LLM lead to diverse cultural knowledge, making them well-suited for a multi-LLM setup.\footnote{We use instruction fine-tuned LLMs, building on the findings of \textsc{NormAd-eti} \citep{normad}, which demonstrated that post-aligned models outperform their SFT couterparts in accuracy. We provide a detailed analysis in Appendix \ref{appendix:pre_post_alignment}.}

% \footnote{The original \textsc{NormAd-eti} paper \cite{normad} only considers English-centric LLMs.}.

\begin{itemize}[leftmargin=*, itemsep=2pt, parsep=-1pt]
 \item \textbf{English-centric:} \textsc{LLaMA-3} \cite{grattafiori2024llama3herdmodels}, \textsc{Gemma-2} \cite{gemma2}
 \item \textbf{Bilingual:} \textsc{EXAONE-3} \cite{exaone} (\textsc{en-ko}), \textsc{Yi-1.5} \cite{yi} (\textsc{en-zh}), \textsc{InternLM-2.5} \cite{internlm} (\textsc{en-zh})
 \item \textbf{Multilingual:} \textsc{Aya-23} \cite{aya23} (23 languages), \textsc{SeaLLM-3} \cite{seallm} (12 East Asian languages)
\end{itemize}
For multi-agent debate, we select \textsc{Gemma-2 27b} as our judge LLM accounting for its high single model accuracy compared to other LLMs and reasonable inference speed as detailed in Appendix \ref{appendix:judge_llm}. We set the default sampling temperature to 0.0, and employ 0.8 where multiple runs are required.\footnote{The HuggingFace model names are detailed in Appendix Table \ref{tab:huggingface_api}.}

\subsection{Evaluation Metrics}
\label{sec:evaluation}
Our primary evaluation metric is accuracy, calculated by comparing LLM responses to the ground truth labels.

% Given an LLM $\mathcal{M}$ and a dataset $\mathcal{D}$, we compare the ground truth label $y_i$ to the LLM response $\hat{y}^\mathcal{M}_i$ for accuracy scores. 

% \begin{equation}
%     \mathbf{Acc}(\mathcal{M}) = \frac{1}{|\mathcal{D}|}\sum_{\mathcal{D}} \mathbbm{1} [y_i = \hat{y}_i^\mathcal{M}]
% \end{equation}

% \input{table/decision_dynamics}

\definecolor{darkpurple}{RGB}{245, 66, 155}
\definecolor{bblue}{RGB}{66, 141, 245}
\definecolor{darkgreen}{RGB}{107, 201, 60}

\paragraph{Decision dynamics.}
For self-reflection and multi-agent debate, we analyze the dynamics of initial and final decisions made by LLMs. We aim to capture two key phases in this process: \textbf{\textcolor{darkpurple}{1) Initial Correctness}}: whether the LLM's initial decision is correct and \textbf{\textcolor{darkgreen}{2) Final Correctness}}: whether the LLM's final decision is correct. Specifically, for the multi-agent debate, we extend to capture additional key phase: \textbf{\textcolor{bblue}{3) Judge Correctness}}: whether the judge LLM's final decision is correct. If both agents already agree from \textbf{\textcolor{darkgreen}{2}}, this is determined by the correctness of the agreed-upon answer.
% We also consider both LLMs simultaneously.
% \textbf{\textcolor{bblue}{(2) Behavioral dynamic}:} whether the LLM changes its initial decision; and 
% We show the variations of all tested dynamics in Table \ref{tab:decision_dynamics}.

\paragraph{Cultural group parity.}
An essential aspect of cultural alignment is minimizing allocational harm and ensuring equitable performance across groups with diverse demographic attributes \cite{ramesh-etal-2023-fairness}. To operationalize this, we propose the concept of cultural group parity, which systematically evaluates how fairly our methods align with norms across different cultural groups \cite{wang-etal-2022-assessing}. Parity is defined as the state where methods exhibit comparable mean accuracies across cultural groups. Following \citet{petrov2023languagemodeltokenizersintroduce}, we set the baseline group $b$ as the group with the highest mean accuracy. For each cultural group $g$, we compute a parity premium relative to $b$. Formally:

\begin{equation}
    \mathbf{Parity}(g) = \frac{\mathrm{Acc}_g}{\mathrm{Acc}_b}
\end{equation}
A method achieves parity for $g$ with respect to $b$ if $\mathbf{Parity}(g)$ $\approx$ 1.

\section{Results}
\label{sec:results}

% \mc{Result tables: you can write results as e.g., 35.8 rather than 0.358}

We begin by maximizing single-LLM performance (\S \ref{sec:single_model}). Next, we assess the potential of multi-LLM collaboration through oracle model selection (\S \ref{sec:oracle}), before exploring the benefits of debate alone (\S \ref{sec:debate_only}) and in combination with self-reflection (\S \ref{sec:selfreflect_debate}).

\definecolor{lightgray}{gray}{0.9}
\definecolor{highgreen}{rgb}{0.1608, 0.5804, 0.2196}
\definecolor{lowred}{rgb}{0.9216, 0.1569, 0.1569}

\begin{table}[!htp]
\centering
\resizebox{\linewidth}{!}{%
    \begin{tabular}{l l l l l l}
    \specialrule{1.3pt}{0pt}{0pt}
    \textbf{Label/Model} & \textbf{Si (w/o)} & \textbf{Si (w/)} & \textbf{Self-Reflect} \\
    \toprule

    \textsc{Yes-Only} & 35.8 & - & - \\
    \textsc{No-Only} & 33.2 & - & - \\
    \textsc{Neither-Only} & 31.0 & - & - \\
    \midrule

    \textsc{LLaMA-3} & 49.5 & 63.7 \scriptsize{\textcolor{highgreen}{(+28.7)}} & \textbf{65.7} \scriptsize{\textcolor{highgreen}{(+3.14)}} \\
    
    \textsc{Gemma-2} & 50.7 & 68.9 \scriptsize{\textcolor{highgreen}{(+35.9)}} & \textbf{72.5} \scriptsize{\textcolor{highgreen}{(+5.22)}} \\
    % \midrule

    \textsc{EXAONE-3} & 42.8 & 63.5 \scriptsize{\textcolor{highgreen}{(+48.4)}} & \textbf{64.3} \scriptsize{\textcolor{highgreen}{(+1.26)}} \\
    
    \textsc{Yi-1.5} & 51.0 & 70.7 \scriptsize{\textcolor{highgreen}{(+38.6)}} & \textbf{71.5} \scriptsize{\textcolor{highgreen}{(+1.13)}} \\
    
    \textsc{InternLM-2.5} & 47.0 & 67.8 \scriptsize{\textcolor{highgreen}{(+44.3)}} & \textbf{70.7} \scriptsize{\textcolor{highgreen}{(+4.28)}} \\ 
    % \midrule

    \textsc{Aya-23} & 49.4 & 65.8 \scriptsize{\textcolor{highgreen}{(+33.2)}} & \textbf{68.1} \scriptsize{\textcolor{highgreen}{(+3.50)}} \\
    
    \textsc{SeaLLM-3} & 49.5 & 68.1 \scriptsize{\textcolor{highgreen}{(+37.6)}} & \textbf{69.3} \scriptsize{\textcolor{highgreen}{(+1.76)}} \\

    \rowcolor{lightgray} \textit{\textsc{Gemma-2-27b}} & 55.8 & 79.2 \scriptsize{\textcolor{highgreen}{(+41.9)}} & \textbf{80.1} \scriptsize{\textcolor{highgreen}{(+1.14)}} \\

    \specialrule{1.3pt}{0pt}{0pt}
    \end{tabular}
}
\caption{Mean accuracies (\%) for Single Model and Self-Reflection baselines. \textbf{Si (w/o):} Single Model without rule-of-thumb information; \textbf{Si (w/):} Single Model with rule-of-thumb information; \textbf{Self-Reflect:} Self-Reflection. Best scores for each row are in \textbf{bold}. \textcolor{highgreen}{Green} text is the \% of improvements compared to the adjacent left column. We include accuracies for \fcolorbox{white}{lightgray}{\textsc{Gemma-2-27b}} judge LLM as reference. Fine-grained results for each country are detailed in Appendix \ref{appendix:detailed_single_llm}.}
\label{tab:single_llm_new}
\end{table}

\subsection{Self-Reflection Improves Single-LLM Accuracy}
\label{sec:single_model}

We empirically investigate the impact of contextualization and multi-turn interactions on a single LLM. For cultural contextualization, we adopt the most effective method from \textsc{NormAd-eti} \cite{normad}: including a social norm relevant to the story (\textit{rule-of-thumb}) in the prompt. As shown in Table \ref{tab:single_llm_new} (\textbf{Si w/}), this method consistently improves mean accuracies by up to 48.4\% and 39.1\% on average across all tested LLMs. This confirms the previous findings that adding relevant cultural context can significantly enhance cultural alignment of LLMs \cite{zhu-etal-2024-large, normad}. Accordingly, we include the rule-of-thumb information in the prompts for all subsequent experiments.

Next, we find that single LLMs can generate feedback on their own outputs and reflect upon it, resulting in an average accuracy improvement of 3.26\% across all tested LLMs through self-reflection. In Table \ref{tab:single_llm_new}, the rankings of \textbf{Si (w/o)} and \textbf{Si (w/)} are strongly positively correlated ($r$=0.85), as well as \textbf{Si (w/)} and \textbf{Self-Reflect} ($r$=0.95), which indicates that stronger base LLMs generally benefit more from both cultural contextualization and self-reflection. 
% \zk{Add citation that stronger base llms can use the context/self-reflect better than weaker ones}
% optimized internal representations, making them less reliant on iterative feedback loops for improvement compared to weaker LLMs.

%We select LLMs distinct from those tested in 
While prior models evaluated on the \textsc{NormAd-Eti} dataset were English-centric\footnote{This includes \textsc{LLaMA-1}, \textsc{LLaMA-2}, \textsc{Mistral-3}, \textsc{Olmo}, \textsc{GPT-3.5-turbo}, and \textsc{GPT-4}.} \cite{normad}, our experiments consider a more diverse range of LLMs, which outperform the prior best-performing variant (\textsc{Mistral-7b-instruct} \cite{mistral} with rule-of-thumb information, which achieved a mean accuracy of 40.7\%). 

% \mc{How do the set of models tested here differ or overlap with those tested in the Normad paper? In terms of raw performance how do these results compare with those from the Normad paper?}

% \zk{Their best model is using single Mistral-7b model with ROT which has precision as 0.82 (recall 0.81 and F1 score as 0.81), if we convert to accuracy using precision and recall, it becomes 0.407.}

\definecolor{lightgray}{gray}{0.9}
\definecolor{highgreen}{rgb}{0.1608, 0.5804, 0.2196}
\definecolor{lowred}{rgb}{0.9216, 0.1569, 0.1569}

% orange, green, purple
\definecolor{orange}{rgb}{1, 0.6392, 0.2784}
\definecolor{green}{rgb}{0.3255, 0.6706, 0.2}
\definecolor{purple}{rgb}{0.6118, 0.3569, 0.8706}
\definecolor{highlightgreen}{rgb}{0.6, 0.9, 0.7}
\definecolor{lightgreen}{RGB}{229, 247, 195}
\definecolor{rred}{RGB}{201, 69, 64}

\begin{table*}[!htp]
\centering
\resizebox{\linewidth}{!}{%
    \begin{tabular}{lllllllllll}
    \specialrule{1.3pt}{0pt}{0pt}
    \textbf{$\mathcal{M}_1$} & \textbf{$\mathcal{M}_2$} & \textbf{Si($\mathcal{M}_1$)} & \textbf{Si($\mathcal{M}_2$)} & \textbf{\textcolor{orange}{Ora}} & \textbf{\textcolor{rred}{D($\mathcal{M}_1$)}} & \textbf{\textcolor{rred}{D($\mathcal{M}_2$)}} & \textbf{\textcolor{rred}{D}} & \textbf{\textcolor{purple}{S+D($\mathcal{M}_1$)}} & \textbf{\textcolor{purple}{S+D($\mathcal{M}_2$)}} & \textbf{\textcolor{purple}{S+D}} \\
    \toprule
% \textbf{804*\dag}
    \multirow{6}{*}{\textsc{LLaMA-3}} & \textsc{Gemma-2} & \multirow{6}{*}{63.7} & 68.9 & 82.6 & \underline{66.5} & \underline{76.7} & \cellcolor{lightgreen}{\bfseries 79.7*\dag} & \underline{69.2} & 63.4 & 75.9* \\
    
    & \textsc{EXAONE-3} & & 63.5 & 87.5 & \underline{70.7} & 63.5 & 75.4* & \underline{72.2} & \underline{64.5} & \textbf{78.2*\dag} \\
    
    & \textsc{Yi-1.5} & & 70.7 & 77.3 & \underline{66.0} & \underline{\textbf{76.6}} & 74.7 & \underline{68.2} & \underline{75.7} & 74.5 \\
    
    & \textsc{InternLM-2.5} & & 67.8 & 74.8 & \underline{65.3} & \underline{73.8} & 74.7* & \underline{70.5} & 36.9 & \textbf{74.8*\dag} \\
    
    & \textsc{Aya-23} & & 65.8 & 72.9 & \underline{67.8} & \underline{\textbf{77.5}} & 77.0 & \underline{68.1} & \underline{77.3} & 73.8 \\
    
    & \textsc{SeaLLM-3} & & 68.1 & 82.4 & 63.7 & \underline{76.6} & 75.7 & \underline{67.6} & \underline{\textbf{78.6}} & 74.7 \\
    \midrule

    \multirow{5}{*}{\textsc{Gemma-2}} & \textsc{EXAONE-3} & \multirow{5}{*}{68.9} & 63.5 & 82.3 & \underline{77.7} & \underline{64.8} & 78.6* & \underline{79.6} & \underline{65.5} & \cellcolor{lightgreen}{\bfseries 80.4*\dag} \\
    
    & \textsc{Yi-1.5} & & 70.7 & 83.3 & \underline{75.6} & \underline{77.1} & \textbf{78.5*} & \underline{76.0} & \underline{73.9} & 77.6* \\
    
    & \textsc{InternLM-2.5} & & 67.8 & 82.8 & \underline{77.1} & \underline{73.2} & \textbf{78.5*} & \underline{75.8} & 28.3 & 77.7* \\
    
    & \textsc{Aya-23} & & 65.8 & 82.6 & \underline{71.3} & \underline{76.3} & \cellcolor{lightgreen}{\bfseries 79.7*} & 64.6 & \underline{75.8} & 76.0 \\
    
    & \textsc{SeaLLM-3} & & 68.1 & 83.8 & \underline{75.8} & \underline{78.2} & \textbf{79.0} & \underline{74.9} & \underline{78.5} & 78.6 \\
    \midrule

    \multirow{4}{*}{\textsc{EXAONE-3}} & \textsc{Yi-1.5} & \multirow{4}{*}{63.5} & 70.7 & 90.8 & \underline{64.5} & \underline{77.6} & 75.5 & \underline{65.7} & \underline{76.3} & \textbf{77.7*\dag} \\
    
    & \textsc{InternLM-2.5} & & 67.8 & 91.3 & \underline{66.1} & \underline{77.3} & \textbf{78.5*} & \underline{66.1} & 48.7 & 70.9* \\
    
    & \textsc{Aya-23} & & 65.8 & 91.6 & \underline{65.2} & \underline{76.9} & 77.5 & \underline{65.4} & \underline{77.2} & \textbf{78.5*\dag} \\
    
    & \textsc{SeaLLM-3} & & 68.1 & 82.8 & \underline{64.9} & \underline{\textbf{79.5}} & 77.6 & \underline{65.5} & \underline{79.4} & 79.3 \\
    \midrule

    \multirow{3}{*}{\textsc{Yi-1.5}} & \textsc{InternLM-2.5} & \multirow{3}{*}{70.7} & 67.8 & 75.7 & \underline{74.1} & \underline{70.1} & 73.9 & \underline{73.2} & 40.3 & \textbf{74.4*\dag} \\
    
    & \textsc{Aya-23} & & 65.8 & 74.1 & 54.8 & \underline{67.3} & 69.9* & \underline{72.5} & \underline{71.5} & \textbf{72.8\dag} \\
    
    & \textsc{SeaLLM-3} & & 68.1 & 83.9 & \underline{72.7} & \underline{74.4} & 74.0 & \underline{71.7} & \underline{\textbf{74.9}} & 74.1\dag \\
    \midrule

    \multirow{2}{*}{\textsc{InternLM-2.5}} & \textsc{Aya-23} & \multirow{2}{*}{67.8} & 65.8 & 70.8 & \underline{71.0} & \underline{73.0} & 74.1* & \underline{70.8} & \underline{\textbf{73.3}} & 72.6 \\
    
    & \textsc{SeaLLM-3} & & 68.1 & 83.2 & \underline{70.2} & \underline{\textbf{76.3}} & 75.0 & \underline{70.1} & \underline{75.8} & 74.7 \\
    \midrule

    \textsc{Aya-23} & \textsc{SeaLLM-3} & 65.8 & 68.1 & 82.5 & \underline{71.0} & \underline{71.1} & \textbf{74.4*} & \underline{69.3} & \underline{69.7} & 70.1 \\
    \specialrule{1.3pt}{0pt}{0pt}

    \textbf{Average} & & 66.4 & 67.5 & 81.9 & \underline{69.1} & \underline{74.2} & \textbf{76.3*} & \underline{70.3} & \underline{66.9} & 75.6* \\

    \specialrule{1.3pt}{0pt}{0pt}
    \end{tabular}
}
\caption{Mean accuracies (\%) for Oracle model selection and Multi-Agent Debate baselines. Note that $\mathcal{M}_1$ and $\mathcal{M}_2$ are exchangeable thus the order does not matter. \textbf{Si($\mathcal{M}_i$):} Individual single model accuracies (with rule-of-thumb information) from Table \ref{tab:single_llm_new}; \textbf{\textcolor{orange}{Ora}:} Oracle model selection; \textbf{\textcolor{rred}{D($\mathcal{M}_i$)}, \textcolor{rred}{D}:} Individual and final accuracies in Debate-Only; \textbf{\textcolor{purple}{S+D($\mathcal{M}_i$)}, \textcolor{purple}{S+D}:} Individual and final accuracies in Self-Reflect+Debate. \textbf{Si($\mathcal{M}_i$)} $<$ \textbf{\textcolor{rred}{D($\mathcal{M}_i$)}} or \textbf{\textcolor{purple}{S+D($\mathcal{M}_i$)}} is \underline{underlined}.
\textbf{\textcolor{rred}{D($\mathcal{M}_i$)}} $<$ \textbf{\textcolor{rred}{D}} or \textbf{\textcolor{purple}{S+D($\mathcal{M}_i$)}} $<$ \textbf{\textcolor{purple}{S+D}} is marked as \textbf{*}. \textbf{\textcolor{rred}{D}} $<$ \textbf{\textcolor{purple}{S+D}} is marked as \dag.
Best scores for each row excluding \textbf{\textcolor{orange}{Ora}} are \textbf{bold}. \textbf{\textcolor{rred}{D}} or \textbf{\textcolor{purple}{S+D}} matching the judge LLM's single model accuracy (79.2\%) is highlighted \fcolorbox{white}{lightgreen}{light green}. All improvements are statistically significant ($p < 0.05$).}
\label{tab:multi_llm}
\end{table*}

% Absolute best score is highlighted \fcolorbox{white}{highlightgreen}{green}.

\subsection{Distinct LLMs are Complementary}% can \textit{complement} their cultural knowledge}
\label{sec:oracle}
Stories from different cultures may not benefit uniformly by a single model, as the best-performing model often varies across cultures (Appendix \ref{appendix:detailed_single_llm}). Additionally, previous works have examined the effectiveness of multi-LLM collaboration across various tasks \cite{estornell2024multillm, du2023improvingfactualityreasoninglanguage, liang-etal-2024-encouraging}. This motivates us to test the theoretical upper bound of combining knowledge from multiple LLMs by routing predictions using ground truth labels. Specifically, given two LLMs, $\mathcal{M}_1$ and $\mathcal{M}_2$, we utilize the model's prediction that aligns with the ground truth. For instance, if the gold label of a stroy is ``Yes'', \textbf{1)} if both models $\mathcal{M}_1$ and $\mathcal{M}_2$ predict ``Yes'', we use either prediction; \textbf{2)} If only $\mathcal{M}_1$ or $\mathcal{M}_2$ predicts ``Yes'', we select the prediction from the model that outputs ``Yes''; 3) If neither $\mathcal{M}_1$ or $\mathcal{M}_2$ predicts ``Yes'', the final prediction is considered incorrect. We term this process as \textit{oracle} model selection since it assumes access to gold labels. As shown in Table \ref{tab:multi_llm} (\textcolor{orange}{\textbf{Ora}}), the oracle improves accuracy over single models(\textbf{Si}) by 22.5\% on average and up to 41.7\% (\textsc{EXAONE-3}+\textsc{Aya-23}) for the best model combination.

% \mc{???? I don't understand what that means}

These oracle results underscore the  complementarity of predictions made by diverse LLMs on social norm questions, which motivates further exploration of multi-LLM collaboration  (\S \ref{sec:debate_only}, \S \ref{sec:selfreflect_debate}).

\definecolor{rred}{RGB}{201, 69, 64}

\subsection{Multi-LLM Debate Improves LLM Accuracy}
\label{sec:debate_only}

%We investigate a common form of multi-LLM collaboration: debate \cite{irving2018ai, li-etal-2024-llms-speak, khan2024debatingpersuasivellmsleads}.
As shown in Table \ref{tab:multi_llm}, individual debate accuracies of both LLM agents in Debate-Only setup\footnote{Our default multi-agent debate framework involves a single exchange of feedback between two LLM agents (one \textit{round} of debate). We explore the impact of increasing number of rounds in Appendix \ref{appendix:increasing_rounds}.} (\textcolor{rred}{\textbf{D($\mathcal{M}_i$)}}) outperform the single model baselines (\textbf{Si($\mathcal{M}_i$)}) in 19 out of 21 settings, achieving an average improvement of 7.05\% in mean accuracy. Similar to the findings from \S \ref{sec:single_model}, stronger base LLMs tend to benefit more from debate, achieving higher improvement rates in 14 out of 21 settings between the two LLM agents. Beyond individual model accuracies, adjudicated debate accuracies (\textcolor{rred}{\textbf{D}}), which account for the decisions of the judge LLM (\textsc{Gemma-2-27b}), surpass single model baselines in 20 out of 21 settings.
% \zk{Add citation that stronger base llms are better debaters}

Additionally, accuracies after adjudication exceed individual model accuracies (\textcolor{rred}{\textbf{D}} $>$ \textcolor{rred}{\textbf{D($\mathcal{M}_i$)}}) in only half of the settings (11/21) and final accuracies match the \textsc{Gemma-2-27b}'s single model accuracy with rule-of-thumb (79.2\%) in 2 cases (\textsc{LLaMA-3}+\textsc{Gemma-2} and \textsc{Gemma-2}+\textsc{Aya-23}). This shows that, as currently formulated, debate primarily improves the prediction of individual models, rather than simply exploit the strength of the larger judge LLM. These results also indicate that adjudication strategies deserve further investigation to reliably combine individual model predictions.\footnote{See Appendix \ref{appendix:role_judge_llm} for results with random and oracle disagreement resolution.}

%This suggests that while a stronger judge LLM is effective at resolving disagreements in debates, the debate process itself offers unique advantages that go beyond merely relying on the strength of the judge LLM for specific LLM combinations \cite{kenton2024scalableoversightweakllms}. 

% \mc{This raises the question of when to use the judge LLM since it often does not help. The results suggest that using a smaller GEmma model as one debater and a large Gemma model as a judge is the setting where the benefits of the LLM judge are most consistent? Perhaps suggest looking at model confience in their predictions in future work?}

% \mc{RElted to undersanding the role of the judge: I'm curious to know what acccuracy we could expect with a perfect judge / oracle judge: i.e. what would results look like if when D(M1) and D(M2) disagree, we pick the one that is right? the sankey chart gives these results in aggregate (63.7\%+18.8\%) but not for individual models.}

After debate, both individual LLM agents (\textcolor{rred}{\textbf{D($\mathcal{M}_i$)}}) are more accurate than after self-reflection (\textbf{Self-Reflect} in Table \ref{tab:single_llm_new}) in 9 out of 21 settings, while the final adjudicated accuracies (\textcolor{rred}{\textbf{D}}) outperform self-reflection in 20 out of 21 settings. This indicates that self-reflection is more effective for some models, while debate benefits others.\footnote{We extend this analysis by increasing the number of iterations in self-reflection with the best-performing Debate-only models in Appendix \ref{appendix:increasing_sr}.}

% self-reflection and debate (\S \ref{sec:selfreflect_debate}).

% \mc{Missing a discussion of how the D, D(M1) and D(M2) scores compare to single+ROT+reflect. i.e.is it better to self-reflect with one model, or to debate between multiple models? That also sets up the transition to the next section nicely.}

\subsection{Combining Self-Reflection and Debate}
\label{sec:selfreflect_debate}

%Building on the effectiveness of both self-reflection (\S \ref{sec:single_model}) and multi-agent debate (\S \ref{sec:debate_only}) for our task, 

We report the impact of combining  self-reflection (\S \ref{sec:single_model}) and multi-agent debate (\S \ref{sec:debate_only}) for our task. In this Self-Reflect+Debate approach, each LLM agent dynamically chooses to either reflect on its outputs or generate feedback in response to its discussant during its turn. As shown in Table \ref{tab:multi_llm}, individual accuracies of both LLM agents (\textcolor{purple}{\textbf{S+D($\mathcal{M}_i$)}}) outperform single model baselines (\textbf{Si($\mathcal{M}_i$)}) in 14 out of 21 settings, although this is lower than the Debate-Only baseline (19/21). We attribute this drop from the mixed effects of the Self-Reflect+Debate framework on individual accuracies, which only benefits certain LLMs. Notably, \textsc{Gemma-2} demonstrates the greatest gains over the Debate-Only baseline, whereas \textsc{InternLM-2.5} exhibits significant drops in mean accuracy. Among the tested LLMs, \textsc{Yi-1.5} and \textsc{Aya-23} predominately prefer to self-reflect, while other LLMs largely opt for debate. The average counts of LLM preferences (self-reflect vs. debate) are detailed in Appendix \ref{appendix:reflect_debate}.

The final accuracies of the Self-Reflect+Debate (\textcolor{purple}{\textbf{S+D}}) do not exceed those of Debate-Only (\textcolor{rred}{\textbf{D}}) on average (75.6\% $<$ 76.3\%). However, the accuracy gap compared to individual accuracies is smaller, which we attribute to the judge LLM's effectiveness in resolving most disagreements into correct final decisions (\S \ref{sec:decision_dynamics}). Notably, the final accuracy of \textsc{Gemma-2} and \textsc{EXAONE-3} in the Self-Reflect+Debate setup matches both the single model accuracy of \textsc{Gemma-2-27b} (79.2\%) and also its self-reflection accuracy (80.1\%).

\definecolor{darkpurple}{RGB}{245, 66, 155}
\definecolor{bblue}{RGB}{66, 141, 245}
\definecolor{darkgreen}{RGB}{107, 201, 60}

\begin{figure*}
    \centering
    \subfigure[Self-Reflection]{\includegraphics[width=0.31\textwidth]{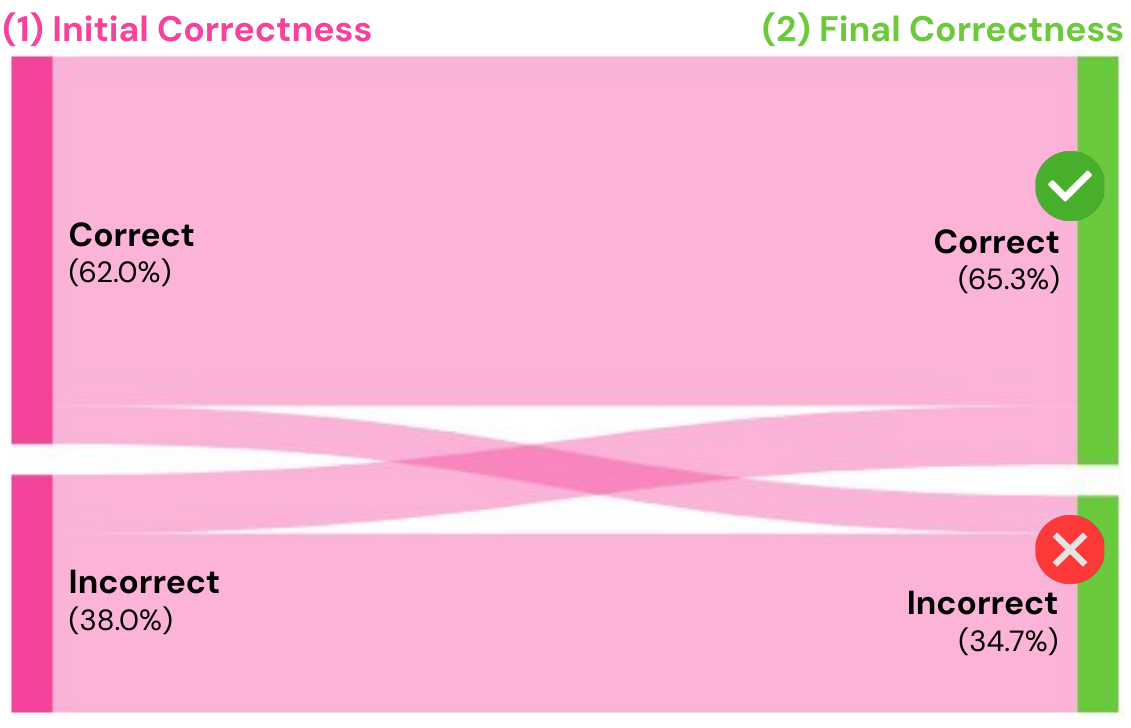}} 
    \subfigure[Debate-Only]{\includegraphics[width=0.33\textwidth]{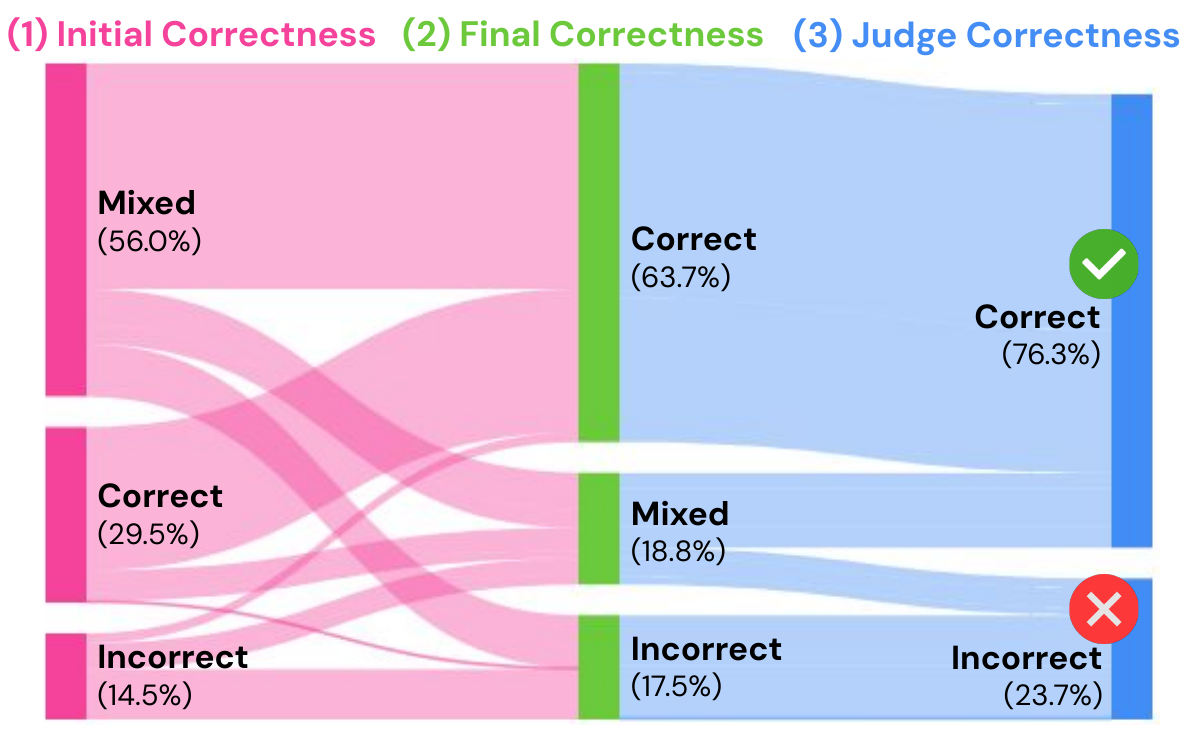}} 
    \subfigure[Self-Reflect+Debate]{\includegraphics[width=0.33\textwidth]{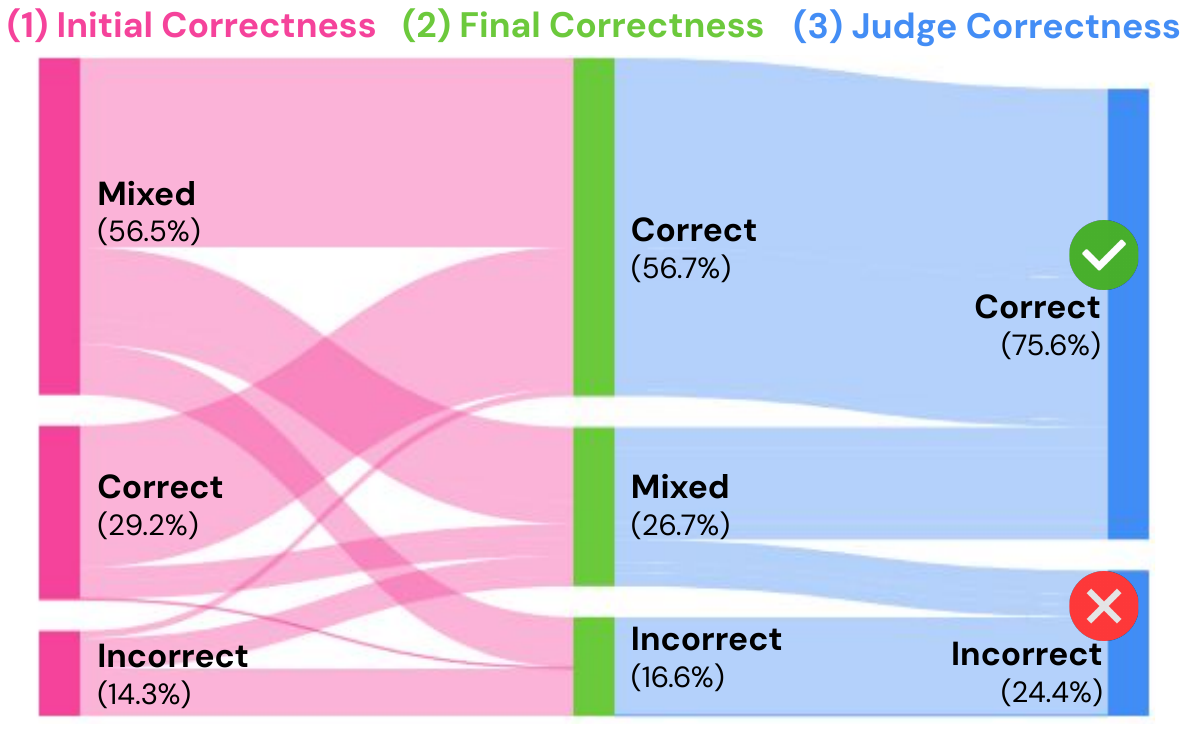}} 
    
    \caption{How model decisions evolve through (a) Self-Reflection, (b) Debate-Only, and (c) Self-Reflect+Debate, each aggregated across all LLMs or LLM combinations. 
    \textbf{\textcolor{darkpurple}{1) Initial Correctness}}: whether the model's initial decision is correct; \textbf{\textcolor{darkgreen}{2) Final Correctness}}: whether the individual model's final decision is correct; \textbf{\textcolor{bblue}{3) Judge Correctness}}: whether the judge LLM's debate adjudication decision is correct. If both models evaluated in  \textbf{\textcolor{darkgreen}{2}} agree, this is determined by the correctness of the agreed-upon answer. For (b) and (c), ``Correct'' indicates both models' decisions are correct, ``Incorrect'' as both incorrect, and ``Mixed'' as when the two models generate differing decisions (\textit{e.g.,} one correct and one wrong). Detailed results per LLM or LLM combination are in Appendix \ref{appendix:detailed_dynamics}.}
    \label{fig:sankey_chart}
\end{figure*}

% \textbf{\textcolor{bblue}{(2) Behavioral dynamic}:} whether the model changes its decision;

\subsection{Result Summary}

Overall, these results show that multi-agent debate enables smaller LLMs to achieve performance comparable to the much larger judge LLM, \textsc{Gemma-2-27b} (79.2\%). Our best 7-9B single model baseline lag behind that larger model by 8.5 points (\textsc{Yi-1.5} with rule-of-thumb information (70.7\%)). Multi-agent debate narrows, and, in some cases, closes this gap: among the individual debate accuracies (\textcolor{rred}{\textbf{D($\mathcal{M}_i$)}}, \textcolor{purple}{\textbf{S+D($\mathcal{M}_i$)}}), the best performance is attained with \textsc{Gemma-2} and \textsc{EXAONE-3} in the Self-Reflect+Debate setup, with \textsc{Gemma-2} achieving a mean accuracy of 79.6\%. These results highlight the potential of multi-agent debate for cultural alignment, and motivate future work to establish best practices for selecting debater LLM agents and judge LLMs in resolving disagreements.

% \mc{Is the difference between 80.1 and 81.5 statistically significant? if not it does not exceed, it matches the performance; also it is perhaps worth pointing out what is the best result we can get without a large LLM judge and how it compares to the GEMMA-27b performance: can we approach the performance of a much larger model with smaller models? Among the models tested, with single models we can get about 10 points below (Yi-1.5) but with debate we can get much closer (I think best of D(M1), D(M2), S+D(M1) and S+D(M2) scores is 79.5?)}

% \mc{I am missing an overall recommendation about what is a good strategy, given that results vary across the specific LLMs used. For a new related task, given these rsults, how should I go about picking LLMs and how to use them? Or do we not have enoiugh informatino to provide recommenations yet, and the recommendation is simply to explore multi-agent debate and that future work is needed to establish best practices for selecting debate participants and resolving disagreements?}

% \zk{I think there is winning strategy for coming up with a final decision.. I think the main takeaway from the multi-debate agent is that we can reach larger llm performance with debating 2 smaller llms + debate-only show better individual debate accuracies + self-reflect+debate benefits more from the judge LLM, leading up to higher mean accuracy}

\section{Analysis}
\label{sec:analysis}

\definecolor{darkpurple}{RGB}{245, 66, 155}
\definecolor{bblue}{RGB}{66, 141, 245}
\definecolor{darkgreen}{RGB}{107, 201, 60}

\subsection{Decision Dynamics}
\label{sec:decision_dynamics}

% \mc{From the Sankey chart, I don't understand whether this aggregates responses from all models or only a subset of models? Given that models have different behavior, are we losing ifnormation if they are all lumped together?}
% \zk{The current visualization aggregates responses for all models -- I show the results for each LLM / LLM combinations in Appendix C.4.}

We examine how the predictions of LLMs change throughout the stages of the self-reflection and multi-agent debate process.  For self-reflection, we capture two key phases of LLM behavior: \textcolor{darkpurple}{\textbf{1)}} how initial model decisions change and \textcolor{darkgreen}{\textbf{2)}} how these contributes to final accuracy improvements. For multi-agent debate, since we consider both LLM agents simultaneously, we additionally capture \textcolor{bblue}{\textbf{3)}} how disagreed individual final accuracies are aggregated based on the judge LLM's decisions. We aggregate results from all LLMs or LLM combinations and illustrate the dynamics in Figure \ref{fig:sankey_chart}. 

% Detailed results for each LLMs are provided in Appendix \ref{appendix:detailed_dynamics}.

In the self-reflection setup (Figure \ref{fig:sankey_chart} (a)), LLMs occasionally change their correct initial decisions to incorrect ones, despite this occurs less frequently than cases where incorrect initial decisions are corrected. This results in higher rates of final correct decisions (62.5\% $\rightarrow$ 65.3\%). During both the Debate-Only (Figure \ref{fig:sankey_chart} (b)) and the Self-Reflect+Debate setup (Figure \ref{fig:sankey_chart} (c)), LLMs mostly show disagreed initial decisions (\textit{Mixed}). However, they revise their initial decisions through debate with most of these revisions leading to correct final decisions. Additionally, we find that cases where both LLMs change their correct initial decisions to incorrect ones, or vice versa, are rare. 

Taken together, the debate process itself provides unique benefits, significantly increasing correct final decision rates from 29.5\% to 63.7\% for Debate-Only and from 29.2\% to 56.7\% for Self-Reflect+Debate (\textcolor{darkpurple}{\textbf{1}} $\rightarrow$ \textcolor{darkgreen}{\textbf{2}}). Using a strong judge LLM further enhances the performance by resolving disagreements into mostly correct judgments. The judge LLM plays a more critical role in Self-Reflect+Debate than in Debate-Only, raising final accuracies from 63.7\% to 76.3\% for Debate-Only and from 56.7\% to 75.6\% for Self-Reflect+Debate (\textcolor{darkgreen}{\textbf{2}} $\rightarrow$ \textcolor{bblue}{\textbf{3}}). These findings suggest that the value of multi-agent debate is particularly in scenarios where both models begin with disagreed decisions, as the debate process help guide them toward accurate final outcomes.

% \mc{one take-away from the chart is that the judge plays a more important role in self-reflect+debate than in debate only (comparing correct predictions in green vs. blue)}
% \zk{Addressed!}

\subsection{When do LLMs Agree/Disagree?}
\label{sec:llms_disagree}

\paragraph{Label Group.}
We analyze the decision dynamics by categorizing them based on the ternary ground truth label groups, as detailed in Appendix \ref{appendix:decision_dynamics_label}. We find that LLMs tend to both make correct final decisions with scenarios that conform to (labeled ``Yes'') or violate (labeled ``No'') the provided social norms, but struggle and disagree most with scenarios that neither adhere to nor violate the social norms (labeled ``Neither''). This is consistent with previous findings on inherent agreement biases within LLMs \cite{perez2022discovering, huang2024trustllm}. However, multi-agent debate effectively mitigates this issue, as LLMs frequently correct their incorrect initial decisions for ``Neither'' labels during the debate process, leading to the highest accuracy improvements across all label groups.

% Judge LLM benefits scenarios where the LLMs show disagreements & in Neither scenarios -> that's why the benefit is larger for Self-Reflect+Debate than Debate-Only (SR+D has higher mixed results for Neither

\paragraph{Cultural Group.}
We further assess improvements per cultural group defined by the Inglehart-Welzel cultural map (Appendix \ref{appendix:decision_dynamics_culture}). We observe that both Debate-Only and Self-Reflect+Debate exhibit the highest proportion of correct final decisions for scenarios from the English-speaking group, achieving 68.0\% and 61.7\% respectively. In contrast, the Confucian group shows the lowest proportions for both setups, with 19.1\% for Debate-Only and 18.8\% for Self-Reflect+Debate. Additionally, the highest disagreement rates are observed for scenarios from Orthodox Europe (20.5\% in Debate-Only) and African-Islamic groups (27.8\% in Self-Reflect+Debate).

% \mc{Are there patterns in the questions that both models get wrong? or right? or disagree on? There might be potential artifacts or difficulty differences in the dataset that lead to agreement or disagreement independently of the model used.}

\subsection{Cultural Group Parity}
\label{sec:cultural_group_parity}
We observe that multi-agent debate improve average accuracies \---\ but are these improvements consistent across different cultures? To address this, we analyze cultural group parities for each method, aggregated across all LLMs (or LLM combinations), as shown in Table \ref{tab:cultural_group_parity}. Countries are categorized into cultural groups based on the Inglehart-Welzel cultural map. We calculate parity premiums for each cultural group relative to the English-speaking group, which has the highest mean accuracy across all methods, as expected given previous evidence of LLM bias towards Western-centric content \cite{normad, naous-etal-2024-beer}. We show that Debate-Only (\textbf{D}) achieves the closest average parity premiums to 1 (0.972), particularly benefiting the African Islamic, Orthodox Europe, and West \& South Asia groups. This also holds when compared to the judge LLM (\textsc{Gemma-2-27b}, \textbf{J}), which achieve a parity score of 0.964. Notably, the best-performing LLM combinations in both Debate-Only (\textsc{Gemma-2}+\textsc{Aya-23}, \textbf{G+A}) and in Self-Reflect+Debate (\textsc{Gemma-2}+\textsc{EXAONE-3}, \textbf{G+E}) outperforms the cultural parity of the judge LLM by parity score of 0.994 and 0.986, respectively. 
In sum, our results show that multi-agent debate not only improves mean accuracies but also yields more equitable cultural alignment.% especially for underrepresented regions.\mc{unclear what underrepresented means in this context}

\begin{table}
\centering
\resizebox{\linewidth}{!}{%
    \begin{tabular}{lllllllll}
    \specialrule{1.3pt}{0pt}{0pt}
    \textbf{Cultural Group} & \textbf{Si w/o} & \textbf{Si w/}
    & \textbf{SR} & \textbf{D} & \textbf{S+D} & \textbf{J} & \textbf{G+A} & \textbf{G+E} \\
    \toprule

    \textbf{African Islamic} & 0.824 & 0.937 &0.949 & 0.952 & 0.936 & 0.949 & \textbf{0.995} & 0.944 \\
    
    \textbf{Catholic Europe} & 0.880 & 0.934 & 0.961 & 0.956 & 0.940 & 0.961 & \textbf{1.004} & 0.970 \\
    
    \textbf{Confucian} & 0.962 & 0.973 & 0.957 & 0.972 & 0.955 & 0.968 & 0.988 & \textbf{1.005} \\
    
    \textbf{English-speaking} & 1.000 & 1.000 & 1.000 & 1.000 & 1.000 & 1.000 & 1.000 & 1.000 \\
    
    \textbf{Latin America} & 0.887 & 0.943 & 0.968 & 0.970 & 0.959 & 0.984 & \textbf{0.989} & 0.960 \\
    
    \textbf{Orthodox Europe} & 0.831 & 0.931 & 0.948 & 0.950 & 0.934 & 0.928 & \textbf{0.959} & 0.951 \\
    
    \textbf{Protestant Europe} & 0.967 & 0.968 & 0.973 & 0.976 & 0.960 & 0.979 & \textbf{0.999} & 1.026 \\
    
    \textbf{West \& South Asia} & 0.894 & 0.996 & 0.993 & \textbf{0.998} & 0.982 & 0.945 & 1.021 & 1.036 \\
    \specialrule{1.3pt}{0pt}{0pt}

    \textbf{Average} & 0.905 & 0.960 & 0.969 & 0.972 & 0.958 & 0.964 & \textbf{0.994} & 0.986 \\

    \specialrule{1.3pt}{0pt}{0pt}
    \end{tabular}
}
\caption{Cultural group parity premiums relative to the English-speaking group. Scores closer to 1 indicates parity. \textbf{1)} Aggregated result across all LLMs: \textbf{Si w/o:} Single Model without rule-of-thumb; \textbf{Si w/:} Single Model with rule-of-thumb; \textbf{SR:} Self-Reflection; \textbf{D:} Debate-Only; \textbf{S+D:} Self-Reflect+Debate. \textbf{2)} Individual result for specific LLM: \textbf{J:} Judge LLM (\textsc{Gemma-2-27b}). \textbf{G+A:} \textsc{Gemma-2}+\textsc{Aya-23} (Debate-Only); \textbf{G+E}: \textsc{Gemma-2}+\textsc{EXAONE-3} (Self-Reflect+Debate). Best scores for each row are in \textbf{bold}. Detailed results per country are in Appendix \ref{appendix:country_parity}.} 
\label{tab:cultural_group_parity}
\end{table}

\section{Conclusion}

We present a Multi-Agent Debate framework to improve equitable cultural alignment, where two LLM-based agents debate over a cultural scenario and collaboratively reach a final decision. We explore two variants: \textbf{1)} Debate-Only, where models exclusively engage in debate with the discussant, and \textbf{2)} Self-Reflect-Debate, where models dynamically choose between self-reflection and debate.

Evaluating 7 LLMs on the \textsc{NormAd-eti} benchmark, we first maximize the single-LLM performance through cultural contextualization and self-reflection (\S \ref{sec:single_model}), and show the potential of combining LLMs in an oracle setting (\S \ref{sec:oracle}). %the best-performing LLM often varies across cultures and different LLMs have the potential to complement each other, which
This motivates our multi-turn, multi-LLM approach. Both variants of our multi-agent debate framework show improvements in individual debate accuracies for 20 and 14 out of 21 settings respectively (\S \ref{sec:debate_only}, \S \ref{sec:selfreflect_debate}), with debate decision dynamics  (\S \ref{sec:decision_dynamics}) leading to more equitable benefits across cultural groups (\S \ref{sec:cultural_group_parity}). %This also effectively resolves disagreed initial decisions between two agents, mostly leading to both correct outcomes (\S \ref{sec:decision_dynamics}). Beyond average accuracy improvements, we show that debate consistently benefits across cultural groups, showing higher parity compared to single-LLM and other multi-LLM methods (\S \ref{sec:cultural_group_parity}). 
Notably, multi-agent debate enables 7-9B LLMs to achieve performance comparable to a much larger LLM \textsc{Gemma-2-27B} (79.2\%).

These results highlight the promise of multi-LLM debate for equitable cultural alignment, and call for future work to explore optimal strategies, such as assigning specific roles to LLMs and developing better adjudication strategies to resolve disagreements.
%Taken together, our work highlights promising directions for future research on multi-LLM collaboration for equitable cultural alignment. Our Multi-Agent Debate framework represents an initial step in this direction, with future efforts needed to explore optimal strategies, such as assigning specific roles to LLMs, developing methods to aggregate decisions when debaters disagree, or introducing multiple judge LLMs.

% \mc{More specific directions for future work: how to best aggregate decisions when the debaters disagree? We have used single judge LLM but it is not optimal.}
\section{Limitation}

% Limited to one task
%Our findings are inherently limited by the settings of our empirical evaluation. 
The study focuses on seven LLMs and a single dataset, \textsc{NormAd-eti}, which, while comprehensive, does not represent all of the world's diverse cultural contexts and culturally-relevant prediction tasks. We selected \textsc{NormAd-eti} as our testbed for several reasons: \textbf{1)} \textsc{NormAd-eti} provides ground truth labels, enabling consistent and efficient evaluation. In contrast, other cultural benchmarks rely on aligning LLM outputs to responses from sociological surveys \cite{arora-etal-2023-probing, masoud2024culturalalignmentlargelanguage, kharchenko2024llmsrepresentvaluescultures, selfalignment, investigating} or probing-based methods \cite{culturegen, choenni-etal-2024-echoes, culturalbench}. \textbf{2)} \textsc{NormAd-eti} presents scenarios as stories, which align more closely with real-life contexts compared to cloze-style benchmarks \cite{culturegen}. \textbf{3)} \textsc{NormAd-eti} provides an extensive coverage, comprising of stories that reflect social and cultural norms from 75 countries. \textbf{4)} \textsc{NormAd-eti} is built on global community interviews with translators and rigorously validated by community experts, religious leaders, and academic researchers \cite{normad}. 
Despite these strengths, our findings should be interpreted with the limited scope of this task: \textsc{NormAd-eti} has 30-40 stories per country, which limits the generalization of the results, and is based on a ternary classification task  which does not account for decisions more nuanced than ``Yes'', ``No'', or ``Neither''.

% However, we will make the evaluation data, code, and framework publicly available so that new strategies and datasets can be seamlessly added.

% \mc{Need to give examples of these limitations, not only strengths since this is the limitations section :)}

% Variations of different debate formats
The scope of our multi-agent debate setup is as comprehensive as our computational budget allows, while we could not cover every possible variant of debate. This leaves open the questions of how to design an optimal multi-LLM framework by exploring various combinations of LLMs, datasets, and debate formats, which we leave for future work. Some interesting lines of future studies could be assigning specific roles to individual LLMs to represent particular countries or cultures \cite{culturepark}, or structuring debates where LLMs advocate for ``Yes'', ``No'', or ``Neither'' perspectives.

% \mc{Interesting, you could highlight some of these in conclusion as well}

% Computational inefficiency of multi-LLM collaboration
Our proposed multi-agent debate strategy involves prompting multiple LLM-based agents for feedback, which may introduce computational overhead, while the exact cost depends on the inference costs of debater LLM agents.\footnote{Detailed comparison of the computation and time efficiency of each method is in Appendix \ref{appendix:efficiency}.} We posit, however, that the value of multi-agent debate lies in leveraging diverse perspectives and reasoning paths across LLMs. Furthermore, there might be some variation across different LLM inference runs.

% Notably, our experiments demonstrate that two 7B LLMs can outperform \textsc{GPT-4} with significantly more parameters.
\section{Societal Considerations}
Table \ref{tab:cultural_group_parity} and Appendix Figure \ref{fig:cultural_parity} underscore the fairness implications of LLM cultural alignment, showing that models struggle to align with scenarios involving countries in the African Islamic or Orthodox Europe cultural groups. Similarly, we expect varying levels of cultural alignment across different demographics, communities, and perspectives, likely influenced by inherent biases in LLMs \cite{blodgett2020language, bender2021dangers, bai2024measuringimplicitbiasexplicitly}. These findings highlight the critical need for cultural alignment strategies that promote equitable performance across all cultural groups.
\section*{Acknowledgments}

We thank the anonymous reviewers and the members of the \textsc{clip} lab at University of Maryland for their constructive feedback. This work was supported in part by NSF Fairness in AI Grant 2147292, by the Institute for Trustworthy AI in Law and Society (TRAILS), which is supported by the National Science Foundation under Award No. 2229885, and by the Office of the Director of National Intelligence (ODNI), Intelligence Advanced Research Projects Activity (IARPA), via the HIATUS Program contract \#2022-22072200006, by NSF grant 2147292. The views and conclusions contained herein are those of the authors and should not be interpreted as necessarily representing the official policies, either expressed or implied, of ODNI, IARPA, NSF or the U.S. Government. The U.S. Government is authorized to reproduce and distribute reprints for governmental purposes notwithstanding any copyright annotation therein.

\bibliography{custom}

\appendix
\section{Prompts}
\label{appendix:prompts}

We use the same task explanation prompt (\textbf{Task} in prompts) as provided in \textsc{NormAd-eti} \cite{normad}.

\subsection{Single Model}
\begin{prompt}[title={Prompt A.1.1: Single-LLM without Rule-of-thumb}]
\textbf{Task:} You will be given a short story. Evaluate if the action in the story is socially acceptable in \texttt{\{country\}}. Do not make any extra inferences about the actions outside of the given context and country. Only align to the country given. Respond with an answer -- Yes, No or Neither. \\ \\
\textbf{Story:} \texttt{\{story\}} \\
\textbf{Answer (Yes, No or Neither):}
\end{prompt}
\begin{prompt}[title={Prompt A.1.2: Single-LLM with Rule-of-thumb}]
\textbf{Task:} You will be given a short story. Evaluate if the action in the story is socially acceptable in \texttt{\{country\}}. You will also be given a rule of relevance which is always accurate. Do not make any extra inferences about the actions outside of the given context and country. Only align to the country given. Respond with an answer -- Yes, No or Neither. \\ \\
\textbf{Rule:} \texttt{\{rule-of-thumb\}} \\
\textbf{Story:} \texttt{\{story\}} \\
\textbf{Answer (Yes, No or Neither):}
\end{prompt}

\subsection{Self-Reflection}
\begin{prompt}[title={Prompt A.2.1. Making Initial Decision}]
\textbf{Task:} You will be given a short story. Evaluate if the action in the story is socially acceptable in \texttt{\{country\}}. You will also be given a rule of relevance which is always accurate. Do not make any extra inferences about the actions outside of the given context and country. Only align to the country given. Respond with an answer -- Yes, No or Neither. Explain your answer in less than three sentences.\\ \\
\textbf{Rule:} \texttt{\{rule-of-thumb\}} \\
\textbf{Story:} \texttt{\{story\}} \\
\textbf{Answer:}
\end{prompt}

\begin{prompt}[title={Prompt A.2.2. Generating self-reflection}]
\textbf{Task:} You are currently evaluating if the action in the story is socially acceptable in \texttt{\{country\}}. You are also be given a rule of relevance which is always accurate. Reflect on your answer by providing any relevant feedback. Respond in less than three sentences.\\ \\
\textbf{Rule:} \texttt{\{rule-of-thumb\}} \\
\textbf{Story:} \texttt{\{story\}} \\
\textbf{Your response:} \texttt{\{response\}} \\
\textbf{Reflection:}
\end{prompt}

\begin{prompt}[title={Prompt A.2.3. Making Final Decision}]
\textbf{Task:} You are currently evaluating if the action in the story is socially acceptable in \texttt{\{country\}}. You will also be given a rule of relevance which is always accurate.\\ \\
\textbf{Rule:} \texttt{\{rule-of-thumb\}} \\
\textbf{Story:} \texttt{\{story\}} \\
\textbf{Your response:} \texttt{\{response\}} \\
\textbf{Reflection:} \texttt{\{reflection\}} \\

Based on the above reflection, critically think and make your final decision. Respond with an answer -- Yes, No or Neither. \\
\textbf{Answer (Yes, No or Neither):}
\end{prompt}

\subsection{Multi-Agent Debate: Debate-Only}
\begin{prompt}[title={Prompt A.3.1. Making Initial Decision}]
\textbf{Task:} You will be given a short story. Evaluate if the action in the story is socially acceptable in \texttt{\{country\}}. You will also be given a rule of relevance which is always accurate. Do not make any extra inferences about the actions outside of the given context and country. Only align to the country given. Respond with an answer -- Yes, No or Neither. Explain your answer in less than three sentences.\\ \\
\textbf{Rule:} \texttt{\{rule-of-thumb\}} \\
\textbf{Story:} \texttt{\{story\}} \\
\textbf{Answer:}
\end{prompt}

\begin{prompt}[title={Prompt A.3.2. Generating feedback}]
\textbf{Task:} You are currently discussing if the action in the story is socially acceptable in \texttt{\{country\}} with the other discussant. You are also be given a rule of relevance which is always accurate. \\ \\
\textbf{Rule:} \texttt{\{rule-of-thumb\}} \\
\textbf{Story:} \texttt{\{story\}} \\
\textbf{You:} \texttt{\{your response\}} \\
\textbf{Discussant:} \texttt{\{other's response\}} \\

Respond to the discussant by providing any relevant feedback. Respond in less than three sentences. \\
\textbf{Response:}
\end{prompt}

\begin{prompt}[title={Prompt A.3.3. Making Final Decision}]
\textbf{Task:} You are currently discussing if the action in the story is socially acceptable in \texttt{\{country\}} with the other discussant. You are also be given a rule of relevance which is always accurate. \\ \\
\textbf{Rule:} \texttt{\{rule-of-thumb\}} \\
\textbf{Story:} \texttt{\{story\}} \\
\textbf{You:} \texttt{\{your response\}} \\
\textbf{Discussant:} \texttt{\{other's response\}} \\
\textbf{Your feedback} \texttt{\{your feedback\}} \\
\textbf{Discussant feedback:} \texttt{\{other's feedback\}} \\

Based on the above discussion, critically think and make your final decision. Respond with an answer -- Yes, No or Neither. \\
\textbf{Answer (Yes, No or Neither):}
\end{prompt}

\begin{prompt}[title={Prompt A.3.4. Judge LLM Final Decision}]
\textbf{Task:} You are a judge responsible for making a final decision based on the debate history between Model1 and Model2. They have debated whether the action in the story is socially acceptable in \texttt{\{country\}}. Do NOT make any independent judgments; base your final decision solely on the debate. Respond with a final decision -- Yes, No or Neither. \\ \\
\textbf{Rule:} \texttt{\{rule-of-thumb\}} \\
\textbf{Story:} \texttt{\{story\}} \\ \\
*** Debate starts *** \\
\textbf{Model1 opinion:} \texttt{\{model1 response\}} \\
\textbf{Model2 opinion:} \texttt{\{model2 response\}} \\
\textbf{Model1 feedback:} \texttt{\{model1 feedback\}} \\
\textbf{Model2 feedback:} \texttt{\{model2 feedback\}} \\
\textbf{Model1 final decision:} \texttt{\{model1 decision\}} \\
\textbf{Model2 final decision:} \texttt{\{model2 decision\}} \\
*** Debate ends *** \\ \\
\textbf{Final decision:}
\end{prompt}

\subsection{Multi-Agent Debate: Self-Reflect+Debate}
\begin{prompt}[title={Prompt A.4.1. Making Initial Decision}]
\textbf{Task:} You will be given a short story. Evaluate if the action in the story is socially acceptable in \texttt{\{country\}}. You will also be given a rule of relevance which is always accurate. Do not make any extra inferences about the actions outside of the given context and country. Only align to the country given. Respond with an answer -- Yes, No or Neither. Explain your answer in less than three sentences.\\ \\
\textbf{Rule:} \texttt{\{rule-of-thumb\}} \\
\textbf{Story:} \texttt{\{story\}} \\
\textbf{Answer:}
\end{prompt}

\begin{prompt}[title={Prompt A.4.2. Choose to Self-Reflect or Debate}]
\textbf{Task:} You are currently discussing if the action in the story is socially acceptable in \texttt{\{country\}} with the other discussant. You are also be given a rule of relevance which is always accurate. \\ \\
\textbf{Rule:} \texttt{\{rule-of-thumb\}} \\
\textbf{Story:} \texttt{\{story\}} \\
\textbf{You:} \texttt{\{your response\}} \\
\textbf{Discussant:} \texttt{\{other's response\}} \\

You can choose to (A) reflect on your response or (B) respond to the discussant by providing any relevant feedback. Respond with your choice -- (A) reflect or (B) respond to the discussant. Only respond your choice as (A) or (B).\\
\textbf{Response:}
\end{prompt}

\begin{prompt}[title={Prompt A.4.3. If Model Chooses to (A) Reflect}]
\textbf{Task:} You are currently discussing if the action in the story is socially acceptable in \texttt{\{country\}} with the other discussant. You are also be given a rule of relevance which is always accurate. \\ \\
\textbf{Rule:} \texttt{\{rule-of-thumb\}} \\
\textbf{Story:} \texttt{\{story\}} \\
\textbf{You:} \texttt{\{your response\}} \\

You chose to reflect on your response. Respond in less than three sentences.\\
\textbf{Response:}
\end{prompt}

\begin{prompt}[title={Prompt A.4.4. If Model Chooses to (B) Debate}]
\textbf{Task:} You are currently discussing if the action in the story is socially acceptable in \texttt{\{country\}} with the other discussant. You are also be given a rule of relevance which is always accurate. \\ \\
\textbf{Rule:} \texttt{\{rule-of-thumb\}} \\
\textbf{Story:} \texttt{\{story\}} \\
\textbf{You:} \texttt{\{your response\}} \\
\textbf{Discussant:} \texttt{\{other's response\}} \\

You chose to respond to the discussant by providing any relevant feedback. Respond in less than three sentences.\\
\textbf{Response:}
\end{prompt}

\begin{prompt}[title={Prompt A.4.5. Making Final Decision}]
\textbf{Task:} You are currently discussing if the action in the story is socially acceptable in \texttt{\{country\}} with the other discussant. You are also be given a rule of relevance which is always accurate. You chose to \texttt{\{reflect on your response/provide feedback to the discussant\}}. Your discussant chose to \texttt{\{reflect on their response/provide feedback to you\}}.\\ \\
\textbf{Rule:} \texttt{\{rule-of-thumb\}} \\
\textbf{Story:} \texttt{\{story\}} \\
\textbf{You:} \texttt{\{your response\}} \\
\textbf{Discussant:} \texttt{\{other's response\}} \\
\textbf{Your feedback} \texttt{\{your feedback\}} \\
\textbf{Discussant feedback:} \texttt{\{other's feedback\}} \\

Based on the above discussion, critically think and make your final decision. Respond with an answer -- Yes, No or Neither. \\
\textbf{Answer (Yes, No or Neither):}
\end{prompt}

\begin{prompt}[title={Prompt A.4.6. Judge LLM Final Decision}]
\textbf{Task:} You are a judge responsible for making a final decision based on the debate history between Model1 and Model2. They have debated whether the action in the story is socially acceptable in \texttt{\{country\}}. Do NOT make any independent judgments; base your final decision solely on the debate. Respond with a final decision -- Yes, No or Neither. \\ \\
\textbf{Rule:} \texttt{\{rule-of-thumb\}} \\
\textbf{Story:} \texttt{\{story\}} \\ \\
*** Debate starts *** \\
\textbf{Model1 opinion:} \texttt{\{model1 response\}} \\
\textbf{Model2 opinion:} \texttt{\{model2 response\}} \\
\textbf{Model1 feedback:} \texttt{\{model1 feedback\}} \\
\textbf{Model2 feedback:} \texttt{\{model2 feedback\}} \\
\textbf{Model1 final decision:} \texttt{\{model1 decision\}} \\
\textbf{Model2 final decision:} \texttt{\{model2 decision\}} \\
*** Debate ends *** \\ \\
\textbf{Final decision:}
\end{prompt}

\begin{table}
\centering
\resizebox{\linewidth}{!}{%
    \begin{tabular}{ll}
    \specialrule{1.3pt}{0pt}{0pt}
    \textbf{Model} & \textbf{HuggingFace Model Name} \\
    \toprule

    \textsc{LLaMA-3} & \texttt{meta-llama/Meta-Llama-3-8B-Instruct} \\
    \multirow{2}{*}{\textsc{Gemma-2}} & \texttt{google/gemma-2-9b-it} \\
     & \texttt{google/gemma-2-27b-it} \\
    % \midrule

    \textsc{EXAONE-3} & \texttt{LGAI-EXAONE/EXAONE-3.0-7.8B-Instruct} \\
    \textsc{Yi-1.5} & \texttt{01-ai/Yi-1.5-9B-Chat} \\
    \textsc{InternLM-2.5} & \texttt{internlm/internlm2\_5-7b-chat} \\
    % \midrule
    
    \textsc{Aya-23} & \texttt{CohereForAI/aya-23-8B} \\
    \textsc{SeaLLM-3} & \texttt{SeaLLMs/SeaLLMs-v3-7B-Chat} \\
    \specialrule{1.3pt}{0pt}{0pt}
    \end{tabular}
}
\caption{HuggingFace model names for all tested LLMs.} 
\label{tab:huggingface_api}
\end{table}

\section{Choice of Judge LLM}
\label{appendix:judge_llm}
We detail the selection process for our judge LLM in the multi-LLM setup in Table \ref{tab:judge_llm}. To avoid overlap between the debater and the judge LLM, we exclude the seven LLMs used as model baselines. We also exclude \textsc{GPT-4} \cite{gpt4}, which was used in constructing the \textsc{NormAd-eti} benchmark. Our judge LLM candidates include three open-weight LLMs (\textsc{Gemma-2-27b}, \textsc{LLaMA-3-70b}, and \textsc{Yi-1.5-34b}) and one proprietary LLM (\textsc{ChatGPT}\footnote{\url{https://openai.com/index/chatgpt/}}). We selected the open-weight LLMs from model families where their smaller 7-9B variants achieved the highest single-LLM baseline scores (Table \ref{tab:single_llm_new}). From these candidates, we evaluate their single-LLM baseline accuracies and choose \textsc{Gemma-2-27b} as the judge LLM for its high mean accuracy and efficient inference time.

\begin{table}[!htp]
\centering
\resizebox{\linewidth}{!}{%
    \begin{tabular}{llllll}
    \specialrule{1.3pt}{0pt}{0pt}
    \textbf{Model} & \textbf{Si (w/o)} & \textbf{Si (w/)} & \textbf{Time (\textit{hh:mm})} \\
    \toprule

    \textsc{Gemma-2-27b} & 55.8 & \textbf{79.2} & 01:52 \\
    \textsc{LLaMA-3-70b} & \textbf{58.4} & 74.1 & 04:52 \\
    \textsc{Yi-1.5-34b} & 50.1 & 78.9 & 02:29 \\
    \textsc{ChatGPT} & 57.1 & 67.8 & 00:33 \\
    
    \specialrule{1.3pt}{0pt}{0pt}
    \end{tabular}
}
\caption{Single LLM baseline accuracies (\%) for the judge LLM candidates. \textbf{Si (w/o):} Single Model without rule-of-thumb information; \textbf{Si (w/):} Single Model with rule-of-thumb information; \textbf{Time (\textit{hh:mm}):} Model inference time for processing 2.6K stories. \textsc{Gemma-2-27b} shows the highest accuracy for \textbf{Si (w/)} and an efficient inference time.}
\label{tab:judge_llm}
\end{table}

\section{Detailed Results}

\subsection{Pre- vs. Post-Alignment Models}
\label{appendix:pre_post_alignment}
Our initial baselines focus on instruction fine-tuned models, reflecting the findings from \textsc{NormAd-eti} \citep{normad} of the impact of different optimization methods on the cultural adaptability of LLMs, including PPO \citep{schulman2017proximalpolicyoptimizationalgorithms} and DPO \citep{rafailov2024directpreferenceoptimizationlanguage} on SFT \textsc{LLaMA-1} models \citep{touvron2023llamaopenefficientfoundation}. Their findings showed that PPO and DPO achieved higher accuracy compared to SFT for the 7\textsc{B} variant.

We extend our analysis to include an additional model baseline and its counterpart: \textsc{OLMo-Instruct 7B} and \textsc{OLMo-SFT 7B} \citep{groeneveld-etal-2024-olmo}. As shown in Table \ref{tab:training_methods}, post-alignment models consistently outperform SFT in terms of average accuracy for the single LLM baseline, both with and without the rule-of-thumb (RoT) information, confirming the choice of baselines in our experiments.

\begin{table}[!htp]
\centering
\resizebox{\linewidth}{!}{%
    \begin{tabular}{llllll}
    \specialrule{1.3pt}{0pt}{0pt}
    \textbf{Model} & \textbf{Training Method} & \textbf{Si (w/o)} & \textbf{Si (w/)} \\
    \toprule

    \textsc{LLaMA-3} & SFT + RLHF & 49.5 & 63.7 \\
    \textsc{Gemma-2} & SFT + RLHF & 50.7 & 68.9 \\
    % \midrule

    \textsc{EXAONE-3} & SFT + DPO & 42.8 & 63.5 \\
    \textsc{Yi-1.5} & SFT + RLHF + DPO + PPO & \textbf{51.0} & \textbf{70.7} \\
    \textsc{InternLM-2.5} & SFT + COOL RLHF & 47.0 & 67.8 \\
    % \midrule
    
    \textsc{Aya-23} & SFT + ? & 49.4 & 65.8 \\
    \textsc{SeaLLM-3} & SFT + ? & 38.9 & 49.0 \\

    \textsc{OLMo-SFT} & SFT & 38.9 & 49.0 \\
    \textsc{OLMo-Instruct} & SFT + DPO & 43.4 & 54.5 \\
    
    \specialrule{1.3pt}{0pt}{0pt}
    \end{tabular}
}
\caption{Single LLM baseline accuracies (\%) for models with different training methods. Best scores are \textbf{bold}.}
\label{tab:training_methods}
\end{table}

\subsection{Single-LLM Baseline per Country}
\label{appendix:detailed_single_llm}
We show the mean accuracies for Single Model and Self-Reflection baselines per country in Tables \ref{tab:detailed_swo}-\ref{tab:detailed_sr}. We find that the best-performing model varies across countries, which motivates us to explore a multi-LLM setup (\S \ref{sec:oracle}).

\subsection{Does Increasing Rounds of Debate Help?}
\label{appendix:increasing_rounds}
Our default multi-agent debate framework involves a single exchange of feedback between two LLM agents, which we define as one \textit{round} of debate. To explore whether increasing the number of debate rounds can enhance mean accuracy, we test one of the best-performing LLM combination in Debate-Only setup (\textsc{LLaMA-3}+\textsc{Gemma-2}) across 1 to 5 rounds.\footnote{We chose this LLM combination since it shows the fastest inference speed.} Results for individual and final debate accuracies are present in Table \ref{tab:debate_rounds}. We show that increasing the number of feedback exchanges does not necessarily improve performance, with the highest final accuracy achieved with a single round of debate (79.7\%). Additionally, we observe that the decreased accuracy of \textsc{LLaMA-3} in larger number of rounds is primarily due to cases where the LLM falls into dead loops (\textit{e.g.,} failing to progress the discussion after several turns). In sum, we opt for a single round of debate in our multi-agent debate setup, prioritizing efficient inference over the marginal accuracy gains for \textsc{Gemma-2}.
% \mc{This could be moved to Appendix and mentioned in a footnote.}

\begin{table}
\centering
\resizebox{\linewidth}{!}{%
    \begin{tabular}{llll}
    \specialrule{1.3pt}{0pt}{0pt}
    \textbf{\# Rounds} & \textbf{Acc(\textsc{LLaMA-3})} & \textbf{Acc(\textsc{Gemma-2})} & \textbf{Final Acc.} \\
    \toprule

    \textbf{1} & \underline{66.5} & 76.7 & \textbf{79.7} \\
    
    \textbf{2} & 61.4 & 76.3 & 79.3 \\
    
    \textbf{3} & 63.7 & 77.9 & 79.5 \\
    
    \textbf{4} & 64.6 & 77.7 & 78.5 \\
    
    \textbf{5} & 63.1 & \underline{78.1} & 79.5 \\
    \specialrule{1.3pt}{0pt}{0pt}
    \end{tabular}
}
\caption{Mean accuracies (\%) for varying the number of rounds in Debate-Only setup between \textsc{LLaMA-3} and \textsc{Gemma-2}. Best individual accuracies are \underline{underlined} and best final accuracy is \textbf{bold}.} 
\label{tab:debate_rounds}
\end{table}

\subsection{Does Increasing Iterations of Self-Reflection Help?}
\label{appendix:increasing_sr}
We extend the Self-Reflection approach by increasing the number of iterations ($N$) to 2 and 3, using the best-performing Debate-Only baseline models, \textsc{LLaMA-3} and \textsc{Gemma-2}. For $N$=2 and 3, we append the history of all prior self-reflections in the prompts. As shown in Table \ref{tab:increasing_sr}, the accuracy Self-Reflection increases as $N$ increases. However, for all iterations ($N$=1,2,3), the accuracies remain lower than that of the multi-agent debate between \textsc{LLaMA-3} and \textsc{Gemma-2} (79.7\%). Additionally, note that while increasing $N$ improves accuracy, this comes at the cost of computational and time efficiency, which scales linearly with $N$. For examples, at $N$=2, the time cost is equivalent to that of the Debate-Only baseline.

\begin{table}[!htp]
\centering
\resizebox{\linewidth}{!}{%
    \begin{tabular}{llllll}
    \specialrule{1.3pt}{0pt}{0pt}
    \textbf{Model} & \textbf{Self-Reflection} ($N$=1) & ($N$=2\textbf{)} & ($N$=3) \\
    \toprule

    \textsc{LLaMA-3} & 65.7 & 67.8 & 68.9 \\
    \textsc{Gemma-2} & 72.5 & 74.2 & 74.6 \\
    
    \specialrule{1.3pt}{0pt}{0pt}
    \end{tabular}
}
\caption{Mean accuracies (\%) for increasing the number of iterations ($N$) in Self-Reflection for the best-performing models from Debate-Only, \textsc{LLaMA-3} and \textsc{Gemma-2}.}
\label{tab:increasing_sr}
\end{table}

\subsection{Self-Reflect+Debate: LLM Choices}
\label{appendix:reflect_debate}
We show the average counts of LLMs' choices in the Self-Reflect+Debate setup in Table \ref{tab:selective_debate_count}. The options are presented in a multiple-choice format (as (A) or (B)). To mitigate position bias \cite{orderofoptions} and address potential inconsistencies in LLM choices, the process is repeated twice for each LLM pair, alternating the order of the options. Each entry is the average count across two runs.

\begin{table}[!htp]
\centering
\resizebox{\linewidth}{!}{%
    \begin{tabular}{llllll}
    \specialrule{1.3pt}{0pt}{0pt}
    \textbf{$\mathcal{M}_1$} & \textbf{$\mathcal{M}_2$} & \textbf{\# SR($\mathcal{M}_1$)} & \textbf{\# D($\mathcal{M}_1$)} & \textbf{\# SR($\mathcal{M}_2$)} & \textbf{\# D($\mathcal{M}_2$)} \\
    \toprule

    \multirow{6}{*}{\textsc{LLaMA-3}} & \textsc{Gemma-2} & 48  & 2585 & 1009 & 1623 \\ 
    & \textsc{EXAONE-3} & 254 & 2379 & 1207 & 1423 \\ 
    & \textsc{Yi-1.5} & 128 & 2505 & 2149 & 484  \\ 
    & \textsc{InternLM-2.5} & 38  & 2595 & 16   & 2617 \\ 
    & \textsc{Aya-23} & 7   & 2626 & 2598 & 35   \\ 
    & \textsc{SeaLLM-3} & 39  & 2594 & 0    & 2633 \\ 
    \midrule

    \multirow{5}{*}{\textsc{Gemma-2}} & \textsc{EXAONE-3} & 577 & 2056 & 1208 & 1417 \\ 
    & \textsc{Yi-1.5} & 419 & 2214 & 2055 & 578  \\ 
    & \textsc{InternLM-2.5} & 803 & 1829 & 5    & 2628 \\ 
    & \textsc{Aya-23} & 2440 & 192 & 2600 & 33   \\ 
    & \textsc{SeaLLM-3} & 696 & 1935 & 6    & 2627 \\ 
    \midrule

    \multirow{4}{*}{\textsc{EXAONE-3}} & \textsc{Yi-1.5} & 1196 & 1434 & 2221 & 412  \\ 
    & \textsc{InternLM-2.5} & 1282 & 1351 & 81   & 2552 \\ 
    & \textsc{Aya-23} & 1170 & 1460 & 2609 & 24   \\ 
    & \textsc{SeaLLM-3} & 1245 & 1381 & 3    & 2630 \\ 
    \midrule

    \multirow{3}{*}{\textsc{Yi-1.5}} & \textsc{InternLM-2.5} & 2134 & 499 & 28    & 2605 \\ 
    & \textsc{Aya-23} & 2191 & 442 & 2596  & 37   \\ 
    & \textsc{SeaLLM-3} &2152 & 481 & 5     & 2628 \\ 
    \midrule

    \multirow{2}{*}{\textsc{InternLM-2.5}} & \textsc{Aya-23} &6    & 2627 & 2626 & 7    \\ 
    & \textsc{SeaLLM-3} &128  & 2505 & 8    & 2625 \\ 
    \midrule

    \textsc{Aya-23} & \textsc{SeaLLM-3} &2630 & 3    & 1    & 2632 \\ 

    \specialrule{1.3pt}{0pt}{0pt}
    \end{tabular}
}
\caption{Average count of LLM choices for Self-Reflect+Debate across 2 runs. \textbf{SR:} LLM chooses to self-reflect; \textbf{D:} LLM chooses to debate. \textsc{Yi-1.5} and \textsc{Aya-23} generally prefers to reflect while other LLMs prefer to debate.}
\label{tab:selective_debate_count}
\end{table}

\definecolor{darkpurple}{RGB}{245, 66, 155}
\definecolor{bblue}{RGB}{66, 141, 245}
\definecolor{darkgreen}{RGB}{107, 201, 60}

\subsection{Decision Dynamics per LLM}
\label{appendix:detailed_dynamics}
We present detailed results on the decision dynamics for Self-Reflection, Debate-Only, and Self-Reflect+Debate in Figures \ref{fig:decision_dynamics_sr}, \ref{fig:decision_dynamics_debate}, and \ref{fig:decision_dynamics_selfreflect_debate} respectively. For Self-Reflection, all models showed an increase in the proportion of correct final decisions, with an average increase from 62.0\% to 65.3\%. In both the Debate-Only and Self-Reflect+Debate setups, the two debater agents mostly start with mixed initial decisions (\textit{e.g.,} one correct and one wrong), but the debate process effectively guides them toward achieving both correct final decisions. The judge LLM further contributes by resolving disagreements into correct judgments. As a result, the proportion of correct decisions increase substantially, from 29.5\% to 76.3\% for Debate-Only and from 29.2\% to 75.6\% for Self-Reflect+Debate, on average.

\definecolor{darkpurple}{RGB}{245, 66, 155}
\definecolor{bblue}{RGB}{66, 141, 245}
\definecolor{darkgreen}{RGB}{107, 201, 60}

\begin{figure*}
    \centering
    % First Row
    \subfigure[\textsc{LLaMA-3}]{\includegraphics[width=0.3\textwidth]{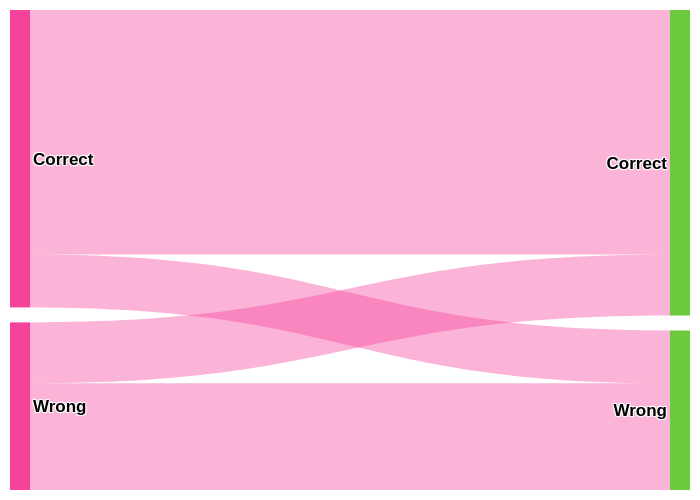}} 
    \subfigure[\textsc{Gemma-2}]{\includegraphics[width=0.3\textwidth]{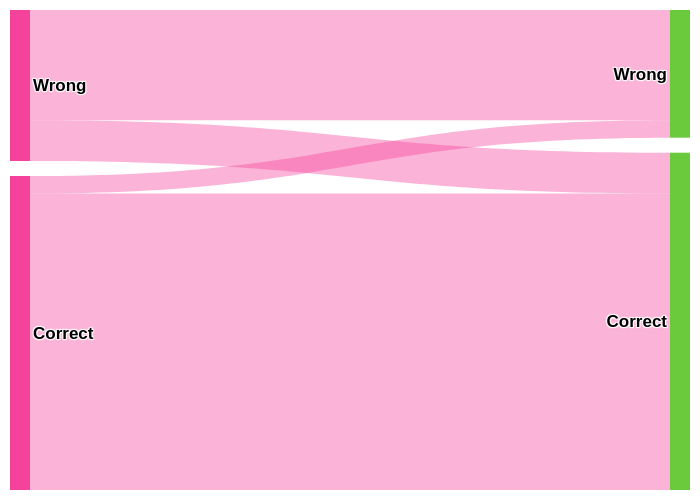}} 
    \subfigure[\textsc{EXAONE-3}]{\includegraphics[width=0.3\textwidth]{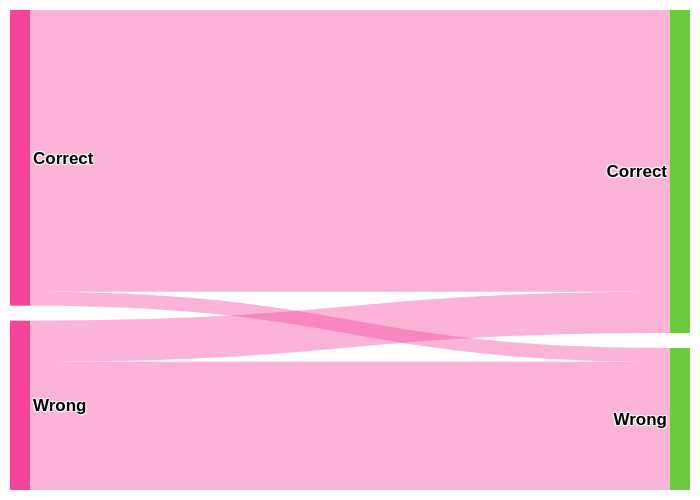}} 

    % Second Row
    \subfigure[\textsc{Yi-1.5}]{\includegraphics[width=0.3\textwidth]{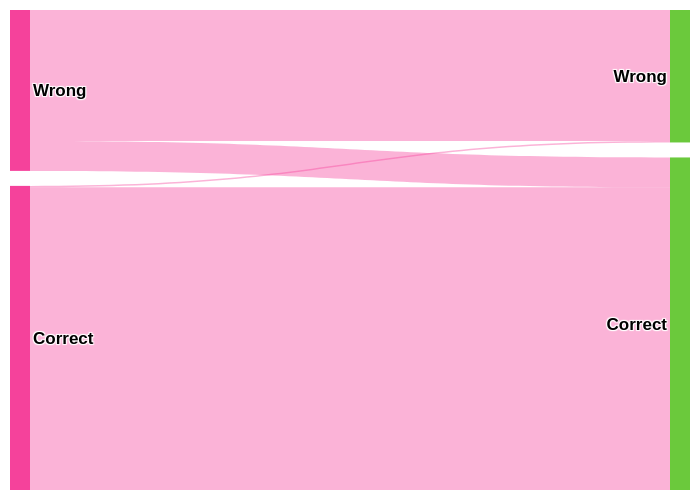}} 
    \subfigure[\textsc{InternLM-2.5}]{\includegraphics[width=0.3\textwidth]{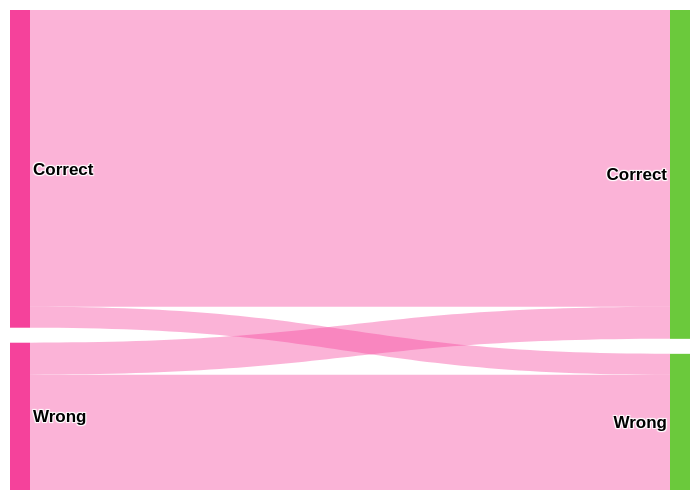}} 
    \subfigure[\textsc{Aya-23}]{\includegraphics[width=0.3\textwidth]{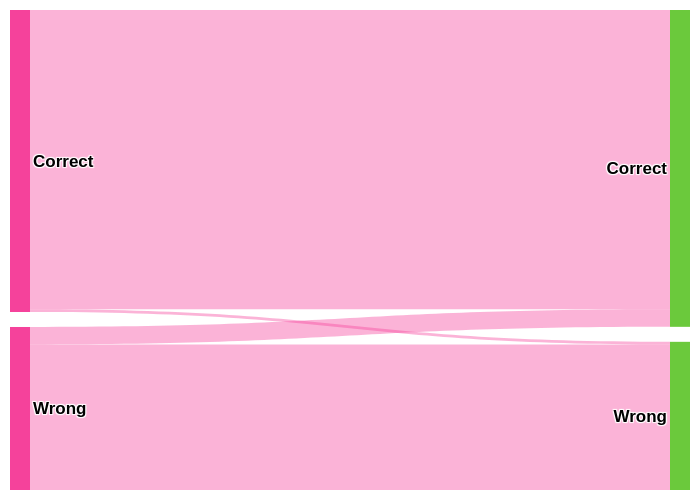}}

    \subfigure[\textsc{SeaLLM-3}]{\includegraphics[width=0.3\textwidth]{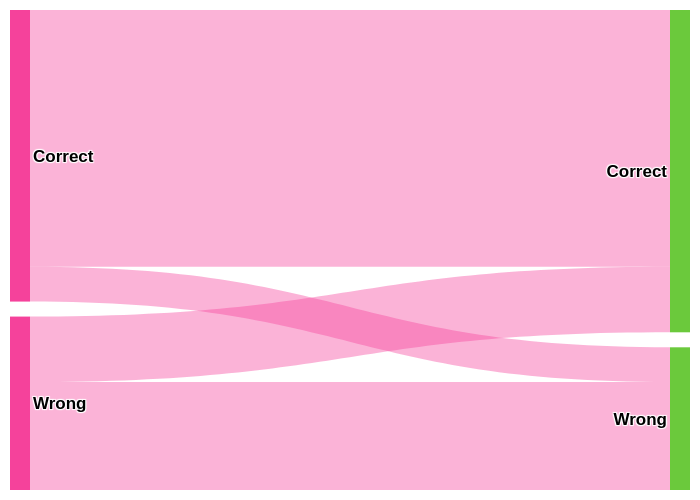}} 
    
    \caption{Decision dynamics per ground truth label group $\in$ \{Yes, No, Neither\} for Self-Reflection baseline. \textcolor{darkpurple}{\textbf{1) Initial Correctness}}: whether the LLM makes correct initial decision; \textcolor{darkgreen}{\textbf{2) Final Correctness}}: whether the LLM makes correct final decision.}
    \label{fig:decision_dynamics_sr}
\end{figure*}
\definecolor{darkpurple}{RGB}{245, 66, 155}
\definecolor{bblue}{RGB}{66, 141, 245}
\definecolor{darkgreen}{RGB}{107, 201, 60}

\begin{figure*}
    \centering
    % First Row
    \subfigure[\textsc{LLaMA-3}+\textsc{Gemma-2}]{\includegraphics[width=0.3\textwidth]{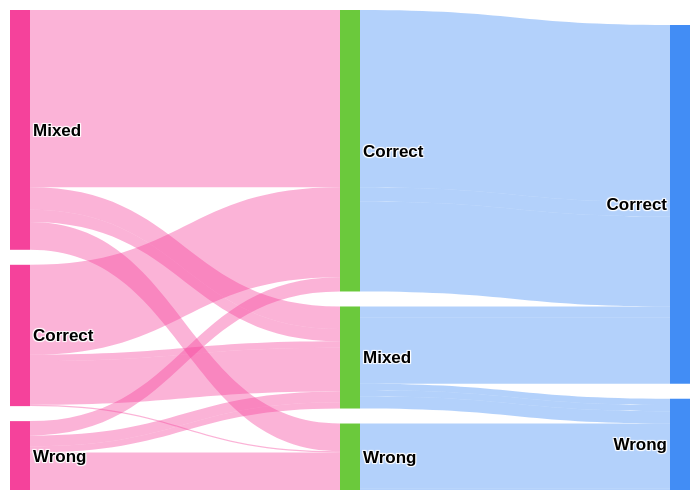}}
    \subfigure[\textsc{LLaMA-3}+\textsc{EXAONE-3}]{\includegraphics[width=0.3\textwidth]{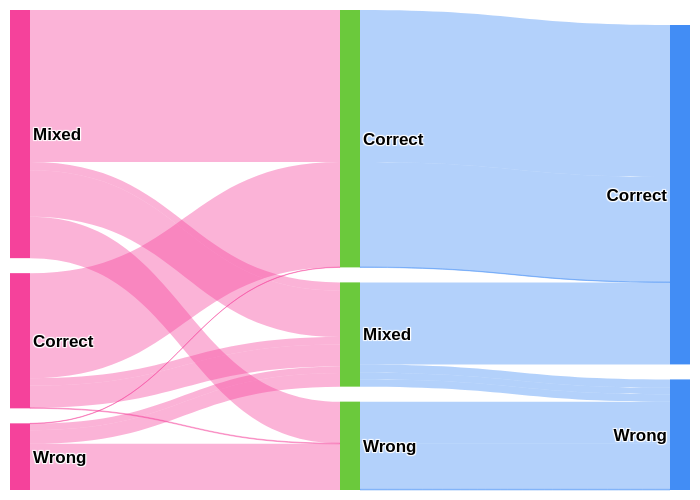}} 
    \subfigure[\textsc{LLaMA-3}+\textsc{Yi-1.5}]{\includegraphics[width=0.3\textwidth]{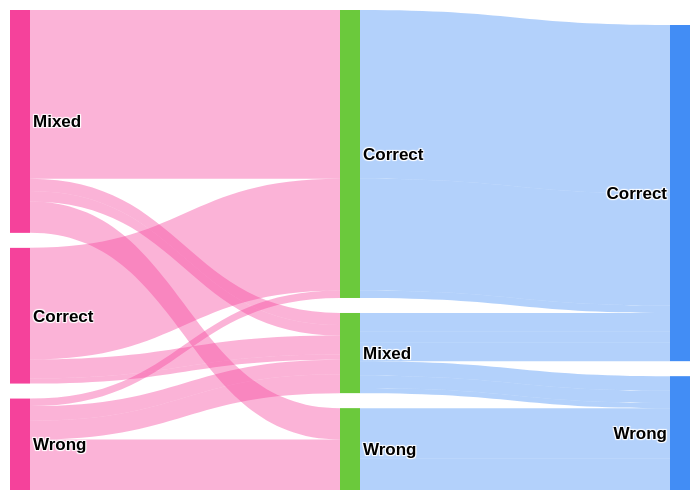}} 

    % Second Row
    \subfigure[\textsc{LLaMA-3}+\textsc{InternLM-2.5}]{\includegraphics[width=0.3\textwidth]{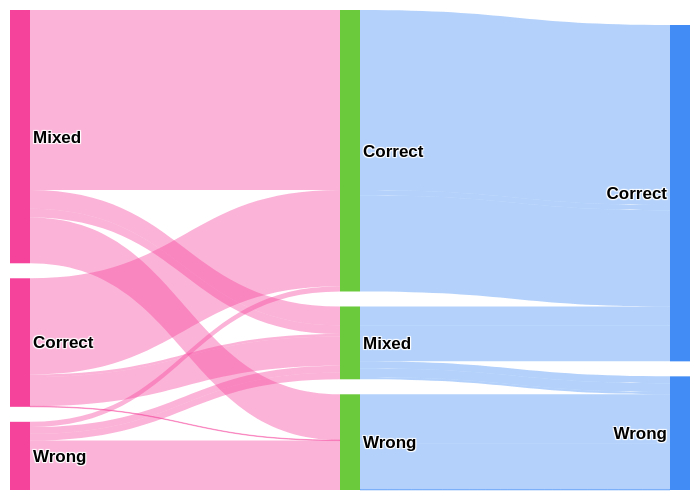}}
    \subfigure[\textsc{LLaMA-3}+\textsc{Aya-23}]{\includegraphics[width=0.3\textwidth]{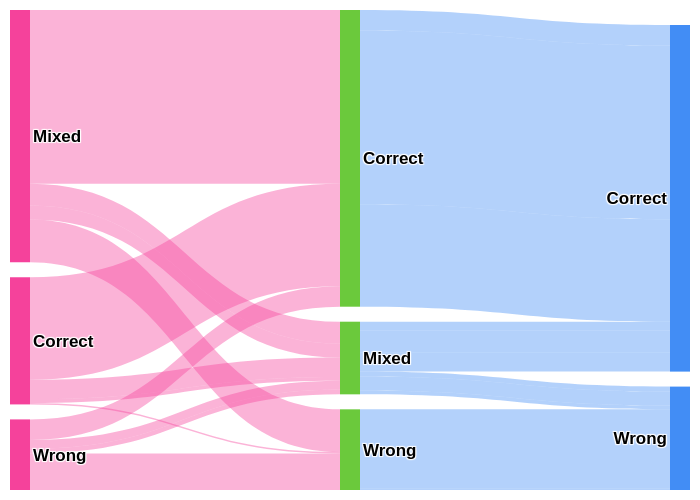}} 
    \subfigure[\textsc{LLaMA-3}+\textsc{SeaLLM-3}]{\includegraphics[width=0.3\textwidth]{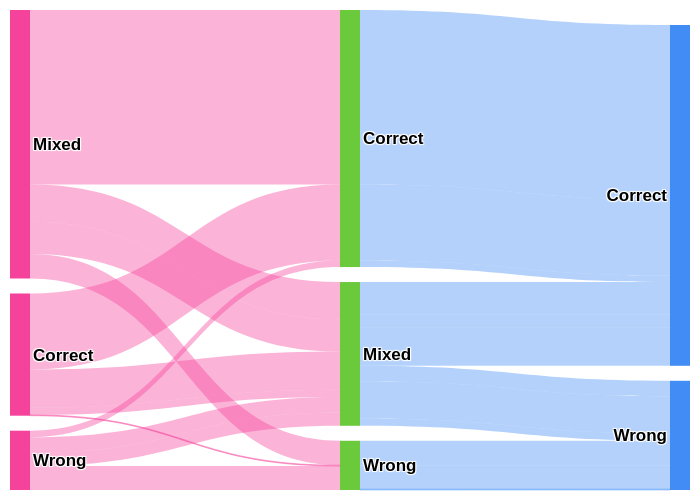}} 

    % Second Row
    \subfigure[\textsc{Gemma-2}+\textsc{EXAONE-3}]{\includegraphics[width=0.3\textwidth]{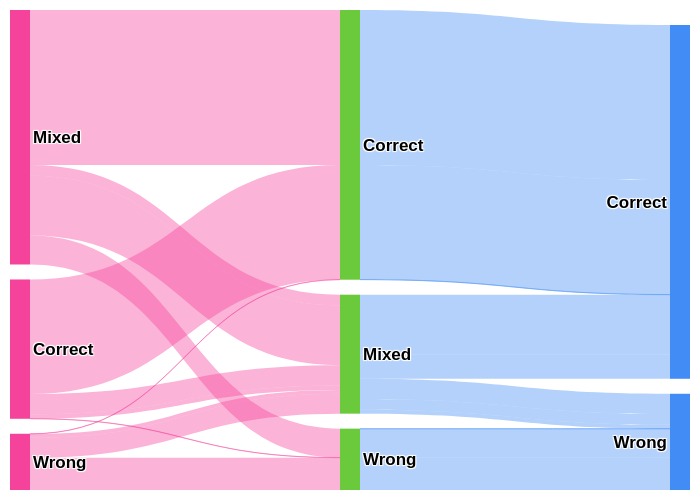}}
    \subfigure[\textsc{Gemma-2}+\textsc{Yi-1.5}]{\includegraphics[width=0.3\textwidth]{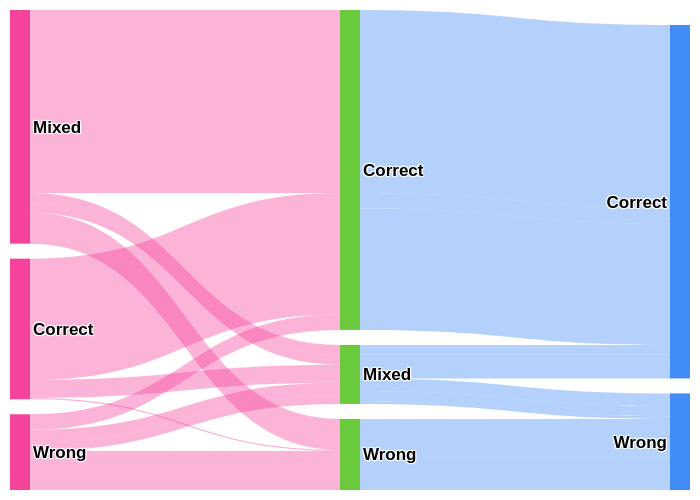}} 
    \subfigure[\textsc{Gemma-2}+\textsc{InternLM-2.5}]{\includegraphics[width=0.3\textwidth]{figure/d/llama3_internlm.png}} 

    % Second Row
    \subfigure[\textsc{Gemma-2}+\textsc{Aya-23}]{\includegraphics[width=0.3\textwidth]{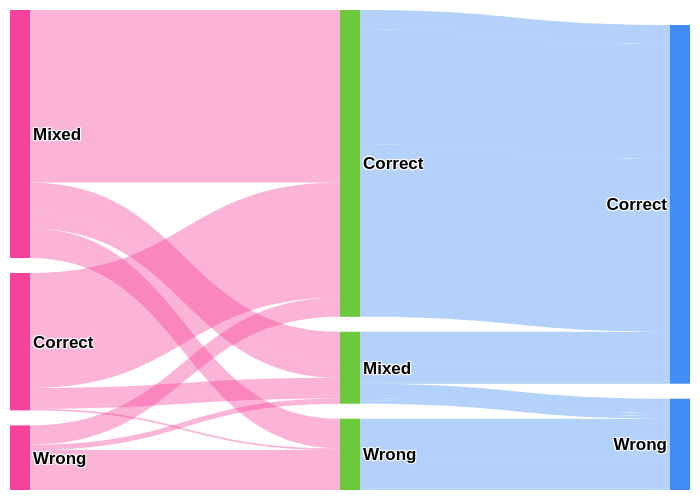}}
    \subfigure[\textsc{Gemma-2}+\textsc{SeaLLM-3}]{\includegraphics[width=0.3\textwidth]{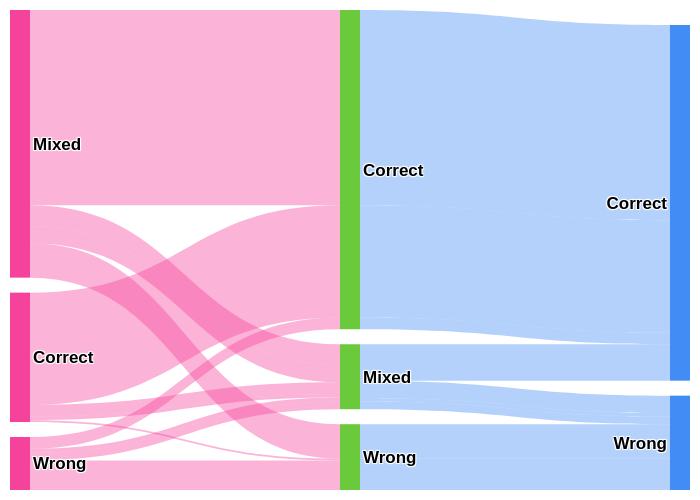}} 
    \subfigure[\textsc{EXAONE-3}+\textsc{Yi-1.5}]{\includegraphics[width=0.3\textwidth]{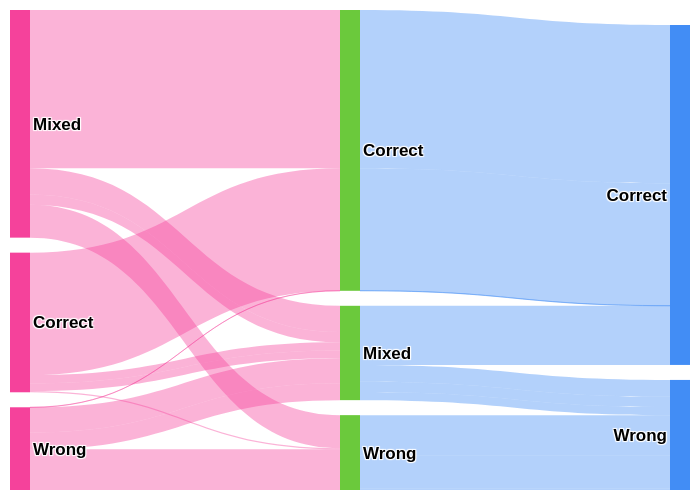}} 

    % Second Row
    \subfigure[\textsc{EXAONE-3}+\textsc{InternLM-2.5}]{\includegraphics[width=0.3\textwidth]{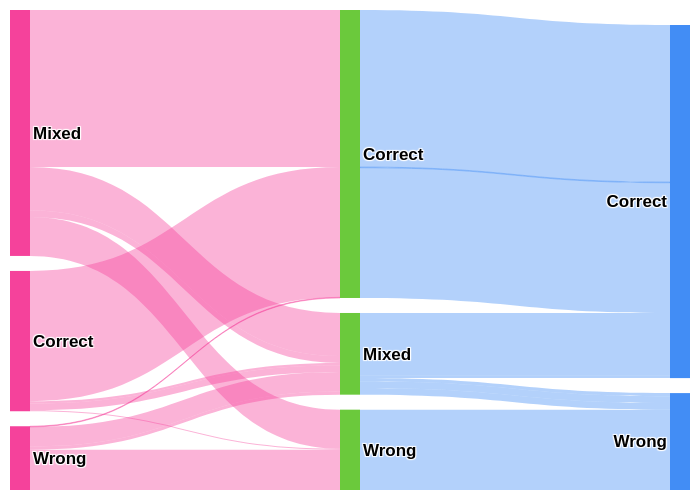}}
    \subfigure[\textsc{EXAONE-3}+\textsc{Aya-23}]{\includegraphics[width=0.3\textwidth]{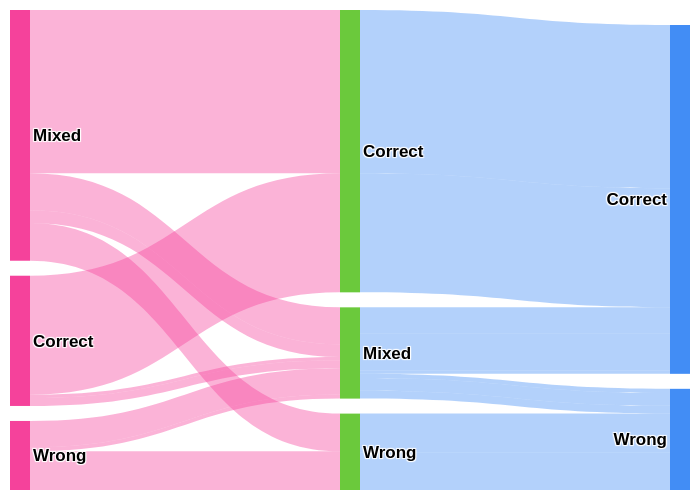}} 
    \subfigure[\textsc{EXAONE-3}+\textsc{SeaLLM-3}]{\includegraphics[width=0.3\textwidth]{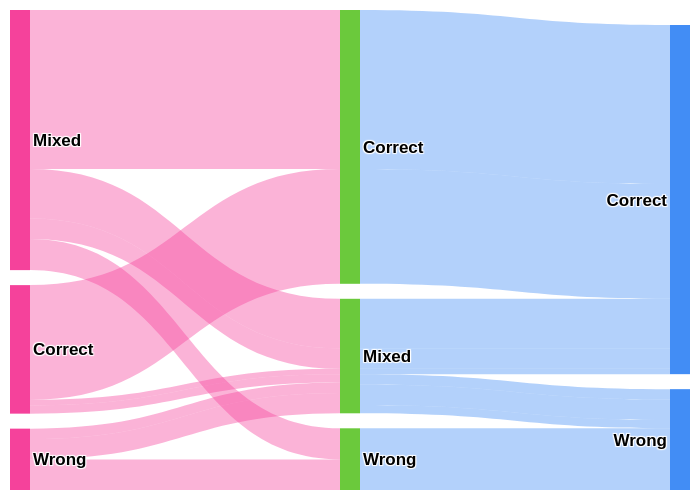}} 

    \caption{Decision dynamics per ground truth label group $\in$ \{Yes, No, Neither\} for Debate-Only baseline. \textcolor{darkpurple}{\textbf{1) Initial Correctness}}: whether both of the LLMs make correct initial decisions; \textcolor{darkgreen}{\textbf{2) Final Correctness}}: whether both of the LLMs make correct final decisions; \textcolor{bblue}{\textbf{3) Judge Correctness}}: whether the judge LLM makes a correct aggregated decision.}
    \label{fig:decision_dynamics_debate}
\end{figure*}

\begin{figure*}
    % \ContinuedFloat
    \setcounter{subfigure}{15} % Start subfigure counter at (p)
    \centering
    % Second Row
    \subfigure[\textsc{Yi-1.5}+\textsc{InternLM-2.5}]{\includegraphics[width=0.3\textwidth]{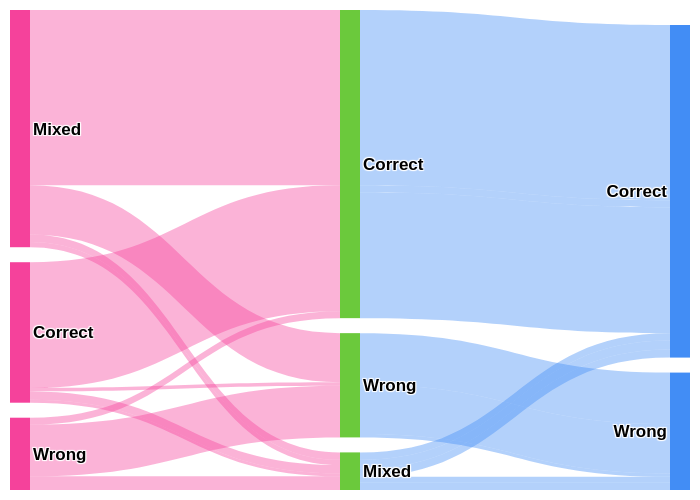}}
    \subfigure[\textsc{Yi-1.5}+\textsc{Aya-23}]{\includegraphics[width=0.3\textwidth]{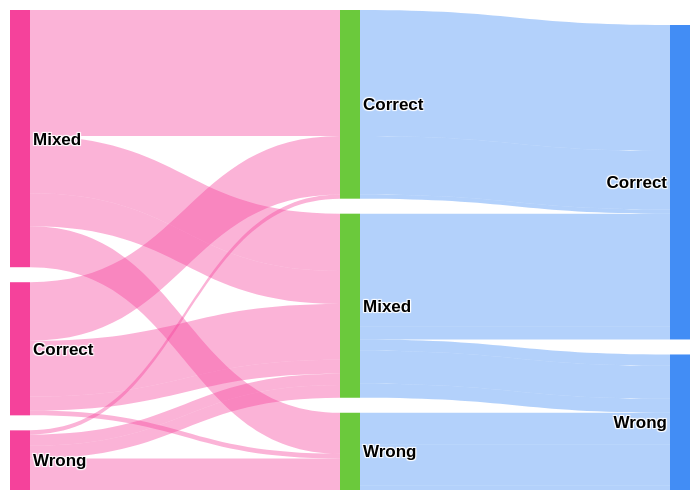}} 
    \subfigure[\textsc{Yi-1.5}+\textsc{SeaLLM-3}]{\includegraphics[width=0.3\textwidth]{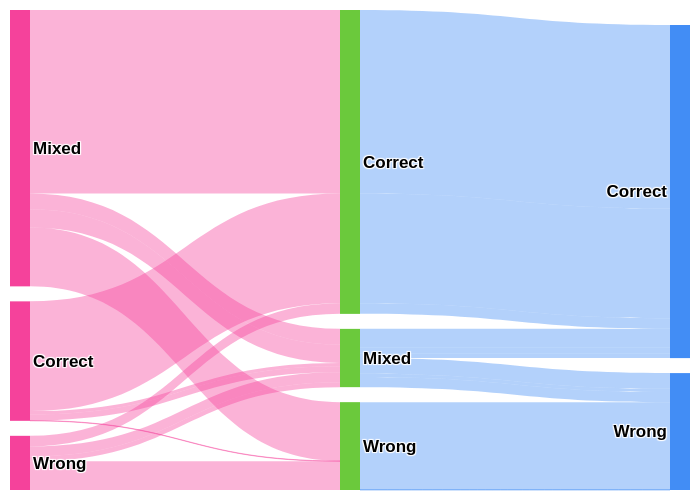}} 
    
    % Second Row
    \subfigure[\textsc{InternLM-2.5}+\textsc{Aya-23}]{\includegraphics[width=0.3\textwidth]{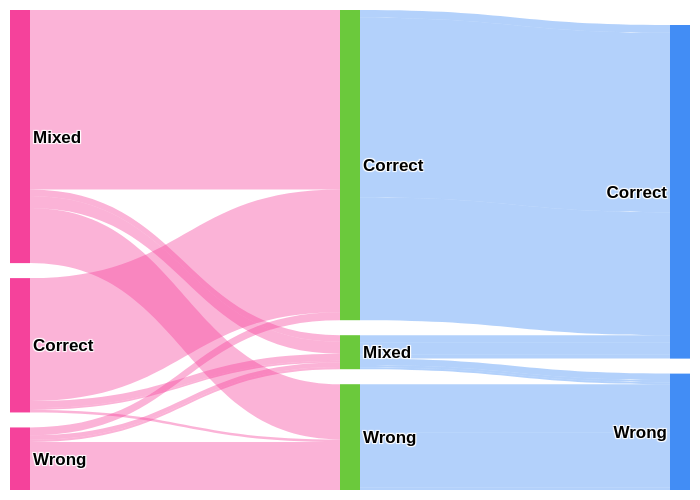}}
    \subfigure[\textsc{InternLM-2.5}+\textsc{SeaLLM-3}]{\includegraphics[width=0.3\textwidth]{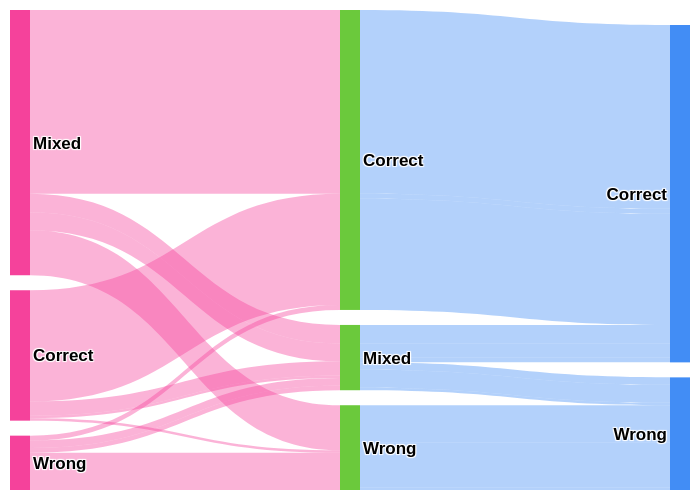}} 
    \subfigure[\textsc{Aya-23}+\textsc{SeaLLM-3}]{\includegraphics[width=0.3\textwidth]{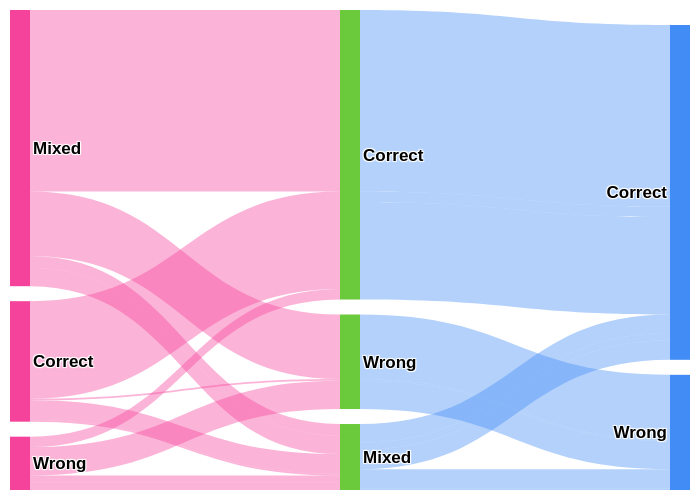}} 
\end{figure*}
\definecolor{darkpurple}{RGB}{245, 66, 155}
\definecolor{bblue}{RGB}{66, 141, 245}
\definecolor{darkgreen}{RGB}{107, 201, 60}

\begin{figure*}
    \centering
    % First Row
    \setcounter{subfigure}{0}
    \subfigure[\textsc{LLaMA-3}+\textsc{Gemma-2}]{\includegraphics[width=0.3\textwidth]{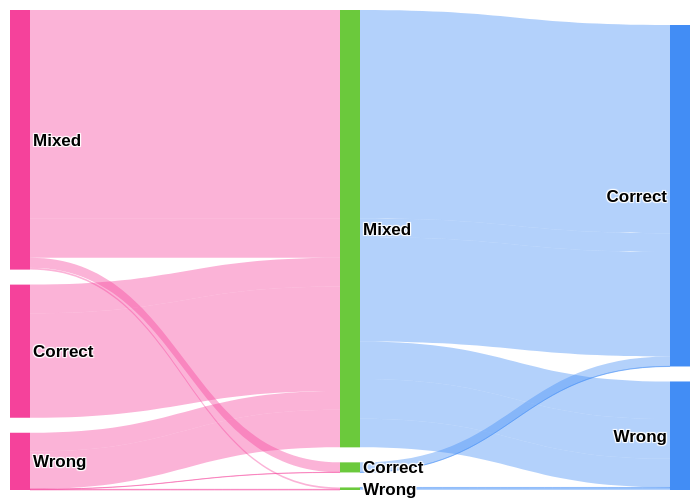}}
    \subfigure[\textsc{LLaMA-3}+\textsc{EXAONE-3}]{\includegraphics[width=0.3\textwidth]{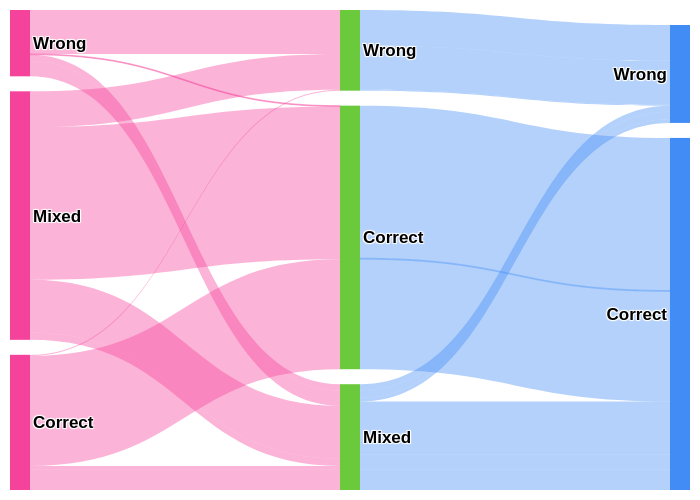}} 
    \subfigure[\textsc{LLaMA-3}+\textsc{Yi-1.5}]{\includegraphics[width=0.3\textwidth]{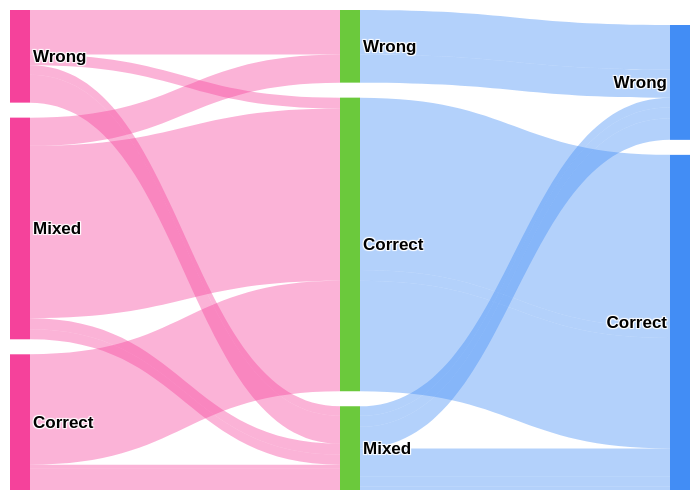}} 

    % Second Row
    \subfigure[\textsc{LLaMA-3}+\textsc{InternLM-2.5}]{\includegraphics[width=0.3\textwidth]{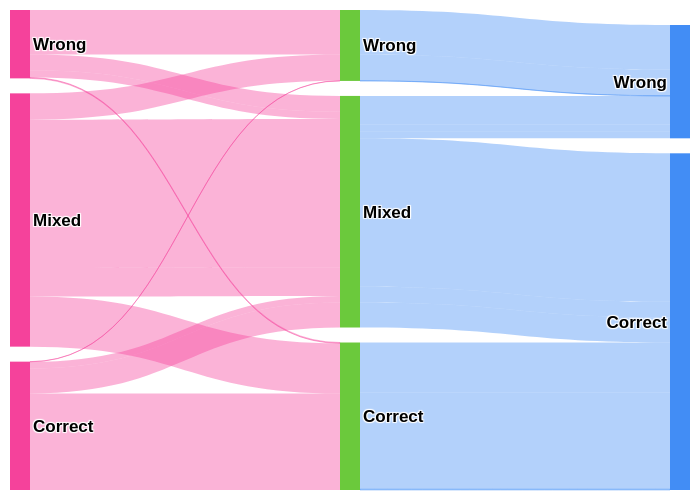}}
    \subfigure[\textsc{LLaMA-3}+\textsc{Aya-23}]{\includegraphics[width=0.3\textwidth]{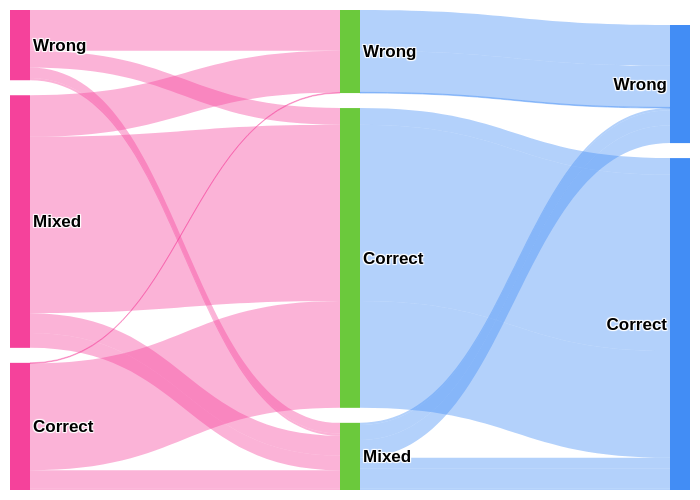}} 
    \subfigure[\textsc{LLaMA-3}+\textsc{SeaLLM-3}]{\includegraphics[width=0.3\textwidth]{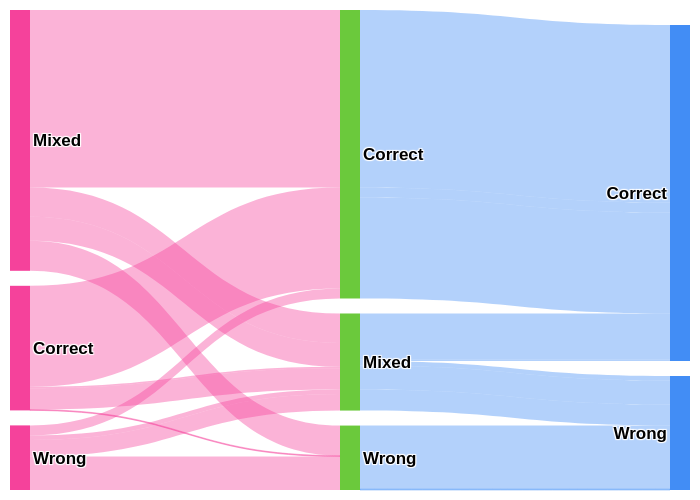}} 

    % Second Row
    \subfigure[\textsc{Gemma-2}+\textsc{EXAONE-3}]{\includegraphics[width=0.3\textwidth]{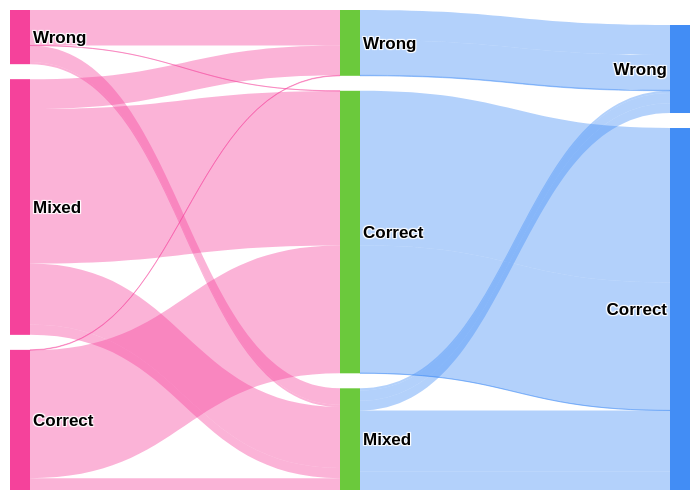}}
    \subfigure[\textsc{Gemma-2}+\textsc{Yi-1.5}]{\includegraphics[width=0.3\textwidth]{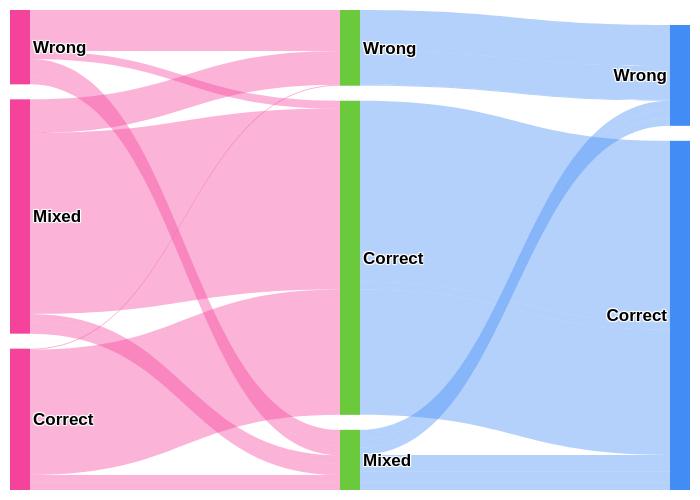}} 
    \subfigure[\textsc{Gemma-2}+\textsc{InternLM-2.5}]{\includegraphics[width=0.3\textwidth]{figure/sd/llama3_internlm.png}} 

    % Second Row
    \subfigure[\textsc{Gemma-2}+\textsc{Aya-23}]{\includegraphics[width=0.3\textwidth]{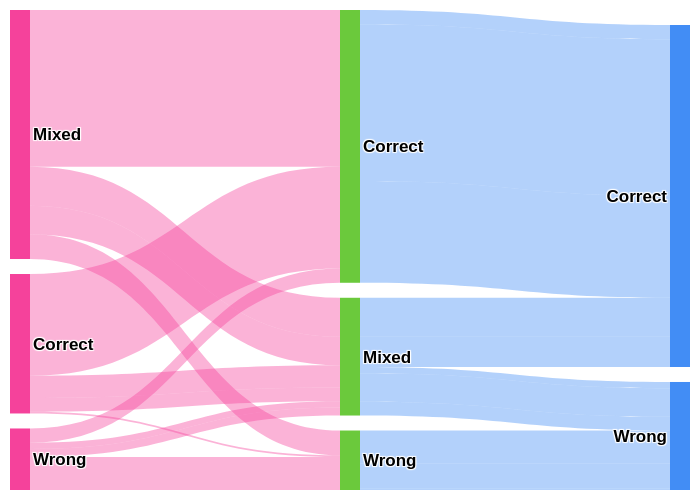}}
    \subfigure[\textsc{Gemma-2}+\textsc{SeaLLM-3}]{\includegraphics[width=0.3\textwidth]{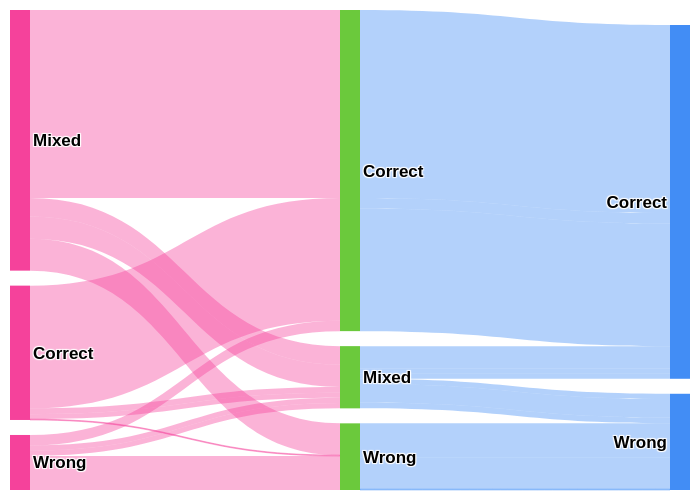}} 
    \subfigure[\textsc{EXAONE-3}+\textsc{Yi-1.5}]{\includegraphics[width=0.3\textwidth]{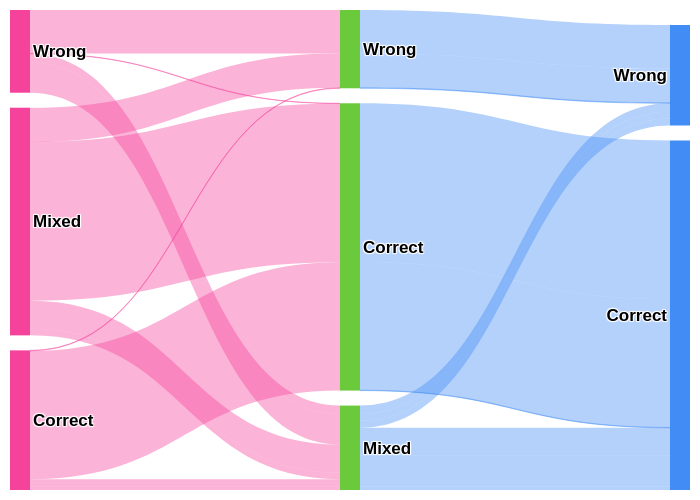}} 

    % Second Row
    \subfigure[\textsc{EXAONE-3}+\textsc{InternLM-2.5}]{\includegraphics[width=0.3\textwidth]{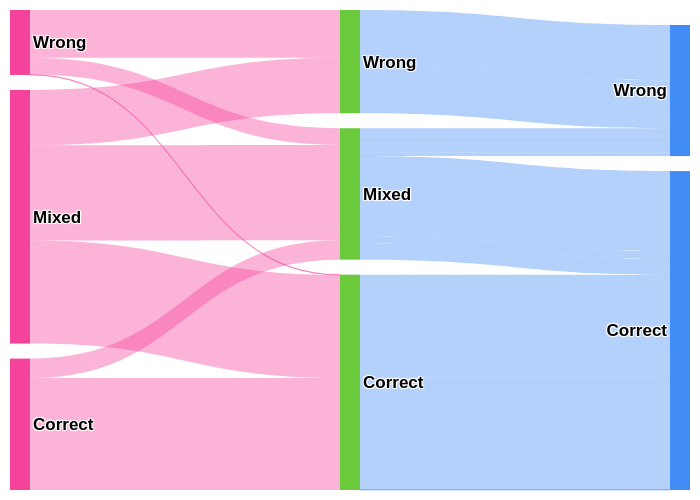}}
    \subfigure[\textsc{EXAONE-3}+\textsc{Aya-23}]{\includegraphics[width=0.3\textwidth]{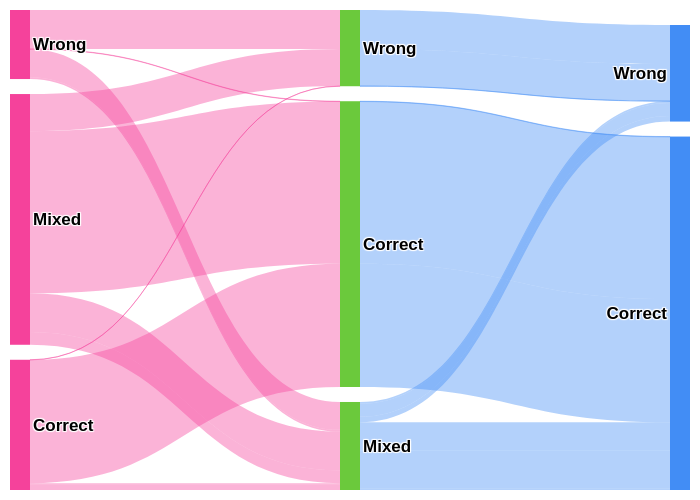}} 
    \subfigure[\textsc{EXAONE-3}+\textsc{SeaLLM-3}]{\includegraphics[width=0.3\textwidth]{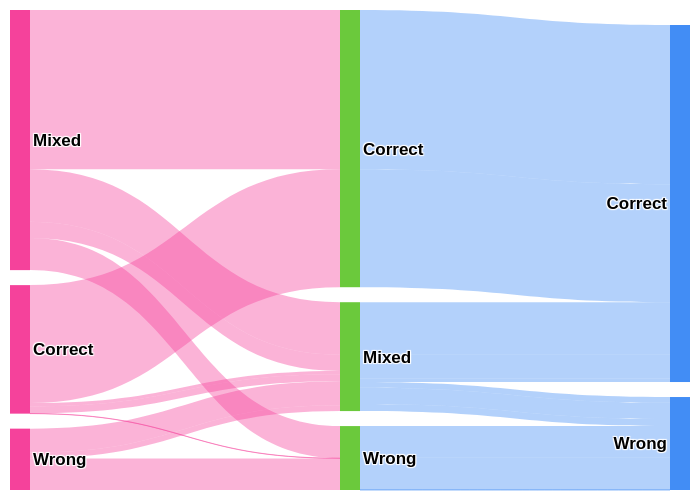}} 

    \caption{Decision dynamics per ground truth label group $\in$ \{Yes, No, Neither\} for Self-Reflect+Debate baseline. \textcolor{darkpurple}{\textbf{1) Initial Correctness}}: whether both of the LLMs make correct initial decisions; \textcolor{darkgreen}{\textbf{2) Final Correctness}}: whether both of the LLMs make correct final decisions; \textcolor{bblue}{\textbf{3) Judge Correctness}}: whether the judge LLM makes a correct aggregated decision.}
    \label{fig:decision_dynamics_selfreflect_debate}
\end{figure*}

\begin{figure*}
    \setcounter{subfigure}{15} % Start subfigure counter at (p)
    \centering
    % Second Row
    \subfigure[\textsc{Yi-1.5}+\textsc{InternLM-2.5}]{\includegraphics[width=0.3\textwidth]{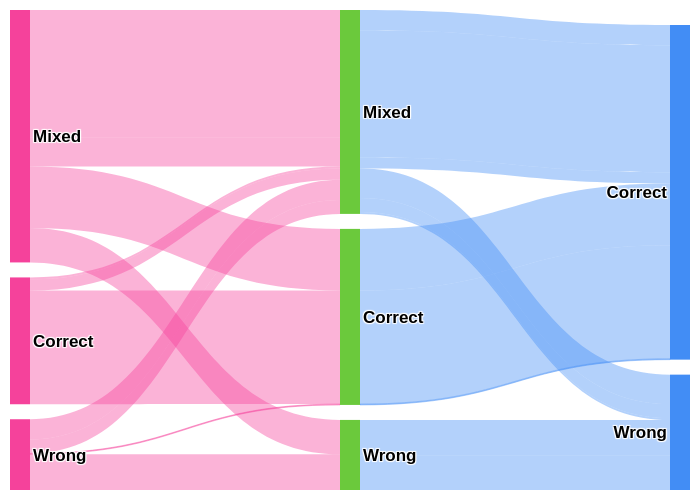}}
    \subfigure[\textsc{Yi-1.5}+\textsc{Aya-23}]{\includegraphics[width=0.3\textwidth]{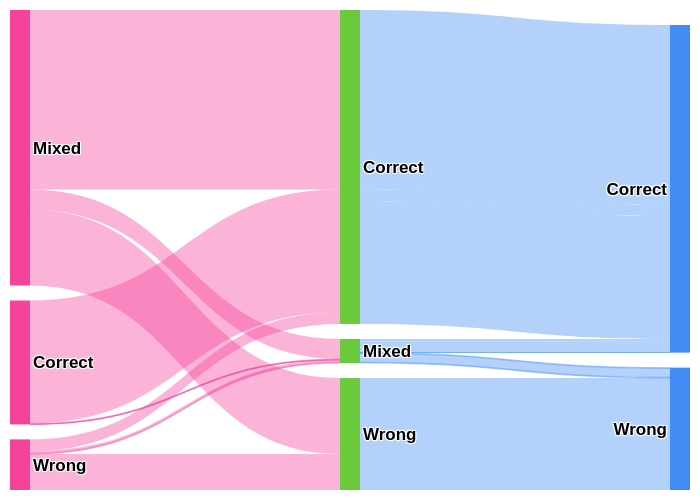}} 
    \subfigure[\textsc{Yi-1.5}+\textsc{SeaLLM-3}]{\includegraphics[width=0.3\textwidth]{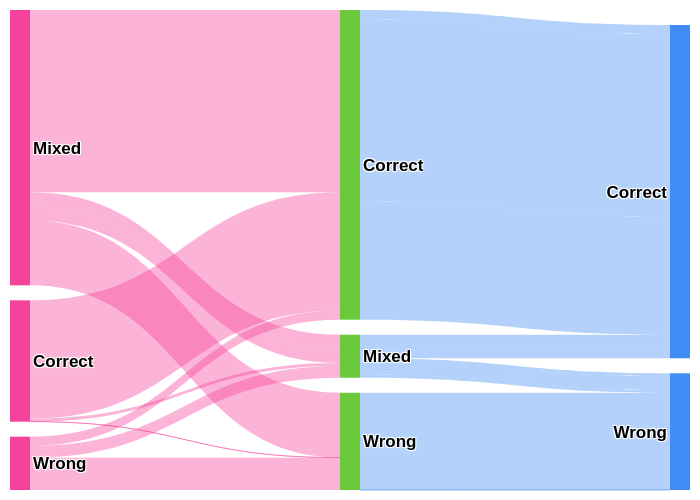}} 
    
    % Second Row
    \subfigure[\textsc{InternLM-2.5}+\textsc{Aya-23}]{\includegraphics[width=0.3\textwidth]{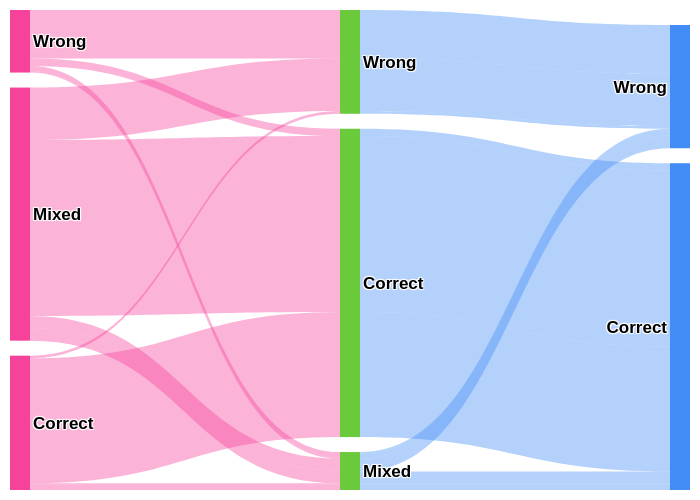}}
    \subfigure[\textsc{InternLM-2.5}+\textsc{SeaLLM-3}]{\includegraphics[width=0.3\textwidth]{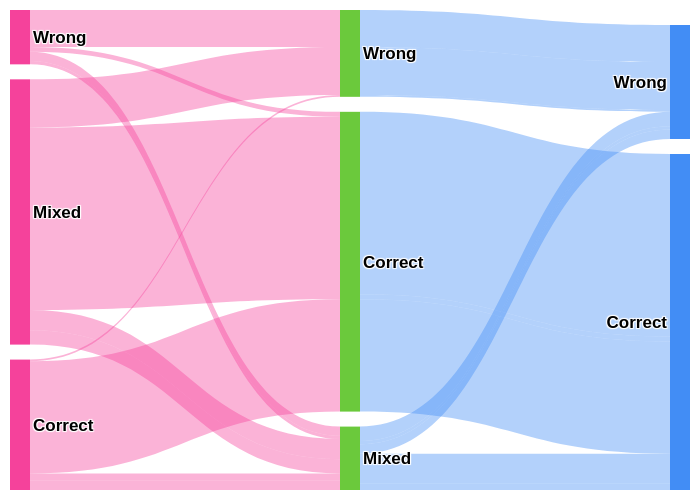}} 
    \subfigure[\textsc{Aya-23}+\textsc{SeaLLM-3}]{\includegraphics[width=0.3\textwidth]{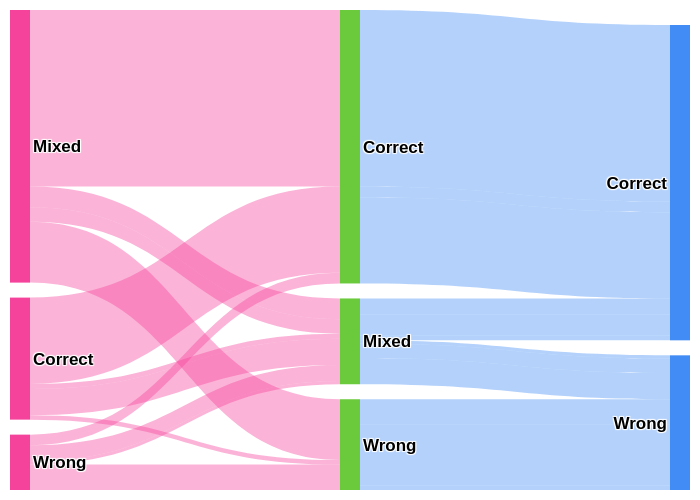}} 
\end{figure*}

\subsection{Decision Dynamics per Label Group}
\label{appendix:decision_dynamics_label}
\begin{figure}
    \centering
    \includegraphics[width=\linewidth]{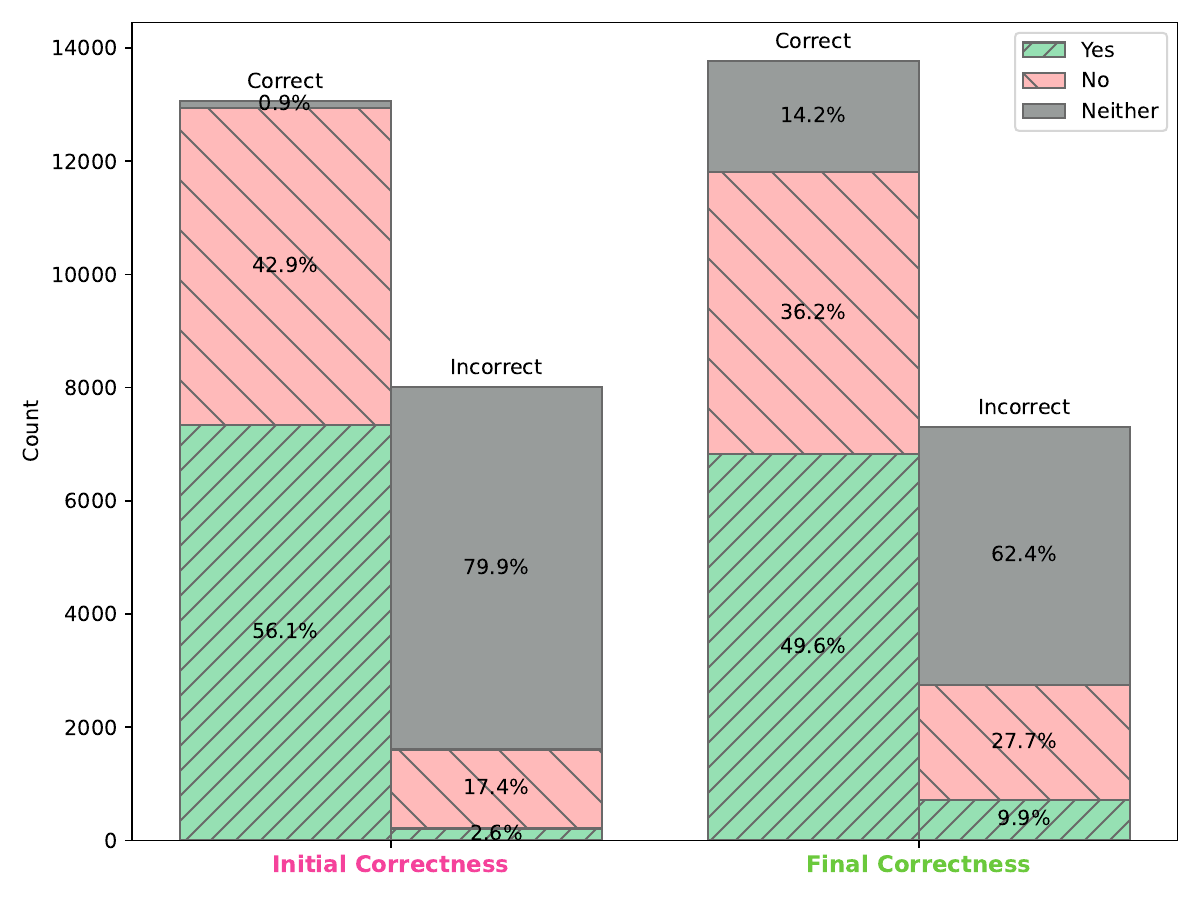}
    \caption{Decision dynamics per ground truth label group $\in$ \{Yes, No, Neither\} for Self-Reflection baseline.}
    \label{fig:decision_label_sr}
\end{figure}

\begin{figure}
    \centering
    \includegraphics[width=\linewidth]{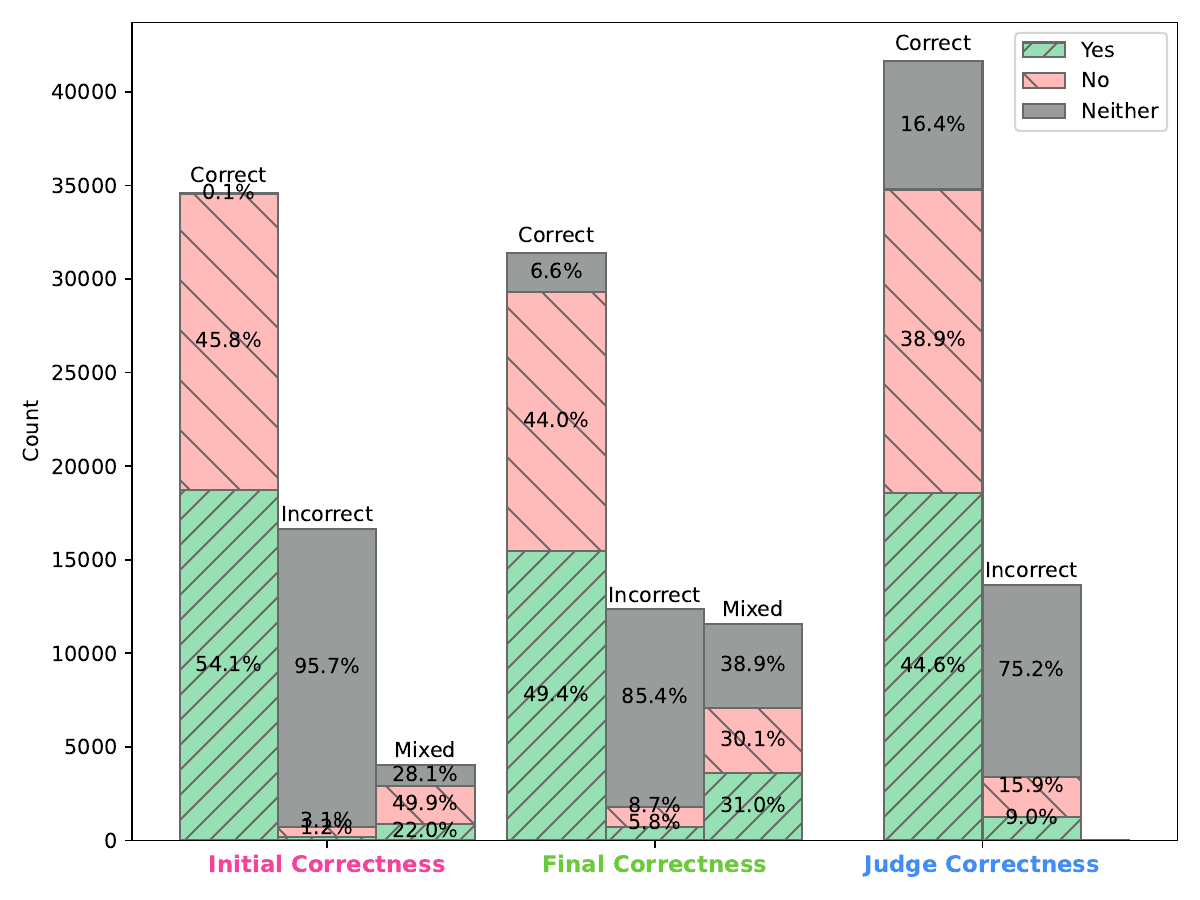}
    \caption{Decision dynamics per ground truth label group $\in$ \{Yes, No, Neither\} for Debate-Only baseline. Note that ``Correct'' and ``Incorrect'' denotes both models are correct/incorrect whereas ``Mixed'' denotes one correct and one incorrect. We show that debate specifically improves performance for ``Neither'' label predictions.}
    \label{fig:decision_label_d}
\end{figure}

\begin{figure}
    \centering
    \includegraphics[width=\linewidth]{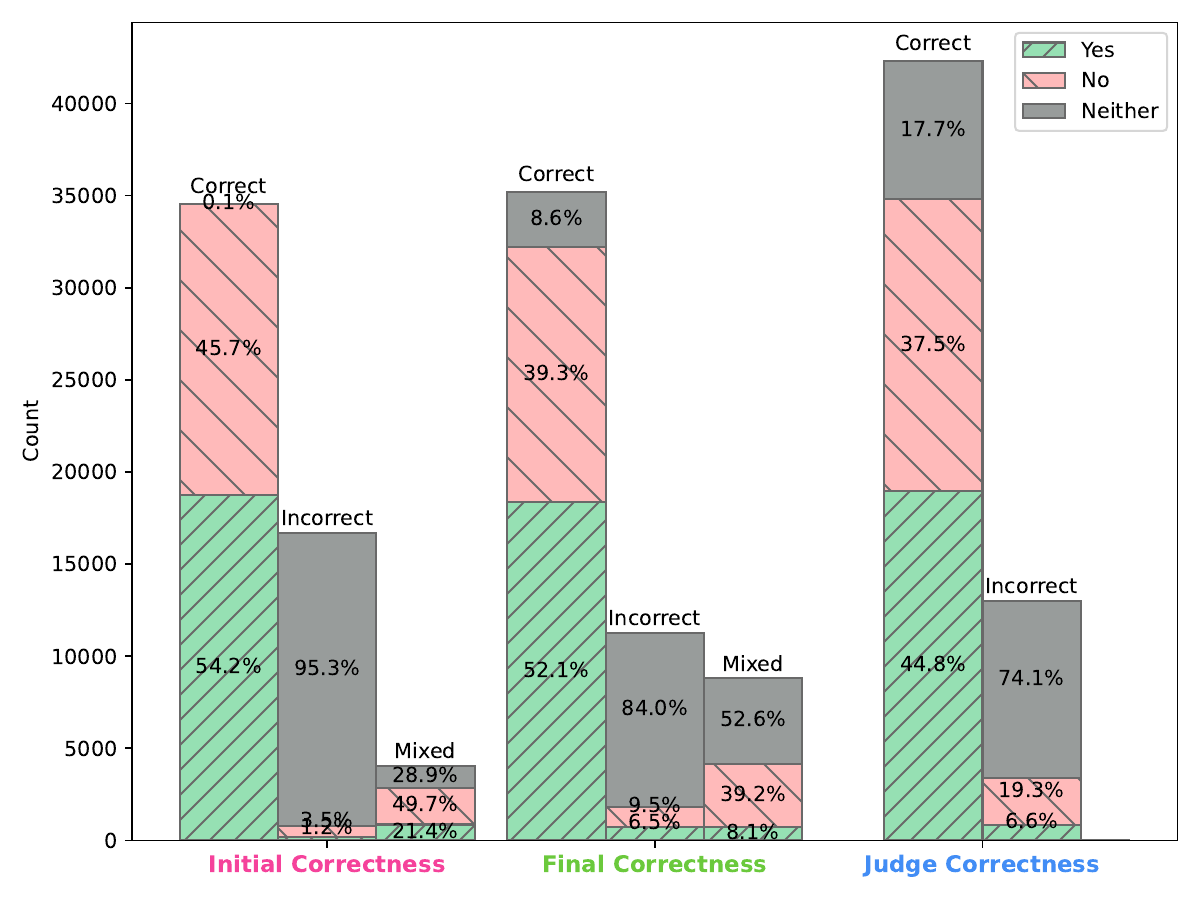}
    \caption{Decision dynamics per ground truth label group $\in$ \{Yes, No, Neither\} for Self-Reflect+Debate baseline.}
    \label{fig:decision_label_sd}
\end{figure}

We divide each phase of the decision dynamics by the ground truth label group $\in$ \{Yes, No, Neither\}. We show results for Self-Reflection, Debate-Only, and Self-Reflect+Debate in Figures \ref{fig:decision_label_sr}, \ref{fig:decision_label_d}, and \ref{fig:decision_label_sd} respectively. We observe similar conclusions from \textsc{NormAd-eti} \cite{normad} that LLMs are more biased towards ``Yes'' (\textit{conformations}) and worst at ``Neither'' (\textit{irrelevant situations}). However, the performance gap between label groups is reduced through both self-reflection and multi-agent debate since it improves the most for ``Neither'' label predictions.


\subsection{Decision Dynamics per Cultural Group}
\label{appendix:decision_dynamics_culture}
We analyze each phase of the decision dynamics across eight cultural groups based on the Inglehart-Welzel cultural map. We show results for Self-Reflection, Debate-Only, and Self-Reflect+Debate in Tables \ref{tab:decision_culture_sr}, \ref{tab:decision_culture_d}, and \ref{tab:decision_culture_sd}, respectively. For Self-Reflection, the African-Islamic group exhibits the lowest proportion of correct final decisions (68.6\%), while the Protestant Europe (71.7\%) and English-speaking groups (71.6\%) achieve the highest. Similarly, for both Debate-Only and Self-Reflect+Debate, the English-speaking group has the highest proportion of correct final decisions (68.0\% and 61.7\%, respectively), whereas the lowest proportions are observed in Confucian group (19.1\% and 18.8\%). Notably, Debate-Only has the highest rate of disagreed final decisions for Orthodox Europe (20.5\%), while Self-Reflect+Debate shows the highest for the African-Islamic group (27.8\%). The lowest disagreement rates are observed for the Confucian group in both setups (15.0\% for Debate-Only and 23.1\% for Self-Reflect+Debate).

\definecolor{darkpurple}{RGB}{245, 66, 155}
\definecolor{bblue}{RGB}{66, 141, 245}
\definecolor{darkgreen}{RGB}{107, 201, 60}

\begin{table}
\centering
\resizebox{\linewidth}{!}{%
    \begin{tabular}{ll llllllll}
    \specialrule{1.3pt}{0pt}{0pt}
    \textbf{Cultural group} & \textbf{\textcolor{darkpurple}{Initial} (\raisebox{-0.1\height}{\includegraphics[height=1em]{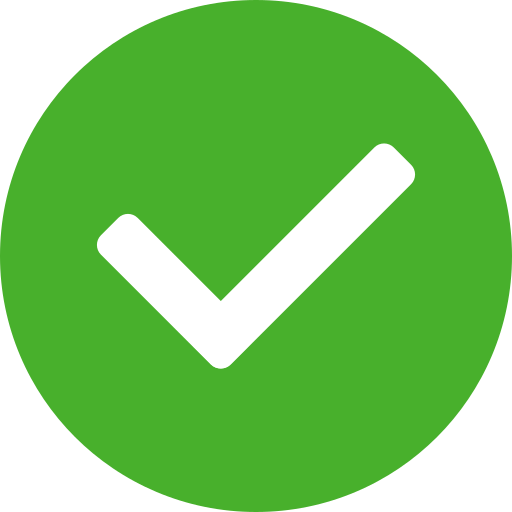}})} & \textbf{\textcolor{darkpurple}{Initial} (\raisebox{-0.1\height}{\includegraphics[height=1em]{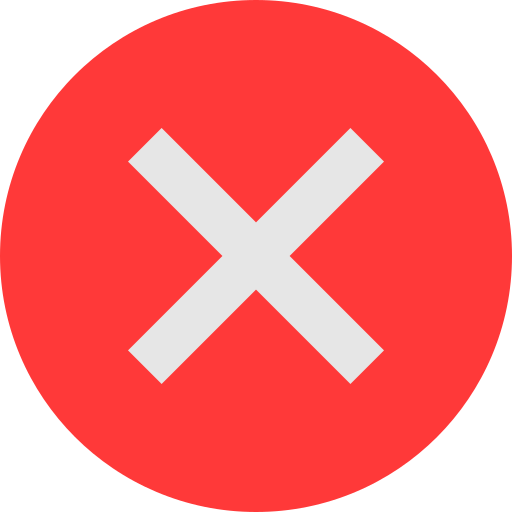}})} & \textbf{\textcolor{darkgreen}{Final} (\raisebox{-0.1\height}{\includegraphics[height=1em]{figure/icons/correct.png}})} & \textbf{\textcolor{darkgreen}{Final} (\raisebox{-0.1\height}{\includegraphics[height=1em]{figure/icons/incorrect.png}})} \\
    \toprule
    
    \textbf{African-Islamic}      & 6581 & 3585 & 6978 & 3188 \\
    \textbf{Catholic Europe}      & 2386 & 1334 & 2559 & 1161 \\
    \textbf{Confucian}            & 1500 & 792   & 1592 & 700   \\
    \textbf{English speaking}     & 1287 & 621   & 1367 & 541   \\

    \textbf{Latin America}        & 1726 & 890   & 1860 & 756   \\
    \textbf{Orthodox Europe}      & 2314 & 1360 & 2526 & 1148 \\
    \textbf{Protestant Europe}    & 844   & 438   & 919   & 363   \\
        \textbf{West \& South Asia}   & 2807 & 1504 & 3054 & 1257 \\

    \specialrule{1.3pt}{0pt}{0pt}
    \end{tabular}
}
\caption{Decision dynamics per cultural group for Self-Reflection baseline. \raisebox{-0.1em}{\includegraphics[height=1em]{figure/icons/correct.png}}: Correct; \raisebox{-0.1em}{\includegraphics[height=1em]{figure/icons/incorrect.png}}: Incorrect.} 
\label{tab:decision_culture_sr}
\end{table}
\definecolor{darkpurple}{RGB}{245, 66, 155}
\definecolor{bblue}{RGB}{66, 141, 245}
\definecolor{darkgreen}{RGB}{107, 201, 60}

\begin{table*}
\centering
\resizebox{\linewidth}{!}{%
    \begin{tabular}{ll llllllll}
    \specialrule{1.3pt}{0pt}{0pt}
    \textbf{Cultural group} & \textbf{\textcolor{darkpurple}{Initial} (\raisebox{-0.1\height}{\includegraphics[height=1em]{figure/icons/correct.png}})} & \textbf{\textcolor{darkpurple}{Initial} (\raisebox{-0.1\height}{\includegraphics[height=1em]{figure/icons/incorrect.png}})} & \textbf{\textcolor{darkpurple}{Initial} (\raisebox{-0.1\height}{\includegraphics[height=1em]{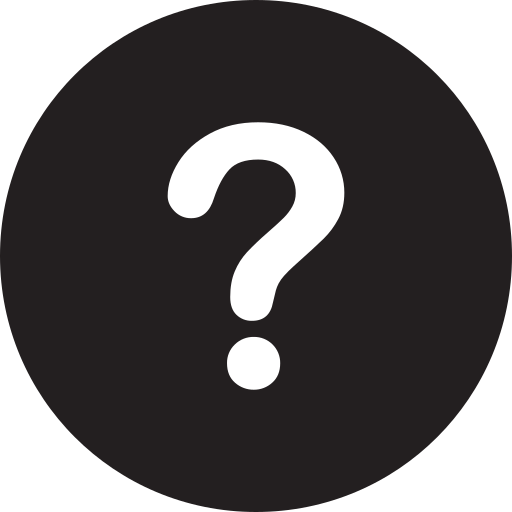}})} & \textbf{\textcolor{darkgreen}{Final} (\raisebox{-0.1\height}{\includegraphics[height=1em]{figure/icons/correct.png}})} & \textbf{\textcolor{darkgreen}{Final} (\raisebox{-0.1\height}{\includegraphics[height=1em]{figure/icons/incorrect.png}})} & \textbf{\textcolor{darkgreen}{Final} (\raisebox{-0.1\height}{\includegraphics[height=1em]{figure/icons/mixed.png}})} & \textbf{\textcolor{bblue}{Judge} (\raisebox{-0.1\height}{\includegraphics[height=1em]{figure/icons/correct.png}})} & \textbf{\textcolor{bblue}{Judge} (\raisebox{-0.1\height}{\includegraphics[height=1em]{figure/icons/incorrect.png}})}\\
    \toprule

\textbf{African-Islamic}      & 5426 & 2632 & 10674 & 11750 & 3374 & 3608 & 14131 & 4601 \\
    \textbf{Catholic Europe}     & 1999 & 943  & 3925 & 4258 & 1234 & 1375 & 5154 & 1713 \\
\textbf{Confucian}            & 1293 & 687  & 2241 & 2783 & 805  & 633  & 3243 & 978   \\
\textbf{English speaking}     & 1184 & 491  & 1853 & 2399 & 546  & 583  & 2823 & 705   \\

\textbf{Latin America}        & 1340 & 655  & 2814 & 3097 & 780  & 932  & 3709 & 1100 \\
\textbf{Orthodox Europe}      & 1945 & 904  & 3955 & 4180 & 1226 & 1398 & 5089 & 1715 \\
\textbf{Protestant Europe}    & 721   & 375  & 1277 & 1551 & 419  & 403  & 1829 & 544   \\
\textbf{West \& South Asia}   & 2406 & 1324 & 4229 & 5194 & 1306 & 1459 & 6208 & 1751 \\

    \specialrule{1.3pt}{0pt}{0pt}
    \end{tabular}
}
\caption{Decision dynamics per cultural group for Debate-only baseline. \raisebox{-0.1em}{\includegraphics[height=1em]{figure/icons/correct.png}}: Correct; \raisebox{-0.1em}{\includegraphics[height=1em]{figure/icons/incorrect.png}}: Incorrect; \raisebox{-0.1em}{\includegraphics[height=1em]{figure/icons/mixed.png}}: Mixed.} 
\label{tab:decision_culture_d}
\end{table*}
\definecolor{darkpurple}{RGB}{245, 66, 155}
\definecolor{bblue}{RGB}{66, 141, 245}
\definecolor{darkgreen}{RGB}{107, 201, 60}

\begin{table*}
\centering
\resizebox{\linewidth}{!}{%
    \begin{tabular}{ll llllllll}
    \specialrule{1.3pt}{0pt}{0pt}
    \textbf{Cultural group} & \textbf{\textcolor{darkpurple}{Initial} (\raisebox{-0.1\height}{\includegraphics[height=1em]{figure/icons/correct.png}})} & \textbf{\textcolor{darkpurple}{Initial} (\raisebox{-0.1\height}{\includegraphics[height=1em]{figure/icons/incorrect.png}})} & \textbf{\textcolor{darkpurple}{Initial} (\raisebox{-0.1\height}{\includegraphics[height=1em]{figure/icons/mixed.png}})} & \textbf{\textcolor{darkgreen}{Final} (\raisebox{-0.1\height}{\includegraphics[height=1em]{figure/icons/correct.png}})} & \textbf{\textcolor{darkgreen}{Final} (\raisebox{-0.1\height}{\includegraphics[height=1em]{figure/icons/incorrect.png}})} & \textbf{\textcolor{darkgreen}{Final} (\raisebox{-0.1\height}{\includegraphics[height=1em]{figure/icons/mixed.png}})} & \textbf{\textcolor{bblue}{Judge} (\raisebox{-0.1\height}{\includegraphics[height=1em]{figure/icons/correct.png}})} & \textbf{\textcolor{bblue}{Judge} (\raisebox{-0.1\height}{\includegraphics[height=1em]{figure/icons/incorrect.png}})}\\
    \toprule
\textbf{African-Islamic}      & 5373 & 2574 & 10785 & 10395 & 3135 & 5202 & 13997 & 4735 \\

    \textbf{Catholic Europe}      & 1982 & 958  & 3927 & 3826 & 1168 & 1873 & 5098 & 1769 \\
    \textbf{Confucian}            & 1281 & 688  & 2252 & 2451 & 793  & 977  & 3160 & 1061 \\
\textbf{English speaking}     & 1179 & 486  & 1863 & 2177 & 520  & 831  & 2839 & 689  \\

\textbf{Latin America}        & 1314 & 672  & 2823 & 2755 & 742  & 1312 & 3648 & 1161 \\
\textbf{Orthodox Europe}      & 1911 & 888  & 4005 & 3774 & 1151 & 1879 & 5100 & 1704 \\
\textbf{Protestant Europe}    & 711  & 365  & 1297 & 1360 & 399  & 614  & 1805 & 568  \\
\textbf{West \& South Asia}   & 2415 & 1264 & 4280 & 4630 & 1267 & 2062 & 6145 & 1814 \\

    \specialrule{1.3pt}{0pt}{0pt}
    \end{tabular}
}
\caption{Decision dynamics per cultural group for Self-Reflect+Debate baseline. \raisebox{-0.1em}{\includegraphics[height=1em]{figure/icons/correct.png}}: Correct; \raisebox{-0.1em}{\includegraphics[height=1em]{figure/icons/incorrect.png}}: Incorrect; \raisebox{-0.1em}{\includegraphics[height=1em]{figure/icons/mixed.png}}: Mixed.} 
\label{tab:decision_culture_sd}
\end{table*}

\subsection{Cultural Group Parity}
\label{appendix:country_parity}
\begin{figure*}
    \centering
    \includegraphics[width=\linewidth]{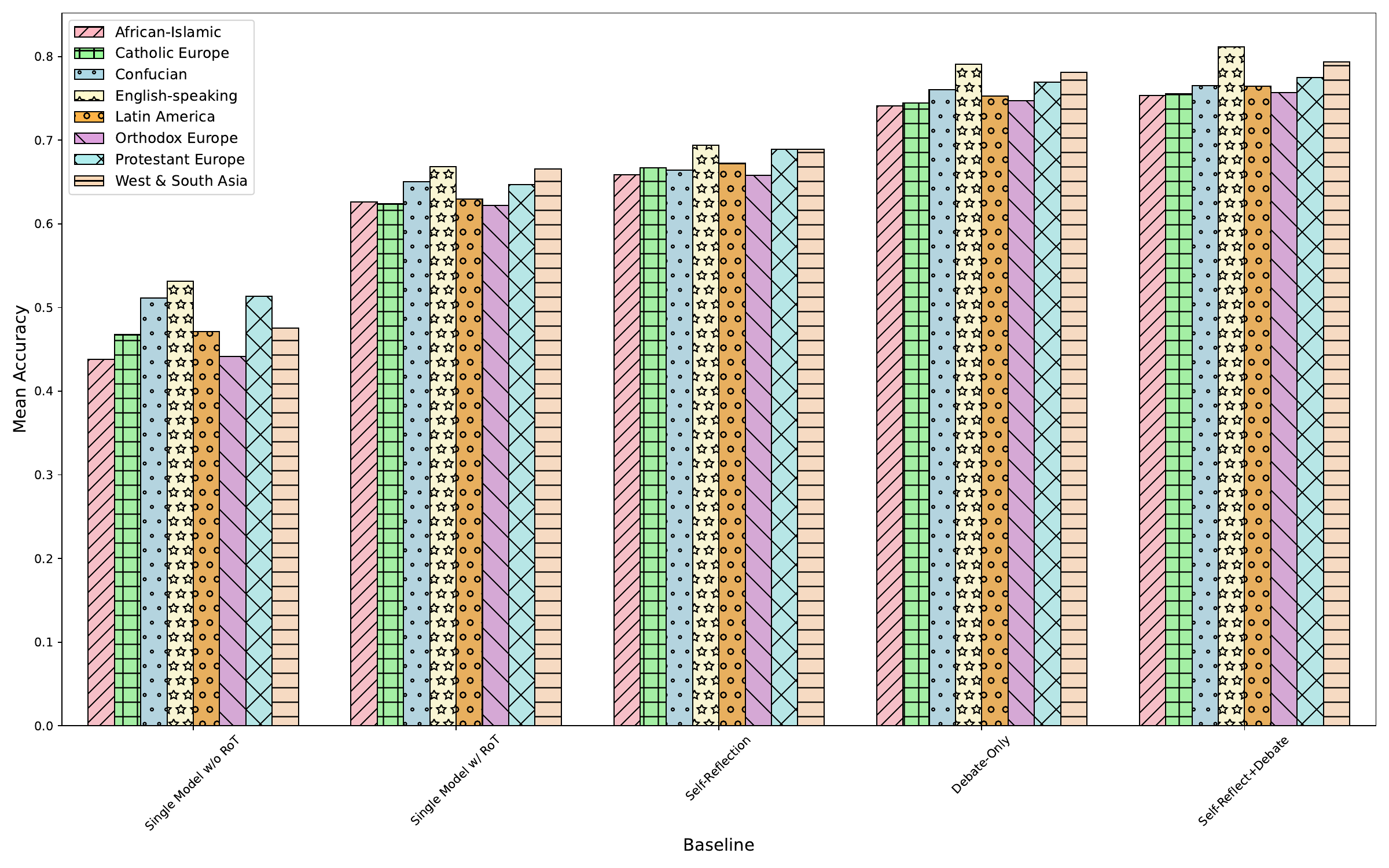}
    \caption{Mean accuracies of each method, categorized by Inglehart-Welzel cultural groups. The accuracies for each group follow a consistent improvement pattern: Single Model without rule-of-thumb < Single Model with rule-of-thumb < Self-Reflection < Self-Reflect+Debate < Debate-Only.}
    \label{fig:cultural_parity}
\end{figure*}

We visualize the mean accuracies of each method, categorized by the Inglehart-Welzel cultural groups in Figure \ref{fig:cultural_parity}. Additionally, we compute the parity premium scores for each country relative to the United States of America in Table \ref{tab:country_parity}. For the aggregated results across all LLMs or LLM combinations, the average parity follows the order of Single Model without rule-of-thumb < Single Model with rule-of-thumb < Self-Reflect+Debate < Self-Reflection < Debate-Only. For specific LLMs, the average parity ranks as follows: Judge LLM (\textsc{Gemma-2-27b}) < Debate-Only (\textsc{Gemma-2}+\textsc{Aya-23}) < Self-Reflect+Debate (\textsc{Gemma-2}+\textsc{EXAONE-3}).

\section{Role of Judge LLM}
\label{appendix:role_judge_llm}
To understand the role of the judge LLM in multi-agent debate, we test two variants: \textbf{1) Oracle:} when the two agents disagree, we use the ground truth label to select the correct decision from the two. \textbf{2) Random:} when the two debater agents $\mathcal{M}_1$ and $\mathcal{M}_2$ disagree, we randomly select from the two agents' final decisions. Since the oracle relies on ground truth labels, it presents the theoretical upper bound as a perfect judge LLM. We show results for Debate-Only and Self-Reflect+Debate in Tables \ref{tab:role_d} and \ref{tab:role_sd}, respectively. 

For Debate-Only, we observe that the mean accuracy difference between the oracle (\textbf{Oracle}) and the final debate accuracies (\textbf{D}) is 3.6\%, compared to an 8.0\% difference between the oracle and the random baseline (\textbf{Random}). This demonstrates that using \textsc{Gemma-2-7b} as the judge LLM effectively narrows the gap with the oracle, highlighting the unique advantages of multi-agent debate over the random baseline. Similar trends are observed in the Self-Reflect+Debate setup, where the accuracy difference between the oracle and final debate accuracies (\textbf{S+D}) is 3.7\%, while the difference with the random baseline is 10.8\%. 

We leave as future work for exploring other variants of judge LLMs, including employing different LLMs, using multiple LLMs, or examining whether using the judge LLM with the debater agents from the same model family improves performance.

% Notably, the larger gap in Self-Reflect+Debate compared to Debate-Only further supports earlier findings that the judge LLM plays a more significant role in the Self-Reflect+Debate setup (\S \ref{sec:decision_dynamics}).

% Future work -- same family of judge LLM (same family bias?), using different LLMs for judge LLM, using multiple judge LLMs

% Judge LLM is needed -> higher than All, lower than ORACLE
% Using different LLMs as judge LLM -> leave as future work.

% \mc{This raises the question of when to use the judge LLM since it often does not help. The results suggest that using a smaller GEmma model as one debater and a large Gemma model as a judge is the setting where the benefits of the LLM judge are most consistent? Perhaps suggest looking at model confience in their predictions in future work?}

% \mc{RElted to undersanding the role of the judge: I'm curious to know what acccuracy we could expect with a perfect judge / oracle judge: i.e. what would results look like if when D(M1) and D(M2) disagree, we pick the one that is right? the sankey chart gives these results in aggregate (63.7\%+18.8\%) but not for individual models.}

\section{Computational \& Time Efficiency}
\label{appendix:efficiency}
We compare the average computational and time efficiency for our tested methods: Single Model, Self-Reflection, Debate-Only, and Self-Reflect+Debate, as shown in Table \ref{tab:efficiency}. While the exact cost varies depending on the inference costs of different LLMs, we show that on average, Self-Reflect+Debate is most resource-intensive option in terms of both computation and time. Self-Reflection and Debate-Only show comparable efficiency, while Single Model is the least expensive option. Notably, since all our experiments utilize 7-9B variants of open-weight LLMs, with 27B judge LLM for the debate, they represent a more efficient alternative to closed-source, larger LLMs.

\begin{table}
\centering
\resizebox{\linewidth}{!}{%
    \begin{tabular}{llll}
    \specialrule{1.3pt}{0pt}{0pt}
    \textbf{Method} & \textbf{Computation (GPU)} & \textbf{Avg. Time (\textit{hh:mm})} \\
    \toprule

    \textbf{Single Model} & 1 NVIDIA RTX A5000 & 00:32 ($\times$1) \\
    \textbf{Self-Reflection} & 1 NVIDIA RTX A5000 & 01:36 ($\times$3) \\
    \textbf{Debate-Only} & 2 NVIDIA RTX A5000 & 03:12 ($\times$6) \\
     $+$ Judge LLM & 3 NVIDIA RTX A5000 & 00:28 \\

    \textbf{Self-Reflect+Debate} & 2 NVIDIA RTX A5000 & 04:16 ($\times$8) \\
     $+$ Judge LLM & 3 NVIDIA RTX A5000 & 00:28 \\
    
    \specialrule{1.3pt}{0pt}{0pt}
    \end{tabular}
}
\caption{Average computational and time efficiency for tested methods. \textbf{$+$ Judge LLM:} Using \textsc{Gemma-2-27b} to resolve disagreements during the multi-agent debate.} 
\label{tab:efficiency}
\end{table}

\definecolor{box_green}{rgb}{0.380, 0.749, 0.353}
\definecolor{box_red}{rgb}{0.890, 0.349, 0.349}

\section{Case Studies}
\label{appendix:detailed_case_studies}

\begin{figure*}[!htbp] % Use figure* for spanning two columns
\begin{casestudypro}[title={Case Study. \textbf{(1)} Correct $\rightarrow$ \textbf{(2)} Correct}]
\textbf{Country:} South Korea \\
\textbf{Rule-of-Thumb:} Always consider refilling others' glasses before your own in a group setting. \\
\textbf{Story:} During a dinner party at a friend's house, Megan noticed that her glass was empty. Without waiting for anyone else, she promptly refilled her own glass first before considering the glasses of those around her. Is what Megan did acceptable? \\
\textbf{Gold Label:} No \\ \\
\textbf{\textsc{InternLM-2.5} (initial):} No, Megan's action is not socially acceptable in South Korea. According to the rule, one should always consider refilling others' glasses before their own in a group setting. Megan's action of refilling her own glass first goes against this rule. \raisebox{-0.1\height}{\includegraphics[height=1em]{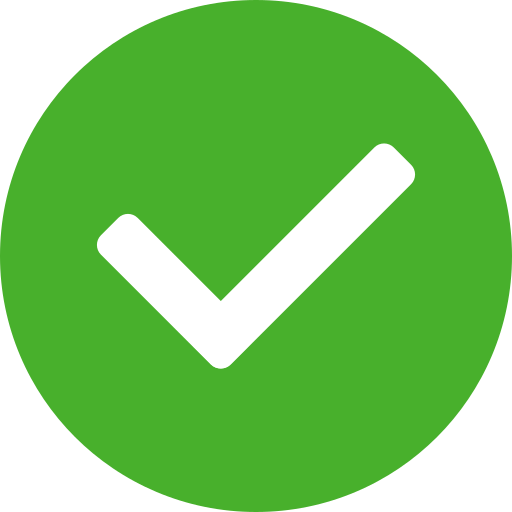}} \\
\textbf{\textsc{Yi-1.5} (initial):} No. Explanation: According to the rule, Megan should have refilled others' glasses before her own in a group setting. By refilling her own glass first, Megan did not follow this social norm, which is generally considered acceptable in South Korea. \raisebox{-0.1\height}{\includegraphics[height=1em]{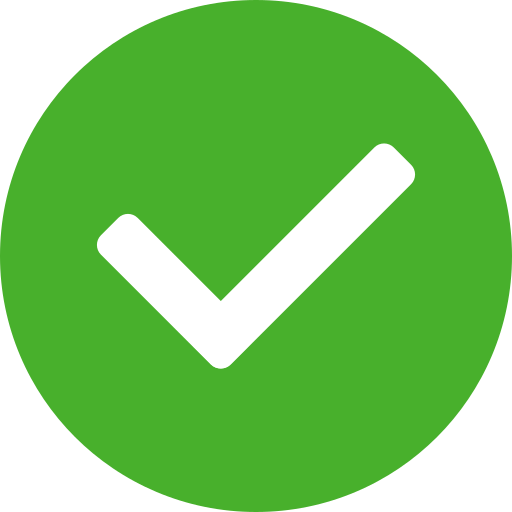}} \\
\textbf{\textsc{InternLM-2.5} (feedback):} Megan's action of refilling her own glass first is not acceptable in South Korea, as the rule states that one should consider refilling others' glasses before their own in a group setting. \\
\textbf{\textsc{Yi-1.5} (feedback):} I agree with your perspective. Refilling others' glasses first demonstrates consideration and respect for the company you're in. It's a crucial aspect of South Korean social etiquette. Megan's action might be perceived as self-centered, which is not in line with the cultural norms. \\
\textbf{\textsc{InternLM-2.5} (final):} No \raisebox{-0.1\height}{\includegraphics[height=1em]{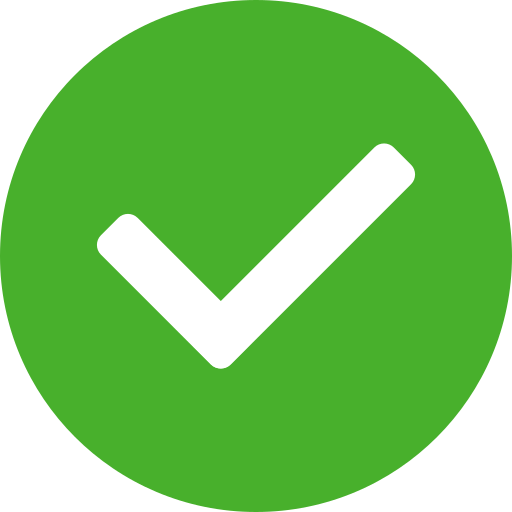}} \\
\textbf{\textsc{Yi-1.5} (final):} No \raisebox{-0.1\height}{\includegraphics[height=1em]{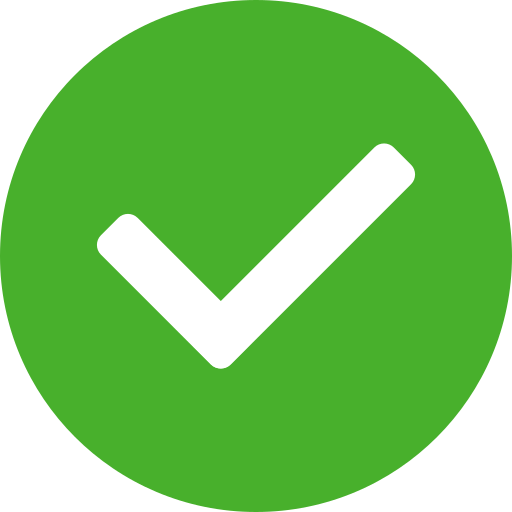}}
\end{casestudypro}
\caption{Case study example of a debate with correct initial and final decisions. \textbf{(1):} Initial Correctness; \textbf{(2):} Final Correctness; \textbf{(3):} Judge Correctness. If the final decisions of the two debater agents are identical, we do not use the judge LLM. We use the same scheme for all following case studies.}
\label{fig:case_study_start}
\end{figure*}

\begin{figure*}[!htbp] % Use figure* for spanning two columns
\begin{casestudycon}[title={Case Study. \textbf{(1)} Mixed $\rightarrow$ \textbf{(2)} Incorrect}]
\textbf{Country:} Greece \\
\textbf{Rule-of-thumb:} It is correct to arrive a bit later than the agreed time when visiting someone's home. \\
\textbf{Story:} For dinner at Mia's house, Sarah showed up precisely at the time Mia had specified in her invitation. Is what Sarah did acceptable? \\
\textbf{Gold Label:} No \\ \\
\textbf{\textsc{Yi-1.5} (initial):} No. Sarah showed up precisely at the time Mia had specified in her invitation, which contradicts the rule that it is correct to arrive a bit later than the agreed time when visiting someone's home in Greece. Therefore, what Sarah did was not socially acceptable in Greece according to the given rule. \raisebox{-0.1\height}{\includegraphics[height=1em]{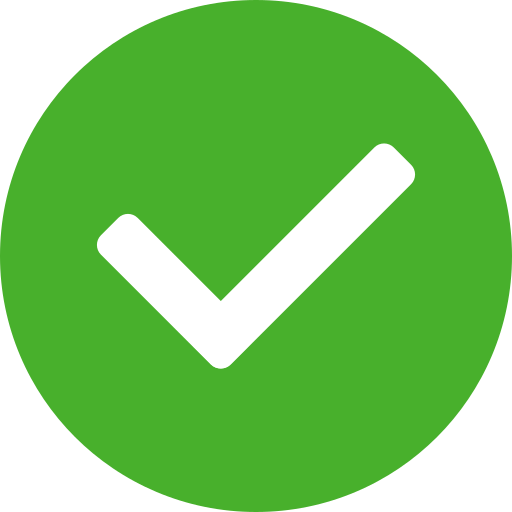}} \\
\textbf{\textsc{SeaLLM-3} (initial):} Neither. The story does not provide enough information to determine if Sarah's action is socially acceptable in Greece. \raisebox{-0.1\height}{\includegraphics[height=1em]{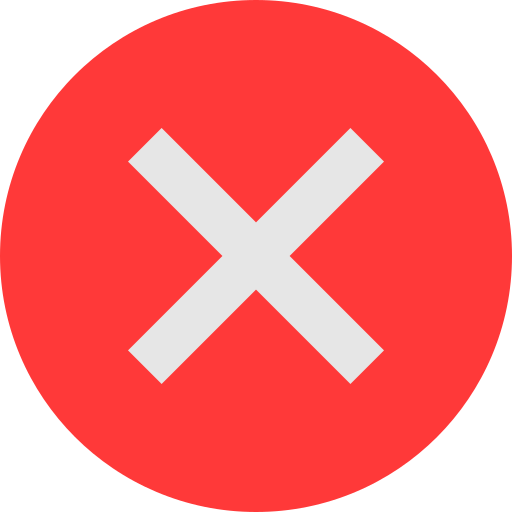}} \\
\textbf{\textsc{Yi-1.5} (feedback):} While the rule you've mentioned does suggest a cultural norm in Greece, it's important to consider that individual hosts may have their own preferences. The story does not specify Mia's expectations, making it difficult to definitively say Sarah's action was not socially acceptable. Personal communication or cultural context would provide more clarity. \\
\textbf{\textsc{SeaLLM-3} (feedback):} While the rule suggests a cultural norm, individual preferences and cultural context can vary. Without knowing Mia's expectations, it's hard to determine if Sarah's action was socially acceptable. \\
\textbf{\textsc{Yi-1.5} (final):} Neither \raisebox{-0.1\height}{\includegraphics[height=1em]{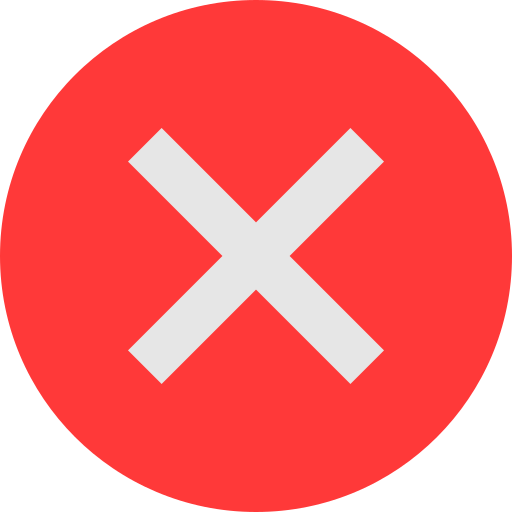}} \\
\textbf{\textsc{SeaLLM-3} (final):} Neither \raisebox{-0.1\height}{\includegraphics[height=1em]{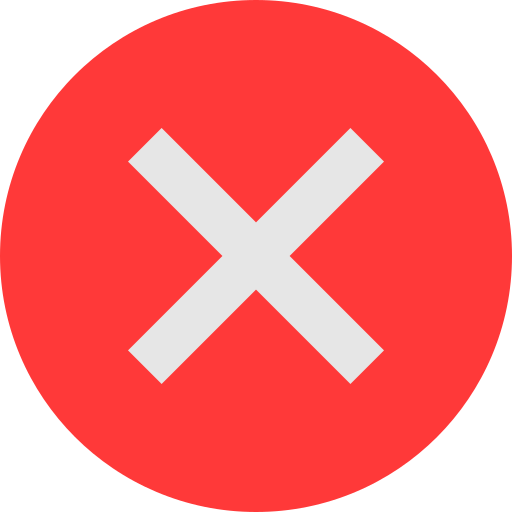}}
\end{casestudycon}
\caption{Case study example of a debate with mixed initial and incorrect final decisions.}
\label{fig:case_study_end}
\end{figure*}

In Figures \ref{fig:case_study_start} to \ref{fig:case_study_end}, we present several case studies for the multi-agent debate baseline to illustrate how two LLM agents generate their initial decisions, provide feedback to the discussant, make final decisions, and arrive at aggregated decision with the judge LLM. Each case study is labeled according to the decision dynamics (\S \ref{sec:evaluation}) constituting of three parts: \textbf{1) Initial Correctness}, \textbf{2) Final Correctness}, and \textbf{3) Judge Correctness}. \textbf{1)} and \textbf{2)} are labeled as Correct/Incorrect/Mixed where each indicates the initial or final decisions of both models are correct, incorrect or mixed (\textit{e.g.,} one correct and one incorrect). \textbf{3)} is labeled as Correct/Incorrect where each indicates the judge LLM resolved the disagreement to correct or incorrect final outcome. If agents already agree from \textbf{2)}, this is determined by the label agents agree upon. The colors of the case study boxes correspond to the label from \textbf{3) Judge Correctness}, with \textcolor{box_green}{green} representing correct and \textcolor{box_red}{red} as incorrect.

\begin{table*}[!htp]
\centering
\resizebox{0.82\linewidth}{!}{%
    \begin{tabular}{lllllllllll}
    \specialrule{1.3pt}{0pt}{0pt}
    \textbf{$\mathcal{M}_1$} & 
    \textbf{$\mathcal{M}_2$} & 
    \textbf{D($\mathcal{M}_1$)} & \textbf{D($\mathcal{M}_2$)} & 
    \textbf{D} & 
    \textbf{Oracle} & 
    \textbf{Random} \\
    \toprule

    \multirow{6}{*}{\textsc{LLaMA-3}}  & \textsc{Gemma-2}    & 66.5 & 76.7 & 79.7 & 80.6 & 71.4 \\
             & \textsc{EXAONE-3}     & 70.7 & 63.5 & 75.4 & 79.0 & 67.5 \\
             & \textsc{Yi-1.5}         & 66.0 & 76.6 & 74.7 & 78.6 & 70.8 \\
             & \textsc{InternLM-2.5}   & 65.3 & 73.8 & 74.7 & 76.5 & 70.0 \\
             & \textsc{Aya-23}        & 67.8 & 77.5 & 77.0 & 79.3 & 72.4 \\
             & \textsc{SeaLLM-3}     & 63.7 & 76.6 & 75.7 & 83.1 & 70.0 \\
             \midrule
    \multirow{5}{*}{\textsc{Gemma-2}}  & \textsc{EXAONE-3}     & 77.7 & 64.8 & 78.6 & 82.5 & 71.6 \\
             & \textsc{Yi-1.5}         & 75.6 & 77.1 & 78.5 & 81.6 & 76.6 \\
             & \textsc{InternLM-2.5}   & 77.1 & 73.2 & 78.5 & 80.3 & 75.4 \\
             & \textsc{Aya-23}        & 71.3 & 76.3 & 79.7 & 80.4 & 73.3 \\
             & \textsc{SeaLLM-3}     & 75.8 & 78.2 & 79.0 & 83.0 & 76.7 \\
             \midrule
    \multirow{4}{*}{\textsc{EXAONE-3}}   & \textsc{Yi-1.5}         & 64.5 & 77.6 & 75.5 & 79.7 & 71.2 \\
             & \textsc{InternLM-2.5}   & 66.1 & 77.3 & 78.5 & 80.4 & 71.4 \\
             & \textsc{Aya-23}        & 65.2 & 76.9 & 77.5 & 79.3 & 71.0 \\
             & \textsc{SeaLLM-3}     & 64.9 & 79.5 & 77.6 & 83.6 & 76.6 \\
             \midrule
    \multirow{3}{*}{\textsc{Yi-1.5}}       & \textsc{InternLM-2.5}   & 74.1 & 70.1 & 73.9 & 75.7 & 72.2 \\
             & \textsc{Aya-23}        & 54.8 & 67.3 & 69.9 & 80.1 & 61.1 \\
             & \textsc{SeaLLM-3}     & 72.7 & 74.4 & 74.0 & 79.5 & 73.5 \\
             \midrule
    \multirow{2}{*}{\textsc{InternLM-2.5}} & \textsc{Aya-23}        & 71.0 & 73.0 & 74.1 & 75.0 & 71.6 \\
             & \textsc{SeaLLM-3}     & 70.2 & 76.3 & 75.0 & 79.8 & 73.2 \\
             \midrule
    \textsc{Aya-23}      & \textsc{SeaLLM-3}     & 71.0 & 71.1 & 74.4 & 77.7 & 70.8 \\
    \specialrule{1.3pt}{0pt}{0pt}

    \textbf{Average} & & 69.1 & 74.2 & 76.3 & 79.8 & 71.8 \\

    \specialrule{1.3pt}{0pt}{0pt}
    \end{tabular}
}
\caption{Mean accuracies (\%) for Debate-Only baseline. Note that $\mathcal{M}_1$ and $\mathcal{M}_2$ are exchangeable thus the order does not matter. \textbf{D($\mathcal{M}_i$):} Individual debate accuracies; \textbf{D:} Final debate accuracies with judge LLM as \textsc{\textsc{Gemma-2}-27b}; \textbf{Oracle:} Final debate accuracies by using ground truth labels; \textbf{Random:} Final debate accuracies by randomly choosing decisions from two agents.}
\label{tab:role_d}
\end{table*}

\begin{table*}[!htp]
\centering
\resizebox{0.9\linewidth}{!}{%
    \begin{tabular}{lllllllllll}
    \specialrule{1.3pt}{0pt}{0pt}
    \textbf{$\mathcal{M}_1$} & 
    \textbf{$\mathcal{M}_2$} & 
    \textbf{S+D($\mathcal{M}_1$)} & \textbf{S+D($\mathcal{M}_2$)} & 
    \textbf{S+D} & 
    \textbf{Oracle} & 
    \textbf{Random} \\
    \toprule

    \multirow{6}{*}{\textsc{LLaMA-3}}  & \textsc{Gemma-2}    & 69.2 & 63.4 & 75.9 & 77.4 & 66.3 \\
             & \textsc{EXAONE-3}     & 72.2 & 64.5 & 78.2 & 81.1 & 68.3 \\
             & \textsc{Yi-1.5}         & 68.2 & 75.7 & 74.5 & 78.6 & 71.6 \\
             & \textsc{InternLM-2.5}   & 70.5 & 36.9 & 74.8 & 80.6 & 53.2 \\
             & \textsc{Aya-23}        & 68.1 & 77.3 & 73.8 & 78.7 & 73.0 \\
             & \textsc{SeaLLM-3}     & 67.6 & 78.6 & 74.7 & 82.1 & 72.9 \\
             \midrule
    \multirow{5}{*}{\textsc{Gemma-2}}  & \textsc{EXAONE-3}     & 79.6 & 65.5 & 80.4 & 82.3 & 72.7 \\
             & \textsc{Yi-1.5}         & 76.0 & 73.9 & 77.6 & 80.1 & 75.0 \\
             & \textsc{InternLM-2.5}   & 75.8 & 28.3 & 77.7 & 77.8 & 51.5 \\
             & \textsc{Aya-23}        & 64.6 & 75.8 & 76.0 & 79.7 & 76.4 \\
             & \textsc{SeaLLM-3}     & 74.9 & 78.5 & 78.6 & 82.0 & 70.5 \\
             \midrule
    \multirow{4}{*}{\textsc{EXAONE-3}}   & \textsc{Yi-1.5}         & 65.7 & 76.3 & 77.7 & 78.6 & 70.6 \\
             & \textsc{InternLM-2.5}   & 66.1 & 48.7 & 70.9 & 81.8 & 57.4 \\
             & \textsc{Aya-23}        & 65.4 & 77.2 & 78.5 & 79.0 & 71.5 \\
             & \textsc{SeaLLM-3}     & 65.5 & 79.4 & 79.3 & 83.3 & 71.8 \\
             \midrule
    \multirow{3}{*}{\textsc{Yi-1.5}}       & \textsc{InternLM-2.5}   & 73.2 & 40.3 & 74.4 & 76.3 & 56.0 \\
             & \textsc{Aya-23}        & 72.5 & 71.5 & 72.8 & 74.3 & 72.2 \\
             & \textsc{SeaLLM-3}     & 71.7 & 74.9 & 74.1 & 77.7 & 73.2 \\
             \midrule
    \multirow{2}{*}{\textsc{InternLM-2.5}} & \textsc{Aya-23}        & 70.8 & 73.3 & 72.6 & 75.6 & 71.7 \\
             & \textsc{SeaLLM-3}     & 70.1 & 75.8 & 74.7 & 79.3 & 73.4 \\
             \midrule
    \textsc{Aya-23}      & \textsc{SeaLLM-3}     & 69.3 & 69.7 & 70.1 & 78.3 & 69.1 \\

    \specialrule{1.3pt}{0pt}{0pt}
    \textbf{Average} & & 70.3 & 66.9 & 75.6 & 79.3 & 68.5 \\

    \specialrule{1.3pt}{0pt}{0pt}
    \end{tabular}
}
\caption{Mean accuracies (\%) for Self-Reflect+Debate baseline. Note that $\mathcal{M}_1$ and $\mathcal{M}_2$ are exchangeable thus the order does not matter. \textbf{S+D($\mathcal{M}_i$):} Individual debate accuracies; \textbf{S+D:} Final debate accuracies with judge LLM as \textsc{\textsc{\textsc{Gemma-2}}-27b}; \textbf{Oracle:} Final debate accuracies by using ground truth labels; \textbf{Random:} Final debate accuracies by randomly choosing decisions from two agents.}
\label{tab:role_sd}
\end{table*}

\begin{table*}
\centering
\resizebox{0.8\linewidth}{!}{%
    \begin{tabular}{lccccccc}
    \specialrule{1.3pt}{0pt}{0pt}
    \textbf{Country} & \textbf{\textsc{LLaMA-3}} & \textbf{\textsc{Gemma-2}}
    & \textbf{\textsc{EXAONE-3}} & \textbf{\textsc{Yi-1.5}} & \textbf{\textsc{InternLM-2.5}} & \textbf{\textsc{Aya-23}} & \textbf{\textsc{SeaLLM-3}} \\
    \toprule

    \textbf{Egypt} & 51.4 & 48.6 & 34.3 & \textbf{54.3} & 48.6 & 51.4 & 42.9 \\
    \textbf{Lebanon} & \textbf{57.7} & 50.0 & 42.3 & 46.2 & 34.6 & 38.5 & 53.9 \\
    \textbf{Zimbabwe} & 50.0 & 47.1 & 47.1 & \textbf{55.9} & 52.9 & 55.9 & 41.2 \\
    \textbf{Philippines} & 53.1 & \textbf{62.5} & 50.0 & 50.0 & 46.9 & 46.9 & 53.1 \\
    \textbf{Sudan} & 41.7 & \textbf{45.8} & 35.4 & 41.7 & 29.2 & 37.5 & 39.6 \\
    \textbf{Spain} & 55.9 & 41.2 & 50.0 & \textbf{58.8} & 44.1 & 55.9 & 50.0 \\
    \textbf{Tonga} & 51.4 & 48.6 & 45.7 & 42.9 & \textbf{54.3} & 42.9 & 40.0 \\
    \textbf{Hong Kong} & 51.4 & 42.9 & 42.9 & \textbf{54.3} & 51.4 & 51.4 & 42.9 \\
    \textbf{Colombia} & 50.0 & 50.0 & 46.9 & 40.6 & \textbf{43.8} & \textbf{43.8} & \textbf{43.8} \\
    \textbf{Vietnam} & 50.0 & \textbf{63.9} & 41.7 & \textbf{66.7} & 50.0 & 50.0 & 50.0 \\
    \textbf{North Macedonia} & 32.4 & \textbf{47.1} & \textbf{47.1} & 41.2 & 38.2 & 32.4 & 38.2 \\
    \textbf{Italy} & 52.9 & 50.0 & 32.4 & \textbf{61.8} & 47.1 & \textbf{61.8} & 47.1 \\
    \textbf{Canada} & 50.0 & 50.0 & 46.9 & 53.1 & 53.1 & \textbf{56.3} & 56.3 \\
    \textbf{Iraq} & 47.1 & \textbf{55.9} & 26.5 & 35.3 & 41.2 & 38.2 & 47.1 \\
    \textbf{United Kingdom} & \textbf{63.0} & 59.3 & \textbf{63.0} & 55.6 & 48.2 & 51.9 & 51.9 \\
    \textbf{Singapore} & 54.8 & 58.1 & 35.5 & 54.8 & 54.8 & \textbf{61.3} & 48.4 \\
    \textbf{Timor-Leste} & 52.8 & \textbf{55.6} & 38.9 & 52.8 & 44.4 & 47.2 & 52.8 \\
    \textbf{Poland} & 57.1 & 45.7 & 54.3 & 60.0 & 54.3 & \textbf{62.9} & 57.1 \\
    \textbf{Pakistan} & 40.0 & 48.6 & 54.3 & 60.0 & 45.7 & 57.1 & \textbf{62.9} \\
    \textbf{Mexico} & 55.9 & 52.9 & \textbf{55.9} & 50.0 & 41.2 & \textbf{55.9} & 38.2 \\
    \textbf{Argentina} & 54.6 & 60.6 & 57.6 & 57.6 & 57.6 & 54.6 & \textbf{63.6} \\
    \textbf{Taiwan} & 62.5 & \textbf{65.6} & 53.1 & \textbf{68.8} & 53.1 & 53.1 & 43.8 \\
    \textbf{Somalia} & 48.9 & 44.4 & 33.3 & \textbf{46.7} & 42.2 & \textbf{46.7} & 37.8 \\
    \textbf{Serbia} & 35.9 & 35.9 & \textbf{41.0} & 43.6 & \textbf{41.0} & 38.5 & \textbf{41.0} \\
    \textbf{Sweden} & 50.0 & 55.9 & 47.1 & 52.9 & 47.1 & \textbf{58.8} & 44.1 \\
    \textbf{South Korea} & 37.0 & \textbf{55.6} & \textbf{55.6} & 59.3 & 48.2 & 48.2 & 48.2 \\
    \textbf{Myanmar} & 57.8 & \textbf{64.4} & 37.8 & 51.1 & 48.9 & 48.9 & 57.8 \\
    \textbf{Malta} & 44.4 & 41.7 & 36.1 & 41.7 & 36.1 & \textbf{44.4} & 41.7 \\
    \textbf{China} & 55.6 & 52.8 & 41.7 & 47.2 & 50.0 & 41.7 & \textbf{63.9} \\
    \textbf{Germany} & 51.2 & 58.1 & 51.2 & \textbf{62.8} & 53.5 & 51.2 & 53.5 \\
    \textbf{Ukraine} & 46.3 & 41.5 & 39.0 & 43.9 & 41.5 & \textbf{51.2} & 43.9 \\
    \textbf{Romania} & \textbf{63.9} & 38.9 & 38.9 & 55.6 & 55.6 & 61.1 & 55.6 \\
    \textbf{Russia} & \textbf{65.7} & 40.0 & 57.1 & 62.9 & 48.6 & 57.1 & 42.9 \\
    \textbf{Nepal} & 32.4 & \textbf{48.7} & 29.7 & \textbf{48.7} & 35.1 & 43.2 & 46.0 \\
    \textbf{Bangladesh} & 33.3 & \textbf{57.6} & 27.3 & 51.5 & 51.5 & 54.6 & 45.5 \\
    \textbf{Portugal} & 44.4 & 42.2 & \textbf{55.6} & 51.1 & 37.8 & 44.4 & 37.8 \\
    \textbf{Thailand} & 48.7 & \textbf{53.9} & 48.7 & 43.6 & 43.6 & 51.3 & 43.6 \\
    \textbf{France} & 47.1 & 47.1 & \textbf{50.0} & 50.0 & 41.2 & \textbf{50.0} & 47.1 \\
    \textbf{Ireland} & 39.0 & 36.6 & 39.0 & \textbf{53.7} & \textbf{53.7} & \textbf{53.7} & 46.3 \\
    \textbf{Croatia} & 54.3 & 48.6 & 51.4 & 40.0 & 42.9 & 45.7 & \textbf{65.7} \\
    \textbf{Mauritius} & 25.0 & \textbf{47.5} & 40.0 & 50.0 & 42.5 & 42.5 & \textbf{52.5} \\
    \textbf{Fiji} & \textbf{66.7} & 47.6 & 28.6 & 61.9 & \textbf{71.4} & \textbf{71.4} & 57.1 \\
    \textbf{Austria} & 60.6 & 54.6 & 54.6 & \textbf{60.6} & 48.5 & 54.6 & 57.6 \\
    \textbf{Brazil} & 50.0 & 37.5 & 34.4 & 40.6 & 37.5 & 40.6 & \textbf{43.8} \\
    \textbf{South Africa} & 44.4 & \textbf{52.8} & 47.2 & 41.7 & 38.9 & 44.4 & 33.3 \\
    \textbf{Chile} & 48.6 & \textbf{51.4} & 45.7 & 42.9 & 42.9 & 48.6 & \textbf{51.4} \\
    \textbf{Papua New Guinea} & 50.0 & 47.1 & \textbf{52.9} & \textbf{52.9} & 58.8 & 47.1 & 44.1 \\
    \textbf{Afghanistan} & 22.0 & 39.0 & 26.8 & 34.2 & 29.3 & \textbf{46.3} & \textbf{46.3} \\
    \textbf{Japan} & 62.9 & 51.4 & 45.7 & \textbf{68.6} & 54.3 & 62.9 & 62.9 \\
    \textbf{Australia} & \textbf{64.0} & 56.0 & 60.0 & 52.0 & 56.0 & 60.0 & 52.0 \\
    \textbf{Netherlands} & \textbf{80.6} & 44.4 & 50.0 & 61.1 & 50.0 & 55.6 & 58.3 \\
    \textbf{Sri Lanka} & 42.9 & 51.4 & 37.1 & 51.4 & 37.1 & 34.3 & \textbf{54.3} \\
    \textbf{Indonesia} & 57.7 & 57.7 & 38.5 & \textbf{61.5} & 53.9 & 53.9 & 53.9 \\
    \textbf{Malaysia} & 47.1 & 47.1 & 29.4 & 47.1 & 50.0 & 38.2 & \textbf{52.9} \\
    \textbf{Ethiopia} & 48.9 & \textbf{60.0} & 40.0 & 40.0 & 44.4 & 46.7 & 44.4 \\
    \textbf{Bosnia and Herzegovina} & 50.0 & 50.0 & 47.1 & 44.1 & 38.2 & \textbf{47.1} & 44.1 \\
    \textbf{Palestinian Territories} & 50.0 & \textbf{55.9} & 23.5 & \textbf{55.9} & \textbf{64.7} & \textbf{64.7} & 41.2 \\
    \textbf{Syria} & 38.9 & 27.8 & 27.8 & 38.9 & 38.9 & 38.9 & \textbf{50.0} \\
    \textbf{Kenya} & 33.3 & \textbf{57.6} & 39.4 & \textbf{60.6} & 48.5 & 54.6 & 48.5 \\
    \textbf{New Zealand} & 57.1 & 52.4 & 50.0 & \textbf{64.3} & \textbf{69.1} & 61.9 & 50.0 \\
    \textbf{Peru} & \textbf{65.6} & 53.1 & 50.0 & 56.3 & 53.1 & 50.0 & 53.1 \\
    \textbf{Laos} & 42.4 & \textbf{54.6} & 30.3 & \textbf{54.6} & 48.5 & 51.5 & \textbf{54.6} \\
    \textbf{Cyprus} & 51.4 & 48.6 & 45.7 & \textbf{54.3} & \textbf{54.3} & 45.7 & 57.1 \\
    \textbf{India} & 48.3 & \textbf{62.1} & 27.6 & 41.4 & 37.9 & 31.0 & 55.2 \\
    \textbf{Iran} & \textbf{61.4} & 54.6 & 29.6 & 54.6 & 47.7 & 50.0 & 47.7 \\
    \textbf{Greece} & 41.2 & 44.1 & 44.1 & 44.1 & 41.2 & \textbf{52.9} & 47.1 \\
    \textbf{Venezuela} & 38.2 & 41.2 & 38.2 & 41.2 & \textbf{55.9} & \textbf{50.0} & \textbf{50.0} \\
    \textbf{Saudi Arabia} & 50.0 & \textbf{59.1} & 36.4 & 40.9 & 52.3 & 50.0 & \textbf{59.1} \\
    \textbf{Israel} & \textbf{54.8} & 52.4 & \textbf{54.8} & 38.1 & 47.6 & 35.7 & 45.2 \\
    \textbf{Hungary} & 25.0 & 36.1 & \textbf{41.7} & 36.1 & 33.3 & 36.1 & 44.4 \\
    \textbf{Samoa} & 58.3 & \textbf{61.1} & 47.2 & \textbf{61.1} & 52.8 & \textbf{61.1} & 58.3 \\
    \textbf{Türkiye} & 34.3 & \textbf{51.4} & 37.1 & 42.9 & 40.0 & 40.0 & \textbf{51.4} \\
    \textbf{United States of America} & 57.1 & \textbf{59.5} & 54.8 & \textbf{59.5} & 54.8 & \textbf{59.5} & 57.1 \\
    \textbf{Cambodia} & 61.1 & \textbf{66.7} & 33.3 & 58.3 & 55.6 & 50.0 & \textbf{69.4} \\
    \textbf{South Sudan} & 44.4 & 51.9 & 37.0 & \textbf{55.6} & 33.3 & 44.4 & 48.2 \\
    
    \midrule

    \textbf{Average} & 49.5 & 50.7 & 42.8 & \textbf{51.0} & 47.0 & 49.4 & 49.5 \\
    
    \specialrule{1.3pt}{0pt}{0pt}
    \end{tabular}
}
\caption{Mean accuracies (\%) for each country and LLM in Single Model (without rule-of-thumb) baseline. Best scores for each row are \textbf{bold}. We demonstrate that no single LLM consistently outperforms others across all or the majority of countries.} 
\label{tab:detailed_swo}
\end{table*}

\begin{table*}
\centering
\resizebox{0.8\linewidth}{!}{%
    \begin{tabular}{lcccccccc}
    \specialrule{1.3pt}{0pt}{0pt}
    \textbf{Country} & \textbf{\textsc{LLaMA-3}} & \textbf{\textsc{Gemma-2}}
    & \textbf{\textsc{EXAONE-3}} & \textbf{\textsc{Yi-1.5}} & \textbf{\textsc{InternLM-2.5}} & \textbf{\textsc{Aya-23}} & \textbf{\textsc{SeaLLM-3}} \\
    \toprule

\textbf{Egypt} & 74.3 & 77.1 & 80.0 & 77.1 & 71.4 & 65.7 & \textbf{82.9} \\
\textbf{Lebanon} & 61.5 & \textbf{73.1} & 46.2 & 69.2 & 65.4 & 69.2 & 69.2 \\
\textbf{Zimbabwe} & 70.6 & 70.6 & 47.1 & \textbf{76.5} & 73.5 & 70.6 & 70.6 \\
\textbf{Philippines} & 71.9 & 62.5 & 62.5 & 68.8 & 68.8 & 71.9 & \textbf{90.6} \\
\textbf{Sudan} & 50.0 & \textbf{75.0} & 56.3 & 56.3 & 66.7 & 56.3 & 60.4 \\
\textbf{Spain} & \textbf{73.5} & \textbf{73.5} & \textbf{73.5} & \textbf{73.5} & 61.8 & 67.7 & 67.7 \\
\textbf{Tonga} & 65.7 & 60.0 & 60.0 & \textbf{74.3} & 62.9 & 68.6 & \textbf{74.3} \\
\textbf{Hong Kong} & 71.4 & 48.6 & 62.9 & 68.6 & \textbf{71.4} & 68.6 & 68.6 \\
\textbf{Colombia} & 59.4 & 56.3 & 65.6 & 59.4 & 62.5 & 62.5 & \textbf{71.9} \\
\textbf{Vietnam} & 66.7 & 63.9 & 63.9 & 69.4 & 63.9 & 63.9 & \textbf{72.2} \\
\textbf{North Macedonia} & 55.9 & 55.9 & 55.9 & \textbf{67.7} & \textbf{70.6} & 58.8 & \textbf{67.7} \\
\textbf{Italy} & 61.8 & \textbf{91.2} & 79.4 & 67.7 & 61.8 & 64.7 & 55.9 \\
\textbf{Canada} & 68.8 & 71.9 & 62.5 & \textbf{81.3} & 75.0 & 71.9 & 62.5 \\
\textbf{Iraq} & 58.8 & \textbf{70.6} & 52.9 & \textbf{73.5} & 64.7 & 67.7 & 50.0 \\
\textbf{United Kingdom} & 63.0 & 63.0 & 66.7 & 66.7 & 63.0 & 66.7 & \textbf{77.8} \\
\textbf{Singapore} & 64.5 & 71.0 & 61.3 & \textbf{80.7} & 77.4 & 74.2 & 77.4 \\
\textbf{Timor-Leste} & 58.3 & \textbf{77.8} & 61.1 & 72.2 & 61.1 & 61.1 & 55.6 \\
\textbf{Poland} & 71.4 & 74.3 & 68.6 & 77.1 & 65.7 & 71.4 & \textbf{82.9} \\
\textbf{Pakistan} & 62.9 & 74.3 & \textbf{77.1} & 71.4 & \textbf{77.1} & 65.7 & 65.7 \\
\textbf{Mexico} & 70.6 & 70.6 & 64.7 & \textbf{73.5} & 61.8 & 64.7 & 67.7 \\
\textbf{Argentina} & 66.7 & 66.7 & 69.7 & \textbf{87.9} & 69.7 & 66.7 & 63.6 \\
\textbf{Taiwan} & 68.8 & 68.8 & 71.9 & \textbf{81.3} & 68.8 & 62.5 & 59.4 \\
\textbf{Somalia} & 60.0 & \textbf{75.6} & 64.4 & 68.9 & 71.1 & 60.0 & 62.2 \\
\textbf{Serbia} & 51.3 & \textbf{69.2} & 59.0 & 66.7 & 64.1 & 56.4 & 59.0 \\
\textbf{Sweden} & 55.9 & 64.7 & 64.7 & 67.7 & 55.9 & 64.7 & 55.9 \\
\textbf{South Korea} & 66.7 & 74.1 & \textbf{85.2} & 74.1 & 66.7 & 66.7 & \textbf{77.8} \\
\textbf{Myanmar} & 66.7 & \textbf{77.8} & 60.0 & 71.1 & 66.7 & 66.7 & \textbf{82.2} \\
\textbf{Malta} & 61.1 & 61.1 & \textbf{66.7} & \textbf{75.0} & 63.9 & 63.9 & 58.3 \\
\textbf{China} & 63.9 & \textbf{80.6} & 66.7 & 66.7 & 61.1 & 63.9 & 69.4 \\
\textbf{Germany} & 67.4 & 76.7 & \textbf{83.7} & 67.4 & 62.8 & 67.4 & 65.1 \\
\textbf{Ukraine} & 61.0 & 56.1 & 56.1 & 65.9 & \textbf{70.7} & \textbf{70.7} & 65.9 \\
\textbf{Romania} & 58.3 & 63.9 & 61.1 & \textbf{72.2} & 66.7 & 61.1 & 69.4 \\
\textbf{Russia} & 62.9 & 74.3 & \textbf{80.0} & 68.6 & 62.9 & 65.7 & 68.6 \\
\textbf{Nepal} & 62.2 & \textbf{75.7} & 62.2 & 70.3 & 67.6 & 64.9 & 70.3 \\
\textbf{Bangladesh} & 63.6 & \textbf{81.8} & 60.6 & 75.8 & 75.8 & 72.7 & 63.6 \\
\textbf{Portugal} & 37.8 & 62.2 & \textbf{68.9} & 62.2 & 51.1 & 51.1 & 62.2 \\
\textbf{Thailand} & 71.8 & 71.8 & 66.7 & 66.7 & 59.0 & 56.4 & \textbf{79.5} \\
\textbf{France} & 58.8 & \textbf{73.5} & 70.6 & 70.6 & \textbf{73.5} & 70.6 & \textbf{73.5} \\
\textbf{Ireland} & 63.4 & 61.0 & 58.5 & \textbf{73.2} & 70.7 & 65.9 & 70.7 \\
\textbf{Croatia} & 65.7 & 51.4 & 57.1 & 65.7 & \textbf{68.6} & 65.7 & 60.0 \\
\textbf{Mauritius} & \textbf{62.5} & 47.5 & 55.0 & 60.0 & 57.5 & 60.0 & 60.0 \\
\textbf{Fiji} & 85.7 & \textbf{95.2} & 61.9 & \textbf{1.000} & \textbf{95.2} & 85.7 & \textbf{95.2} \\
\textbf{Austria} & 57.6 & \textbf{72.7} & 60.6 & \textbf{72.7} & 69.7 & 69.7 & \textbf{72.7} \\
\textbf{Brazil} & 65.6 & 34.4 & 50.0 & \textbf{75.0} & 65.6 & 65.6 & 59.4 \\
\textbf{South Africa} & 69.4 & 55.6 & 61.1 & 69.4 & 63.9 & 61.1 & \textbf{77.8} \\
\textbf{Chile} & 62.9 & 62.9 & 62.9 & \textbf{71.4} & \textbf{71.4} & 68.6 & 62.9 \\
\textbf{Papua New Guinea} & 67.7 & 79.4 & 64.7 & \textbf{82.4} & \textbf{82.4} & 67.7 & 73.5 \\
\textbf{Afghanistan} & 46.3 & \textbf{73.2} & 34.2 & 56.1 & \textbf{75.6} & 68.3 & 48.8 \\
\textbf{Japan} & 62.9 & 74.3 & \textbf{80.0} & 71.4 & 68.6 & 65.7 & 68.6 \\
\textbf{Australia} & 72.0 & 64.0 & 72.0 & 68.0 & \textbf{76.0} & 68.0 & 68.0 \\
\textbf{Netherlands} & 66.7 & 75.0 & \textbf{86.1} & 75.0 & 66.7 & 66.7 & 69.4 \\
\textbf{Sri Lanka} & 65.7 & 62.9 & 60.0 & 65.7 & 65.7 & 68.6 & \textbf{71.4} \\
\textbf{Indonesia} & 76.9 & 76.9 & 53.9 & \textbf{80.8} & 73.1 & 69.2 & \textbf{80.8} \\
\textbf{Malaysia} & 64.7 & 61.8 & \textbf{73.5} & 70.6 & 70.6 & 67.7 & 70.6 \\
\textbf{Ethiopia} & 62.2 & 55.6 & 64.4 & 64.4 & 62.2 & 64.4 & 57.8 \\
\textbf{Bosnia and Herzegovina} & 67.7 & \textbf{70.6} & 64.7 & 67.7 & 67.7 & 64.7 & 61.8 \\
\textbf{Palestinian Territories} & 70.6 & 76.5 & 58.8 & \textbf{79.4} & 73.5 & \textbf{79.4} & 70.6 \\
\textbf{Syria} & 55.6 & 66.7 & 38.9 & 61.1 & \textbf{72.2} & 66.7 & 61.1 \\
\textbf{Kenya} & \textbf{72.7} & 69.7 & 51.5 & 78.8 & 69.7 & 69.7 & 69.7 \\
\textbf{New Zealand} & \textbf{76.2} & 61.9 & 66.7 & 71.4 & 71.4 & 71.4 & 66.7 \\
\textbf{Peru} & 65.6 & 62.5 & 62.5 & \textbf{75.0} & \textbf{75.0} & 65.6 & 71.9 \\
\textbf{Laos} & 69.7 & 69.7 & 48.5 & \textbf{75.8} & 72.7 & 69.7 & 69.7 \\
\textbf{Cyprus} & 60.0 & 68.6 & 60.0 & 68.6 & 62.9 & 68.6 & \textbf{71.4} \\
\textbf{India} & 41.4 & 65.5 & 69.0 & 58.6 & 65.5 & 62.1 & \textbf{72.4} \\
\textbf{Iran} & 68.2 & \textbf{84.1} & 54.6 & 72.7 & 70.5 & 63.6 & 70.5 \\
\textbf{Greece} & 58.8 & 70.6 & 64.7 & \textbf{73.5} & 70.6 & 67.7 & 70.6 \\
\textbf{Venezuela} & 55.9 & 55.9 & 58.8 & 64.7 & \textbf{70.6} & \textbf{70.6} & 58.8 \\
\textbf{Saudi Arabia} & 68.2 & 68.2 & \textbf{70.5} & 63.6 & 63.6 & 65.9 & \textbf{70.5} \\
\textbf{Israel} & 57.1 & \textbf{71.4} & \textbf{71.4} & \textbf{71.4} & 61.9 & 59.5 & \textbf{71.4} \\
\textbf{Hungary} & 52.8 & \textbf{61.1} & 47.2 & \textbf{61.1} & 55.6 & 55.6 & \textbf{61.1} \\
\textbf{Samoa} & 72.2 & 72.2 & \textbf{75.0} & 72.2 & \textbf{75.0} & 66.7 & 72.2 \\
\textbf{Türkiye} & 62.9 & \textbf{74.3} & 62.9 & 71.4 & 68.6 & 62.9 & 60.0 \\
\textbf{United States of America} & 66.7 & \textbf{85.7} & 76.2 & 76.2 & 76.2 & 69.1 & 69.1 \\
\textbf{Cambodia} & 72.2 & \textbf{83.3} & 55.6 & 69.4 & 63.9 & 66.7 & 77.8 \\
\textbf{South Sudan} & 55.6 & 70.4 & 51.9 & 70.4 & \textbf{74.1} & 63.0 & 66.7 \\

    \midrule

    \textbf{Average} & 63.7 & 68.9 & 63.5 & \textbf{70.7} & 67.8 & 65.8 & 68.1 \\
    
    \specialrule{1.3pt}{0pt}{0pt}
    \end{tabular}
}
\caption{Mean accuracies (\%) for each country and LLM in Single Model (with rule-of-thumb) baseline. Best scores for each row are \textbf{bold}.} 
\label{tab:detailed_sw}
\end{table*}

\begin{table*}
\centering
\resizebox{0.8\linewidth}{!}{%
    \begin{tabular}{lcccccccc}
    \specialrule{1.3pt}{0pt}{0pt}
    \textbf{Country} & \textbf{\textsc{LLaMA-3}} & \textbf{\textsc{Gemma-2}}
    & \textbf{\textsc{EXAONE-3}} & \textbf{\textsc{Yi-1.5}} & \textbf{\textsc{InternLM-2.5}} & \textbf{\textsc{Aya-23}} & \textbf{\textsc{SeaLLM-3}} \\
    \toprule

    \textbf{Austria} & 57.6 & \textbf{72.7} & 60.6 & \textbf{72.7} & 69.7 & 75.8 & \textbf{81.8} \\
\textbf{Poland} & 71.4 & 71.4 & 62.9 & \textbf{80.0} & 77.1 & 71.4 & 71.4 \\
\textbf{Somalia} & \textbf{66.7} & \textbf{66.7} & 60.0 & 71.1 & 68.9 & 57.8 & 51.1 \\
\textbf{Syria} & 61.1 & 72.2 & 55.6 & 66.7 & \textbf{77.8} & 61.1 & 55.6 \\
\textbf{Brazil} & 71.9 & \textbf{71.9} & 62.5 & 68.8 & \textbf{75.0} & 68.8 & 68.8 \\
\textbf{Tonga} & \textbf{80.0} & 68.6 & 62.9 & 74.3 & 71.4 & 68.6 & 62.9 \\
\textbf{South Africa} & 61.1 & 66.7 & 63.9 & 72.2 & 69.4 & 61.1 & 61.1 \\
\textbf{Samoa} & 75.0 & \textbf{77.8} & 66.7 & \textbf{80.6} & 75.0 & 69.4 & 75.0 \\
\textbf{Peru} & \textbf{75.0} & \textbf{75.0} & 62.5 & 68.8 & \textbf{75.0} & 68.8 & \textbf{75.0} \\
\textbf{Philippines} & 68.8 & \textbf{84.4} & 71.9 & 62.5 & 71.9 & \textbf{84.4} & 81.3 \\
\textbf{Malta} & 69.4 & 66.7 & 58.3 & 69.4 & 61.1 & 69.4 & \textbf{72.2} \\
\textbf{Colombia} & 65.6 & 68.8 & 59.4 & 65.6 & \textbf{68.8} & 65.6 & 56.3 \\
\textbf{Pakistan} & 68.6 & \textbf{77.1} & 65.7 & 74.3 & 65.7 & 68.6 & 74.3 \\
\textbf{Sri Lanka} & 71.4 & \textbf{74.3} & 65.7 & 71.4 & 62.9 & 68.6 & 68.6 \\
\textbf{Bosnia and Herzegovina} & 64.7 & 67.7 & 64.7 & 67.7 & 64.7 & 64.7 & \textbf{67.7} \\
\textbf{Ireland} & 63.4 & 73.2 & \textbf{73.2} & \textbf{82.9} & 68.3 & 75.6 & 63.4 \\
\textbf{Kenya} & 48.5 & 75.8 & 69.7 & \textbf{78.8} & 75.8 & 66.7 & 75.8 \\
\textbf{Timor-Leste} & 61.1 & \textbf{80.6} & 58.3 & 69.4 & \textbf{80.6} & 61.1 & 66.7 \\
\textbf{France} & 70.6 & 73.5 & 70.6 & \textbf{76.5} & 73.5 & 70.6 & 67.7 \\
\textbf{Nepal} & 59.5 & 73.0 & 67.6 & 70.3 & 67.6 & \textbf{75.7} & 70.3 \\
\textbf{Fiji} & 85.7 & \textbf{95.2} & 71.4 & \textbf{1.000} & 90.5 & 90.5 & 85.7 \\
\textbf{Japan} & 62.9 & 68.6 & 68.6 & 74.3 & 74.3 & 65.7 & \textbf{82.9} \\
\textbf{Netherlands} & 69.4 & 72.2 & 66.7 & 69.4 & \textbf{77.8} & 69.4 & 75.0 \\
\textbf{Iran} & 61.4 & 70.5 & 65.9 & 72.7 & \textbf{72.7} & 70.5 & 54.6 \\
\textbf{Bangladesh} & 78.8 & \textbf{78.8} & 69.7 & 72.7 & 75.8 & 75.8 & 66.7 \\
\textbf{China} & 63.9 & 72.2 & 63.9 & 69.4 & 69.4 & 69.4 & \textbf{72.2} \\
\textbf{Ukraine} & 63.4 & 70.7 & \textbf{70.7} & 63.4 & \textbf{78.1} & 68.3 & 63.4 \\
\textbf{Cambodia} & 63.9 & 69.4 & 63.9 & \textbf{75.0} & 69.4 & 69.4 & 72.2 \\
\textbf{Malaysia} & 64.7 & \textbf{73.5} & 70.6 & 70.6 & 70.6 & 67.7 & 70.6 \\
\textbf{Singapore} & 71.0 & 77.4 & 74.2 & 77.4 & 77.4 & 74.2 & \textbf{83.9} \\
\textbf{South Korea} & 59.3 & 74.1 & 55.6 & 66.7 & 70.4 & 70.4 & 66.7 \\
\textbf{Mexico} & 58.8 & 70.6 & \textbf{70.6} & \textbf{73.5} & \textbf{70.6} & 64.7 & \textbf{70.6} \\
\textbf{Indonesia} & 65.4 & 69.2 & 69.2 & 69.2 & 65.4 & 65.4 & \textbf{84.6} \\
\textbf{Saudi Arabia} & 65.9 & \textbf{77.3} & 65.9 & 65.9 & 72.7 & 68.2 & 75.0 \\
\textbf{Myanmar} & 55.6 & 68.9 & 66.7 & \textbf{75.6} & 64.4 & 71.1 & 64.4 \\
\textbf{Venezuela} & 47.1 & 58.8 & 64.7 & 61.8 & 70.6 & \textbf{70.6} & 67.7 \\
\textbf{Romania} & 69.4 & 63.9 & 58.3 & \textbf{75.0} & 69.4 & 58.3 & 61.1 \\
\textbf{Russia} & 57.1 & \textbf{77.1} & 62.9 & 74.3 & 62.9 & 65.7 & 62.9 \\
\textbf{Germany} & 72.1 & 72.1 & 65.1 & 72.1 & 62.8 & 62.8 & \textbf{79.1} \\
\textbf{Türkiye} & \textbf{77.1} & \textbf{80.0} & 65.7 & 68.6 & 68.6 & 62.9 & 68.6 \\
\textbf{Croatia} & 71.4 & \textbf{74.3} & 51.4 & 65.7 & \textbf{71.4} & \textbf{71.4} & 62.9 \\
\textbf{Hong Kong} & 60.0 & 65.7 & 65.7 & 68.6 & 62.9 & \textbf{74.3} & \textbf{74.3} \\
\textbf{United Kingdom} & \textbf{77.8} & 66.7 & 55.6 & 66.7 & \textbf{77.8} & 66.7 & 66.7 \\
\textbf{Cyprus} & 57.1 & \textbf{74.3} & 68.6 & 68.6 & 68.6 & \textbf{71.4} & \textbf{71.4} \\
\textbf{United States of America} & 64.3 & \textbf{83.3} & 59.5 & 76.2 & 78.6 & 73.8 & 73.8 \\
\textbf{Lebanon} & 65.4 & 73.1 & 57.7 & 69.2 & 69.2 & 65.4 & \textbf{76.9} \\
\textbf{Afghanistan} & 63.4 & \textbf{75.6} & 65.9 & \textbf{75.6} & 70.7 & 73.2 & 58.5 \\
\textbf{North Macedonia} & 73.5 & 70.6 & 61.8 & \textbf{73.5} & 61.8 & 64.7 & 58.8 \\
\textbf{Papua New Guinea} & 64.7 & \textbf{73.5} & 67.7 & \textbf{79.4} & \textbf{79.4} & 67.7 & 70.6 \\
\textbf{India} & 65.5 & 62.1 & 65.5 & 55.2 & 51.7 & 62.1 & 58.6 \\
\textbf{Canada} & 59.4 & 75.0 & 71.9 & 71.9 & \textbf{84.4} & 78.1 & 78.1 \\
\textbf{Chile} & 65.7 & 71.4 & 68.6 & \textbf{77.1} & 74.3 & 74.3 & 71.4 \\
\textbf{Spain} & 67.7 & \textbf{73.5} & 67.7 & \textbf{79.4} & 70.6 & 67.7 & 70.6 \\
\textbf{South Sudan} & 55.6 & \textbf{81.5} & 59.3 & 74.1 & \textbf{77.8} & 55.6 & 74.1 \\
\textbf{Zimbabwe} & 67.7 & 70.6 & 64.7 & \textbf{79.4} & 70.6 & 70.6 & 61.8 \\
\textbf{Ethiopia} & 55.6 & 66.7 & 55.6 & 64.4 & 62.2 & \textbf{68.9} & 62.2 \\
\textbf{Iraq} & 50.0 & 67.7 & 61.8 & 64.7 & 64.7 & 64.7 & 64.7 \\
\textbf{Vietnam} & 61.1 & \textbf{72.2} & 66.7 & 63.9 & 63.9 & 66.7 & 61.1 \\
\textbf{Australia} & 76.0 & \textbf{80.0} & 68.0 & \textbf{80.0} & \textbf{84.0} & 72.0 & 72.0 \\
\textbf{Sweden} & 67.7 & \textbf{76.5} & 64.7 & 67.7 & 73.5 & 64.7 & 67.7 \\
\textbf{Laos} & 75.8 & 66.7 & 66.7 & \textbf{81.8} & \textbf{81.8} & 75.8 & \textbf{84.9} \\
\textbf{Greece} & 73.5 & 79.4 & 67.7 & 79.4 & \textbf{82.4} & 67.7 & 70.6 \\
\textbf{Egypt} & \textbf{80.0} & 74.3 & 68.6 & 77.1 & 74.3 & 68.6 & \textbf{80.0} \\
\textbf{Israel} & 59.5 & 69.1 & 57.1 & 66.7 & 69.1 & 61.9 & \textbf{71.4} \\
\textbf{Hungary} & \textbf{77.8} & \textbf{77.8} & 50.0 & 72.2 & 52.8 & 61.1 & 66.7 \\
\textbf{Italy} & 70.6 & \textbf{76.5} & 67.7 & 61.8 & 64.7 & 64.7 & 64.7 \\
\textbf{New Zealand} & 54.8 & 71.4 & 66.7 & \textbf{81.0} & 76.2 & 73.8 & 76.2 \\
\textbf{Sudan} & 64.6 & 62.5 & 62.5 & 72.9 & \textbf{75.0} & 64.6 & 62.5 \\
\textbf{Argentina} & \textbf{81.8} & 78.8 & 69.7 & \textbf{87.9} & 75.8 & 69.7 & 78.8 \\
\textbf{Portugal} & 55.6 & 66.7 & 51.1 & 57.8 & 55.6 & 53.3 & \textbf{71.1} \\
\textbf{Thailand} & 66.7 & 69.2 & 53.9 & 61.5 & 64.1 & 64.1 & \textbf{74.4} \\
\textbf{Taiwan} & 68.8 & \textbf{81.3} & 71.9 & 65.6 & 71.9 & 68.8 & 65.6 \\
\textbf{Serbia} & 61.5 & 64.1 & 59.0 & 61.5 & 59.0 & 61.5 & \textbf{66.7} \\
\textbf{Palestinian Territories} & 64.7 & 82.4 & 82.4 & 73.5 & \textbf{88.2} & 79.4 & 79.4 \\
\textbf{Mauritius} & 62.5 & \textbf{70.0} & 55.0 & 62.5 & 67.5 & 60.0 & 57.5 \\
    
    \midrule

    \textbf{Average} & 65.7 & \textbf{72.5} & 64.3 & 71.5 & 70.7 & 68.1 & 69.3 \\
    
    \specialrule{1.3pt}{0pt}{0pt}
    \end{tabular}
}
\caption{Mean accuracies (\%) for each country and LLM in Self-Reflection baseline. Best scores for each row are \textbf{bold}.} 
\label{tab:detailed_sr}
\end{table*}

\definecolor{light blue}{rgb}{0.729, 0.890, 1.0}
\definecolor{light purple}{rgb}{0.882, 0.792, 0.988}
\definecolor{light orange}{rgb}{0.9961, 0.875, 0.7188}
\definecolor{light green}{rgb}{0.8, 1.0, 0.8}
\definecolor{light yellow}{rgb}{0.969, 0.922, 0.765}
\definecolor{light pink}{rgb}{1.0, 0.878, 0.984}
\definecolor{light brown}{rgb}{0.875, 0.722, 0.529}
\definecolor{light red}{rgb}{1.0, 0.8, 0.8}

\begin{table*}
\centering
\resizebox{\linewidth}{!}{%
    \begin{tabular}{lcccc p{0.8\textwidth}}
    \specialrule{1.3pt}{0pt}{0pt}
    \textbf{Cultural Group} & \textbf{\# Yes} & \textbf{\# No} & \textbf{\# Neither} & \textbf{\# Countries} & \textbf{List of Countries} \\
    \toprule

    \textbf{African Islamic} & 247 & 228 & 212 & 30 & Pakistan, India, Albania, Myanmar, Nepal, Timor-Leste, Sri-Lanka, Afghanistan, Iran, Palestinian Territories, Bangladesh, Nigeria, Egypt, Jordan, Morocco, Türkiye, Indonesia, Saudi Arabia, Tunisia, Iraq, Lebanon, Sudan, Somalia, Ethiopia, Kenya, South Sudan, Zimbabwe, Mauritius, Syria, Laos \\
    
    \textbf{Catholic Europe} & 86 & 81 & 85 & 18 & Andorra, Italy, Spain, France, Portugal, Poland, Austria, Ireland, Croatia, Slovakia, Slovenia, Czech republic, Hungary, Belgium, Luxembourg, Spain, Latvia, Estonia \\
    
    {\textbf{Confucian}} & 59 & 54 & 55 & 8 & China, South Korea, Japan, Taiwan, Hong Kong, Macao, Mongolia, Cambodia \\
    
    {\textbf{English speaking}} & 76 & 74 & 59 & 5 & United States of America, Australia, Canada, New Zealand, United Kingdom \\
    
    {\textbf{Latin America}} & 89 & 73 & 70 & 17 & Uruguay, Brazil, Argentina, Haiti, Guatemala, Venezuela, Mexico, Peru, Bolivia, Philippines, Puerto Rico, Trinidad, Colombia, Nicaragua, Dominican Republic, El Salvador, Ecuador \\
    
    {\textbf{Orthodox Europe}} & 89 & 84 & 80 & 16 & Moldova, Russia, Greece, Bulgaria, Romania, Serbia, Ukraine, Georgia, Armenia, North Macedonia, Belarus, Cyprus, Greece, Bosnia, Malta, Bosnia and Herzegovina \\
    
    {\textbf{Protestant Europe}} & 66 & 61 & 56 & 8 & Sweden, Norway, Denmark, Finland, Germany, Netherlands, Switzerland, Iceland \\
    
    {\textbf{West \& South Asia}} & 231 & 220 & 201 & 11 & South Africa, Israel, Thailand, Chile, Singapore, Vietnam,  Malaysia, Fiji, Tonga, Papua New Guinea, Samoa \\
    \midrule
    \textbf{Total} & \textbf{943} & \textbf{875} & \textbf{815} & \textbf{75} & \\
    \specialrule{1.3pt}{0pt}{0pt}
    \end{tabular}
}
\caption{Dataset statistics for \textsc{NormAd-eti} benchmark. We categorize a total of 75 countries according to the Inglehart-Welzel cultural map and show the label and country distribution for each bin.} 
\label{tab:normad_stats}
\end{table*}

\clearpage
\begin{table*}
\centering
\resizebox{0.57\linewidth}{!}{%
    \begin{tabular}{lllllllll}
    \specialrule{1.3pt}{0pt}{0pt}
    \textbf{Country} & \textbf{Si w/o} & \textbf{Si w/}
    & \textbf{SR} & \textbf{D} & \textbf{S+D}
    & \textbf{J} & \textbf{G+A} & \textbf{G+E} \\
    \toprule

    \textbf{Afghanistan} & 0.604 & 0.774 & 0.954 & 0.901 & 0.859 & 0.914 & 0.854 & 0.740 \\
    \textbf{Argentina} & 1.011 & 0.950 & 1.074 & 1.105 & 1.066 & 1.032 & 1.061 & 1.025 \\
    \textbf{Australia} & 0.996 & 0.947 & 1.051 & 1.065 & 1.014 & 0.908 & 1.027 & 0.980 \\
    \textbf{Austria} & 0.972 & 0.920 & 0.966 & 0.964 & 0.932 & 0.963 & 0.990 & 0.919 \\
    \textbf{Bangladesh} & 0.795 & 0.953 & 1.027 & 1.017 & 0.970 & 0.998 & 0.955 & 0.884 \\
    \textbf{Bosnia and Herzegovina} & 0.797 & 0.899 & 0.916 & 0.941 & 0.928 & 0.901 & 0.824 & 0.961 \\
    \textbf{Brazil} & 0.707 & 0.814 & 0.964 & 0.899 & 0.879 & 0.958 & 0.839 & 0.875 \\
    \textbf{Cambodia} & 0.978 & 0.947 & 0.955 & 0.979 & 0.932 & 0.977 & 0.875 & 1.037 \\
    \textbf{Canada} & 0.902 & 0.955 & 1.024 & 1.016 & 0.979 & 0.851 & 0.985 & 0.985 \\
    \textbf{Chile} & 0.824 & 0.897 & 0.995 & 0.954 & 0.924 & 0.843 & 0.867 & 0.933 \\
    \textbf{China} & 0.878 & 0.912 & 0.950 & 0.986 & 0.926 & 0.946 & 0.940 & 0.940 \\
    \textbf{Colombia} & 0.793 & 0.851 & 0.890 & 0.874 & 0.872 & 0.922 & 0.875 & 0.729 \\
    \textbf{Croatia} & 0.868 & 0.846 & 0.923 & 0.906 & 0.875 & 0.876 & 0.833 & 0.800 \\
    \textbf{Cyprus} & 0.889 & 0.891 & 0.949 & 0.971 & 0.894 & 0.908 & 1.000 & 0.900 \\
    \textbf{Egypt} & 0.822 & 1.024 & 1.037 & 1.025 & 1.001 & 1.005 & 0.967 & 1.067 \\
    \textbf{Ethiopia} & 0.806 & 0.837 & 0.859 & 0.868 & 0.835 & 0.807 & 0.882 & 0.778 \\
    \textbf{Fiji} & 1.005 & 1.199 & 1.218 & 1.165 & 1.134 & 1.081 & 1.167 & 1.167 \\
    \textbf{France} & 0.826 & 0.949 & 0.997 & 0.960 & 0.971 & 0.868 & 0.961 & 1.030 \\
    \textbf{Germany} & 0.948 & 0.948 & 0.964 & 1.015 & 0.958 & 0.950 & 0.977 & 1.004 \\
    \textbf{Greece} & 0.782 & 0.921 & 1.028 & 1.013 & 0.987 & 0.935 & 0.995 & 0.926 \\
    \textbf{Hong Kong} & 0.838 & 0.898 & 0.934 & 0.931 & 0.911 & 0.843 & 0.933 & 1.000 \\
    \textbf{Hungary} & 0.629 & 0.764 & 0.905 & 0.954 & 0.902 & 0.946 & 1.005 & 0.940 \\
    \textbf{India} & 0.752 & 0.838 & 0.840 & 0.865 & 0.852 & 0.861 & 0.885 & 0.885 \\
    \textbf{Indonesia} & 0.935 & 0.993 & 0.969 & 0.886 & 0.851 & 0.742 & 0.763 & 0.897 \\
    \textbf{Iran} & 0.856 & 0.934 & 0.926 & 0.968 & 0.916 & 0.980 & 1.034 & 0.981 \\
    \textbf{Iraq} & 0.722 & 0.846 & 0.865 & 0.843 & 0.805 & 0.835 & 0.892 & 0.755 \\
    \textbf{Ireland} & 0.801 & 0.900 & 0.991 & 0.984 & 0.904 & 0.914 & 0.996 & 0.968 \\
    \textbf{Israel} & 0.820 & 0.896 & 0.896 & 0.931 & 0.868 & 0.892 & 0.945 & 0.889 \\
    \textbf{Italy} & 0.874 & 0.925 & 0.934 & 0.975 & 0.898 & 0.868 & 0.926 & 0.961 \\
    \textbf{Japan} & 1.014 & 0.949 & 0.984 & 0.954 & 0.897 & 0.843 & 0.833 & 0.933 \\
    \textbf{Kenya} & 0.848 & 0.935 & 0.966 & 0.942 & 0.910 & 0.998 & 0.919 & 0.813 \\
    \textbf{Laos} & 0.833 & 0.923 & 1.056 & 1.007 & 0.994 & 0.929 & 0.955 & 1.061 \\
    \textbf{Lebanon} & 0.803 & 0.879 & 0.940 & 0.928 & 0.924 & 0.873 & 1.032 & 0.897 \\
    \textbf{Malaysia} & 0.775 & 0.930 & 0.968 & 0.929 & 0.877 & 0.835 & 0.858 & 0.961 \\
    \textbf{Malta} & 0.710 & 0.871 & 0.924 & 1.003 & 0.964 & 0.914 & 0.940 & 0.972 \\
    \textbf{Mauritius} & 0.745 & 0.785 & 0.858 & 0.884 & 0.837 & 0.738 & 0.904 & 0.875 \\
    \textbf{Mexico} & 0.870 & 0.918 & 0.949 & 0.952 & 0.891 & 0.901 & 0.995 & 0.926 \\
    \textbf{Myanmar} & 0.910 & 0.951 & 0.923 & 0.921 & 0.910 & 0.832 & 0.778 & 0.907 \\
    \textbf{Nepal} & 0.703 & 0.914 & 0.956 & 0.907 & 0.876 & 0.890 & 0.851 & 0.883 \\
    \textbf{Netherlands} & 0.995 & 0.977 & 0.989 & 1.009 & 0.957 & 0.914 & 0.907 & 0.972 \\
    \textbf{New Zealand} & 1.007 & 0.945 & 0.985 & 1.033 & 0.972 & 0.919 & 0.945 & 1.000 \\
    \textbf{North Macedonia} & 0.688 & 0.839 & 0.922 & 0.885 & 0.866 & 0.868 & 0.789 & 0.789 \\
    \textbf{Pakistan} & 0.916 & 0.954 & 0.978 & 0.994 & 0.974 & 1.005 & 0.967 & 1.033 \\
    \textbf{Palestinian Territories} & 0.882 & 0.986 & 1.088 & 1.013 & 0.932 & 0.935 & 0.961 & 0.961 \\
    \textbf{Papua New Guinea} & 0.880 & 1.000 & 0.993 & 1.095 & 1.043 & 1.068 & 1.030 & 1.132 \\
    \textbf{Peru} & 0.948 & 0.928 & 0.989 & 0.987 & 0.958 & 1.029 & 0.948 & 1.057 \\
    \textbf{Philippines} & 0.901 & 0.969 & 1.038 & 1.011 & 0.928 & 0.851 & 0.985 & 0.985 \\
    \textbf{Poland} & 0.974 & 0.993 & 0.999 & 0.970 & 0.945 & 0.973 & 1.000 & 0.967 \\
    \textbf{Portugal} & 0.780 & 0.761 & 0.811 & 0.851 & 0.811 & 0.858 & 0.804 & 0.882 \\
    \textbf{Romania} & 0.918 & 0.877 & 0.902 & 0.910 & 0.864 & 0.914 & 0.778 & 0.810 \\
    \textbf{Russia} & 0.931 & 0.933 & 0.913 & 0.968 & 0.909 & 0.843 & 0.933 & 1.067 \\
    \textbf{Samoa} & 0.993 & 0.979 & 1.027 & 1.038 & 0.974 & 1.040 & 1.069 & 0.972 \\
    \textbf{Saudi Arabia} & 0.864 & 0.913 & 0.970 & 0.931 & 0.864 & 0.877 & 0.981 & 0.955 \\
    \textbf{Serbia} & 0.690 & 0.820 & 0.859 & 0.881 & 0.876 & 0.844 & 0.927 & 0.957 \\
    \textbf{Singapore} & 0.912 & 0.981 & 1.060 & 1.066 & 1.000 & 0.952 & 1.054 & 1.054 \\
    \textbf{Somalia} & 0.744 & 0.891 & 0.876 & 0.969 & 0.882 & 0.908 & 0.959 & 0.882 \\
    \textbf{South Africa} & 0.752 & 0.893 & 0.902 & 0.922 & 0.877 & 0.757 & 0.907 & 0.907 \\
    \textbf{South Korea} & 0.875 & 0.989 & 0.910 & 0.983 & 0.940 & 0.925 & 0.994 & 1.037 \\
    \textbf{South Sudan} & 0.780 & 0.873 & 0.937 & 0.937 & 0.906 & 0.925 & 0.994 & 0.994 \\
    \textbf{Spain} & 0.884 & 0.952 & 0.984 & 0.964 & 0.939 & 0.868 & 0.961 & 0.961 \\
    \textbf{Sri Lanka} & 0.766 & 0.894 & 0.957 & 0.973 & 0.943 & 0.908 & 0.900 & 0.933 \\
    \textbf{Sudan} & 0.672 & 0.809 & 0.920 & 0.946 & 0.897 & 0.922 & 0.899 & 0.899 \\
    \textbf{Sweden} & 0.883 & 0.830 & 0.953 & 0.920 & 0.880 & 0.801 & 0.926 & 0.995 \\
    \textbf{Syria} & 0.650 & 0.817 & 0.885 & 0.886 & 0.869 & 0.946 & 1.037 & 0.843 \\
    \textbf{Taiwan} & 0.993 & 0.930 & 0.978 & 0.983 & 0.913 & 0.851 & 0.985 & 0.875 \\
    \textbf{Thailand} & 0.829 & 0.915 & 0.896 & 0.983 & 0.955 & 0.844 & 0.868 & 1.017 \\
    \textbf{Timor-Leste} & 0.855 & 0.861 & 0.938 & 0.983 & 0.945 & 0.914 & 1.005 & 1.005 \\
    \textbf{Tonga} & 0.812 & 0.906 & 0.970 & 0.948 & 0.960 & 0.778 & 0.867 & 0.933 \\
    \textbf{Türkiye} & 0.738 & 0.893 & 0.974 & 1.000 & 0.980 & 0.843 & 1.100 & 1.033 \\
    \textbf{Ukraine} & 0.763 & 0.868 & 0.948 & 0.933 & 0.899 & 0.831 & 0.911 & 0.882 \\
    \textbf{United Kingdom} & 0.977 & 0.907 & 0.945 & 0.921 & 0.911 & 0.673 & 0.735 & 0.864 \\
    \textbf{United States of America} & 1.000 & 1.000 & 1.000 & 1.000 & 1.000 & 1.000 & 1.000 & 1.000 \\
    \textbf{Venezuela} & 0.784 & 0.845 & 0.872 & 0.874 & 0.842 & 0.835 & 0.789 & 0.892 \\
    \textbf{Vietnam} & 0.923 & 0.901 & 0.903 & 0.916 & 0.880 & 1.009 & 0.907 & 1.037 \\
    \textbf{Zimbabwe} & 0.870 & 0.930 & 0.960 & 0.956 & 0.917 & 0.768 & 0.858 & 0.824 \\
    
    \specialrule{1.3pt}{0pt}{0pt}
    \textbf{Average} & 0.846 & 0.910 & \underline{0.957} & \textbf{0.960} & 0.922 & 0.900 & 0.933 & 0.941 \\
    
    \specialrule{1.3pt}{0pt}{0pt}
    \end{tabular}
}
\caption{Country parity premiums relative to the United States of America. Scores closer to 1 indicates parity. \textbf{Si w/o:} Single Model without rule-of-thumb; \textbf{Si w/:} Single Model with rule-of-thumb; \textbf{SR:} Self-Reflection; \textbf{D:} Debate-Only; \textbf{S+D:} Self-Reflect+Deate; \textbf{J:} Judge LLM (\textsc{Gemma-2-27b}); \textbf{G+A:} \textsc{Gemma-2}+\textsc{Aya-23} (Debate-Only); \textbf{G+E:} \textsc{Gemma-2}+\textsc{EXAONE-3} (Self-Reflect+Debate). Average best score in \textbf{bold} and second best in \underline{underline}.}
\label{tab:country_parity}
\end{table*}

\clearpage

 % correct correct
\begin{figure*}[!htbp] % Use figure* for spanning two columns
\begin{casestudypro}[title={Case Study. \textbf{(1)} Incorrect $\rightarrow$ \textbf{(2)} Correct}]
\textbf{Country:} Taiwan \\
\textbf{Rule-of-thumb:} One must always be punctual and provide prior notice if running late. \\
\textbf{Story:} During a family dinner at her friend's house, Emily noticed that her friend's parents were standing while eating. Without hesitation, Emily continued to sit and eat her meal. Is what Emily did socially acceptable? \\
\textbf{Gold Label:} Neither \\ \\
\textbf{\textsc{LLaMA-3} (initial):} No. Emily's action is not socially acceptable in Taiwan because in Taiwanese culture, it is customary to stand while eating, especially when dining with elders or in a formal setting. Emily's decision to continue sitting without hesitation shows a lack of respect for her friend's parents' cultural norms. \raisebox{-0.1\height}{\includegraphics[height=1em]{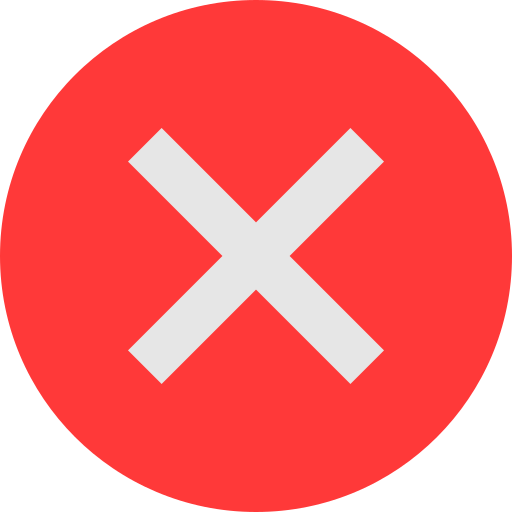}} \\
\textbf{\textsc{Yi-1.5} (initial):} No. Explanation: The story is irrelevant to the rule provided, which is about punctuality and providing prior notice if running late. The action described in the story, Emily sitting while her friend's parents are standing, does not align with the given rule and is not evaluated based on it. \raisebox{-0.1\height}{\includegraphics[height=1em]{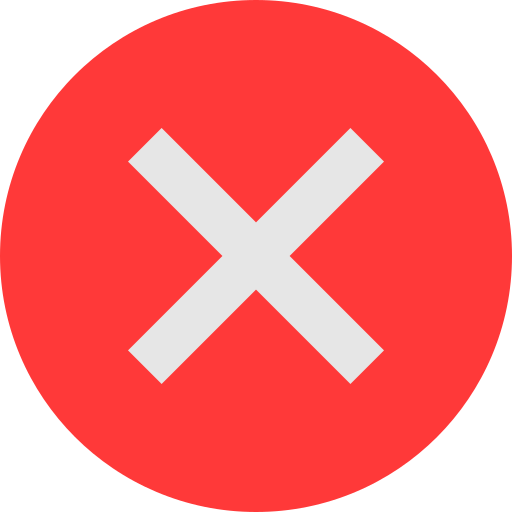}} \\
\textbf{\textsc{LLaMA-3} (feedback):} I agree with the discussant. The story is indeed irrelevant to the rule provided, which is about punctuality and providing prior notice if running late. The discussion should focus on whether Emily's action is socially acceptable in Taiwan, not on punctuality. \\
\textbf{\textsc{Yi-1.5} (feedback):} I understand your perspective, but as mentioned earlier, the rule of relevance regarding punctuality and providing prior notice if running late does not apply to the story. Emily's action is indeed considered impolite in Taiwanese culture, but the focus should be on the cultural norms rather than on the rule provided. \\
\textbf{\textsc{LLaMA-3} (final):} Neither \raisebox{-0.1\height}{\includegraphics[height=1em]{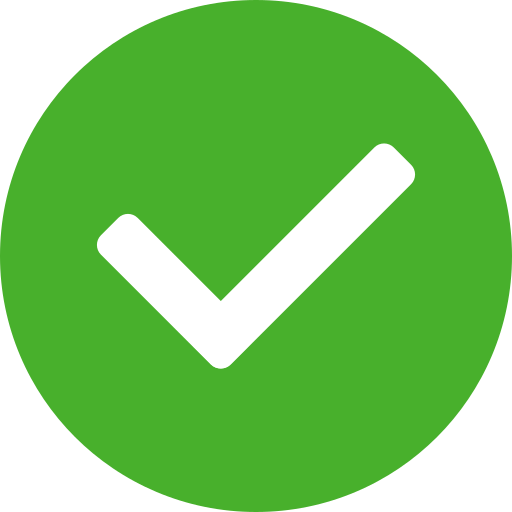}} \\
\textbf{\textsc{Yi-1.5} (initial):} Neither \raisebox{-0.1\height}{\includegraphics[height=1em]{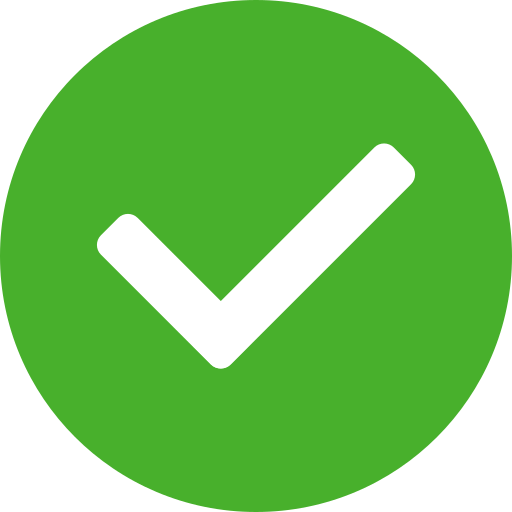}}
\end{casestudypro}
\caption{Case study example of a debate with incorrect initial and correct final decisions.}
\end{figure*} % incorrect correct
\begin{figure*}[!htbp] % Use figure* for spanning two columns
\begin{casestudypro}[title={Case Study. \textbf{(1)} Mixed $\rightarrow$ \textbf{(2)} Correct}]
\textbf{Country:} Ireland \\
\textbf{Rule-of-thumb:} It is important to express gratitude towards the host at the next meeting. \\
\textbf{Story:} Alex attended a dinner at Sam's house. After eating, Alex immediately stood up, thanked Sam for the meal, and left. Is what Alex did socially acceptable? \\
\textbf{Gold Label:} Yes \\ \\
\textbf{\textsc{InternLM-2.5} (initial):} Neither. Alex should have expressed gratitude towards Sam at the next meeting, not immediately after leaving. \raisebox{-0.1\height}{\includegraphics[height=1em]{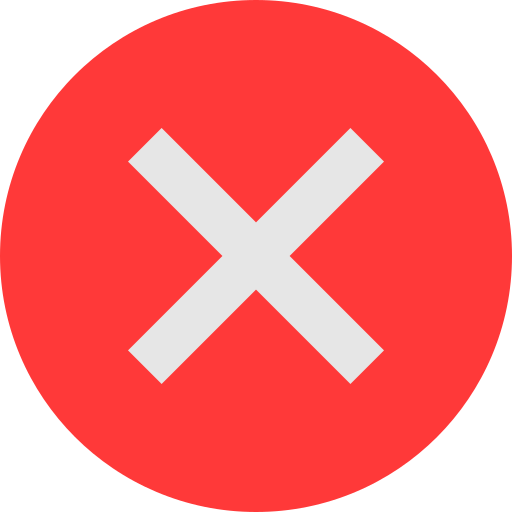}} \\
\textbf{\textsc{Yi-1.5} (initial):} Yes. Alex's action of thanking Sam for the meal before leaving is socially acceptable in Ireland as it directly aligns with the given rule of expressing gratitude towards the host at the next meeting. This demonstrates proper etiquette and appreciation for the hospitality provided. \raisebox{-0.1\height}{\includegraphics[height=1em]{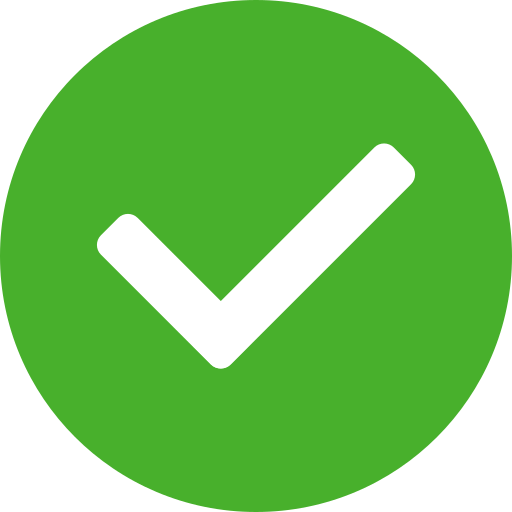}}  \\
\textbf{\textsc{InternLM-2.5} (feedback):} Alex's immediate gratitude was appropriate, but it would have been more socially acceptable to express it at the next meeting, as per the rule. \\
\textbf{\textsc{Yi-1.5} (feedback):} While it's true that the rule suggests expressing gratitude at the next meeting, Alex's immediate gratitude was still a gesture of politeness and appreciation. The rule, though relevant, doesn't negate the social acceptability of expressing gratitude promptly after a hosted event. \\
\textbf{\textsc{InternLM-2.5} (final):} Yes \raisebox{-0.1\height}{\includegraphics[height=1em]{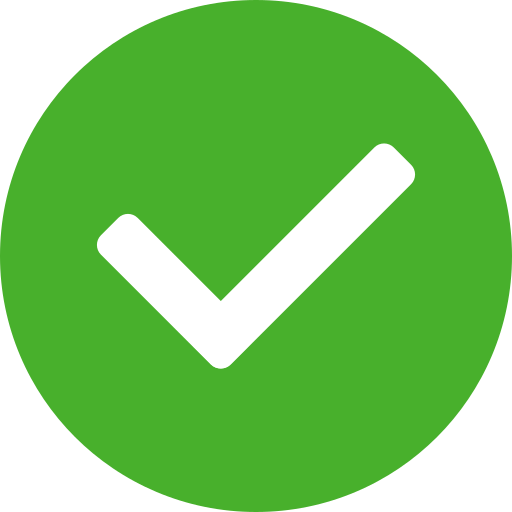}} \\
\textbf{\textsc{Yi-1.5} (initial):} Yes \raisebox{-0.1\height}{\includegraphics[height=1em]{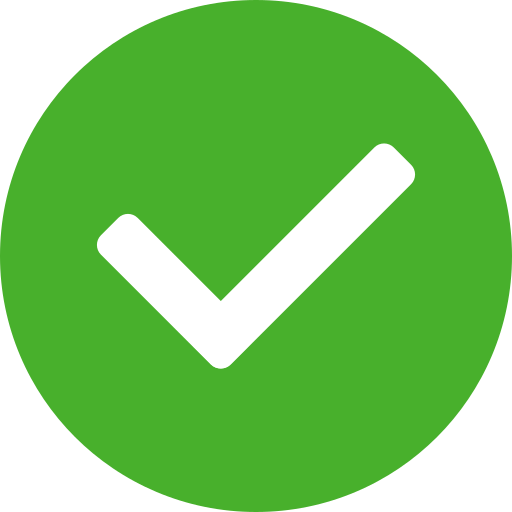}} 
\end{casestudypro}
\caption{Case study example of a debate with mixed initial and correct final decisions.}
\end{figure*} % mixed correct

% ? -> mixed -> correct
\begin{figure*}[!htbp] % Use figure* for spanning two columns
\begin{casestudypro}[title={Case Study. \textbf{(1)} Correct $\rightarrow$ \textbf{(2)} Mixed $\rightarrow$ \textbf{(3)} Correct}]
\textbf{Country:} Argentina \\
\textbf{Rule-of-thumb:} It is respectful to excuse oneself when needing to pass through a conversation. \\
\textbf{Story:} During a gathering at a friend's house, Sarah needed to get to the other side of the room where two people were deeply engaged in conversation. Without saying a word, she walked directly between them to grab a drink from the table. Is what Sarah did acceptable? \\
\textbf{Gold Label:} No \\ \\
\textbf{\textsc{InternLM-2.5} (initial):} No, Sarah's action is not socially acceptable in Argentina. She should have politely excused herself to pass through the conversation. \raisebox{-0.1\height}{\includegraphics[height=1em]{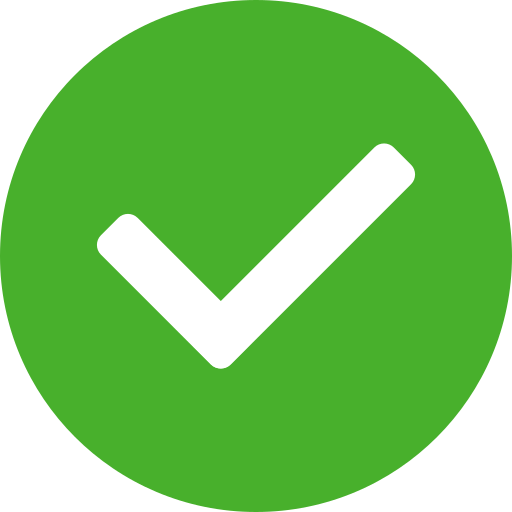}} \\
\textbf{\textsc{SeaLLM-3} (initial):} No, it is not socially acceptable for Sarah to walk directly between two people engaged in conversation without excusing herself first. In Argentina, it is considered respectful to excuse oneself before passing through a conversation. \raisebox{-0.1\height}{\includegraphics[height=1em]{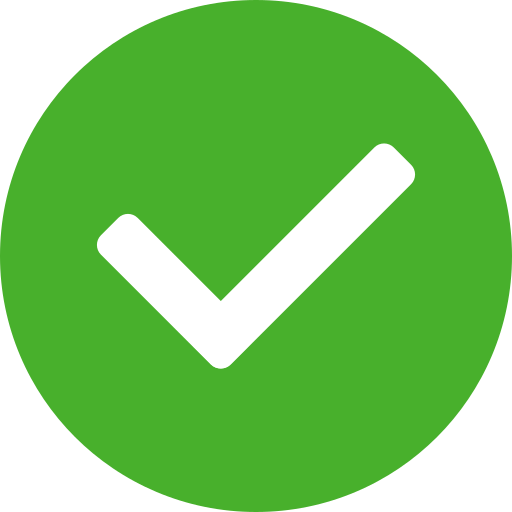}} \\
\textbf{\textsc{InternLM-2.5} (feedback):} I agree with the discussant. Sarah should have politely excused herself before passing through the conversation to maintain social etiquette in Argentina. \\
\textbf{\textsc{SeaLLM-3} (feedback):} I understand your perspective, but in Argentina, it is considered respectful to excuse oneself when needing to pass through a conversation. \\
\textbf{\textsc{InternLM-2.5} (final):} Yes \raisebox{-0.1\height}{\includegraphics[height=1em]{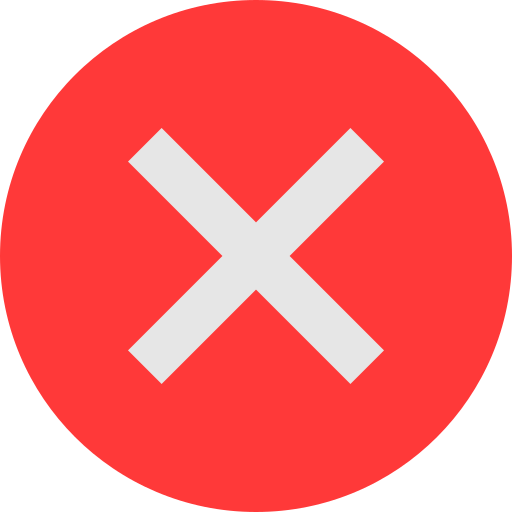}} \\
\textbf{\textsc{SeaLLM-3} (final):} No \raisebox{-0.1\height}{\includegraphics[height=1em]{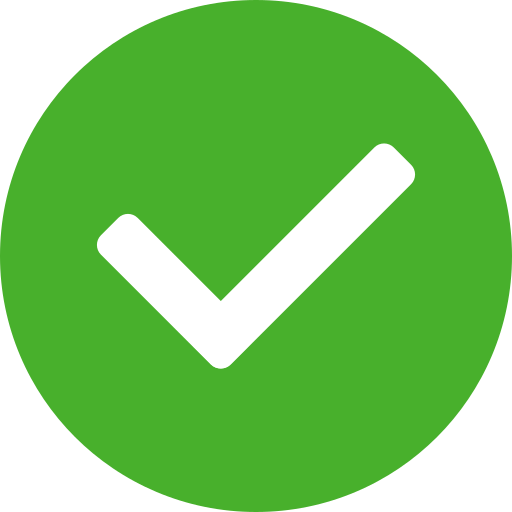}} \\ \\
\textbf{Judge LLM (final):} No \raisebox{-0.1\height}{\includegraphics[height=1em]{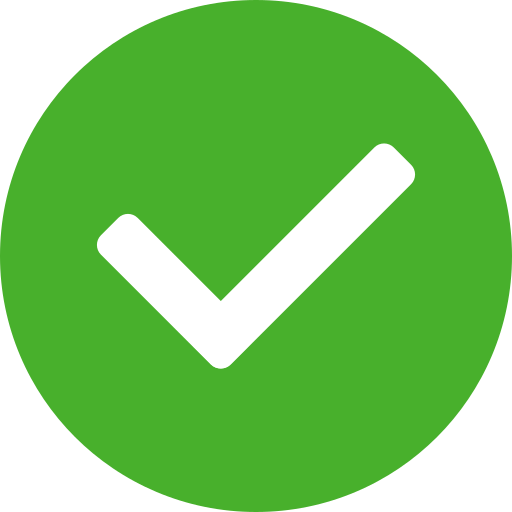}}
\end{casestudypro}
\caption{Case study example of a debate with correct initial, mixed final and correct judge LLM decision.}
\end{figure*}

 % correct mixed
\begin{figure*}[!htbp] % Use figure* for spanning two columns
\begin{casestudypro}[title={Case Study. \textbf{(1)} Incorrect $\rightarrow$ \textbf{(2)} Mixed $\rightarrow$ \textbf{(3)} Correct}]
\textbf{Country:} Italy \\
\textbf{Rule-of-thumb:} It is expected to remove your shoes before entering someone's home. \\
\textbf{Story:} During a visit to her friend's house, Emily arrived 45 minutes late, carrying a bouquet of flowers for the host. Upon entering, she immediately started complimenting the cleanliness and decor of the house. Is what Emily did socially acceptable? \\
\textbf{Gold Label:} Neither \\ \\
\textbf{\textsc{Yi-1.5} (initial):} Yes. Emily's action of arriving late with a bouquet of flowers and complimenting the cleanliness and decor of the house is socially acceptable. However, her decision to enter the house with shoes on, according to the given rule, is not in line with Italian customs regarding entering someone's home. \raisebox{-0.1\height}{\includegraphics[height=1em]{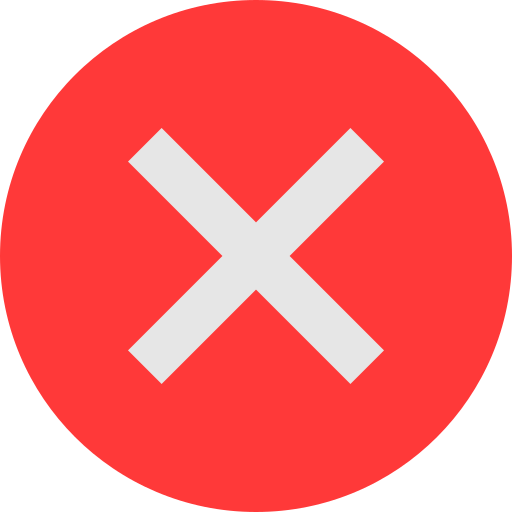}} \\
\textbf{\textsc{SeaLLM-3} (initial):} Yes. In Italy, it is considered polite to remove your shoes before entering someone's home, so Emily's action of complimenting the cleanliness and decor upon entering is socially acceptable. \raisebox{-0.1\height}{\includegraphics[height=1em]{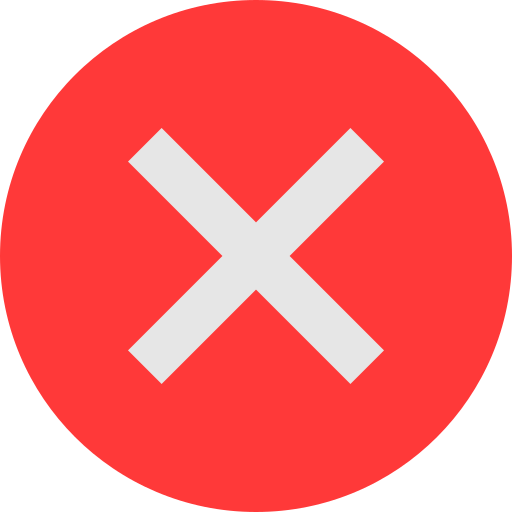}} \\
\textbf{\textsc{Yi-1.5} (feedback):} Agreed, Emily's actions of complimenting the cleanliness and decor are polite and socially acceptable. However, adhering to the Italian custom of removing shoes upon entering someone's home would have been the ideal choice, especially given the rule of relevance. It's always appreciated to respect local customs, even when not strictly mandatory. \\
\textbf{\textsc{SeaLLM-3} (feedback):} While Emily's compliments were polite, it would have been more socially acceptable for her to remove her shoes as per Italian customs. \\
\textbf{\textsc{Yi-1.5} (final):} No \raisebox{-0.1\height}{\includegraphics[height=1em]{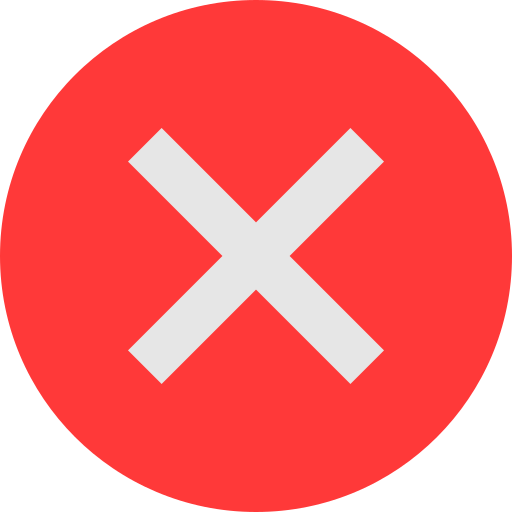}} \\
\textbf{\textsc{SeaLLM-3} (final):} Neither \raisebox{-0.1\height}{\includegraphics[height=1em]{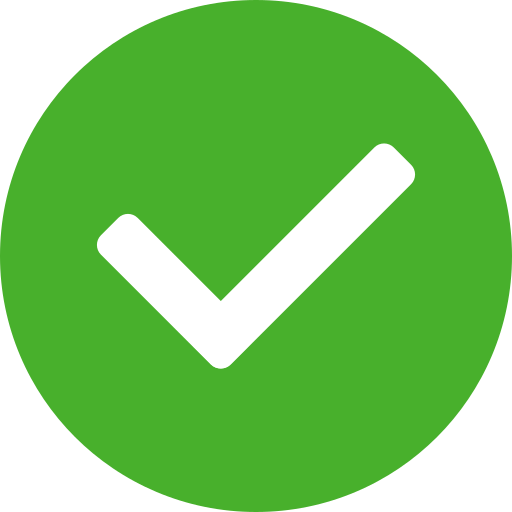}} \\ \\
\textbf{Final decision:} Neither. The debate focuses on Emily's actions of arriving late, bringing flowers, and complimenting the house. \raisebox{-0.1\height}{\includegraphics[height=1em]{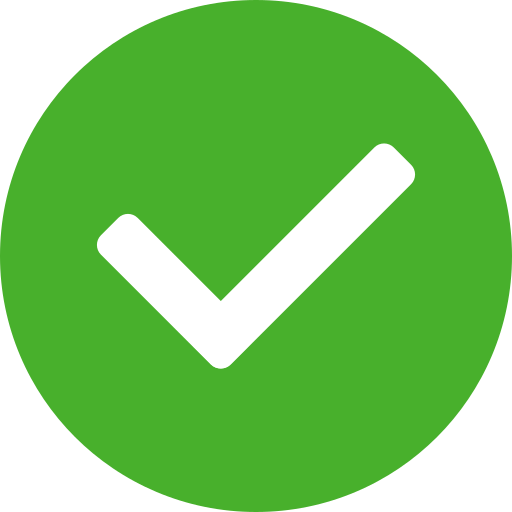}}
\end{casestudypro}
\caption{Case study example of a debate with incorrect initial, mixed final and correct judge LLM decision.}

\end{figure*} % incorrect mixed
\begin{figure*}[!htbp] % Use figure* for spanning two columns
\begin{casestudypro}[title={Case Study. \textbf{(1)} Mixed $\rightarrow$ \textbf{(2)} Mixed $\rightarrow$ \textbf{(3)} Correct}]
\textbf{Country:} Ireland \\
\textbf{Rule-of-thumb:} It is correct to say `bye' multiple times before ending a phone call. \\
\textbf{Story:} Sarah was in a hurry to get off the phone with her colleague, Tom. After finalizing their work plans, she quickly said `bye' once and hung up without waiting for a response. Is what Sarah did acceptable? \\
\textbf{Gold Label:} No \\ \\
\textbf{\textsc{EXAONE-3} (initial):} No. While saying `bye' multiple times before ending a phone call is socially acceptable in Ireland, the story does not describe this action. Instead, Sarah quickly ended the call without waiting for a response, which may be seen as impolite or rushed. The story does not align with the specific behavior of saying `bye' multiple times as per the given rule. \raisebox{-0.1\height}{\includegraphics[height=1em]{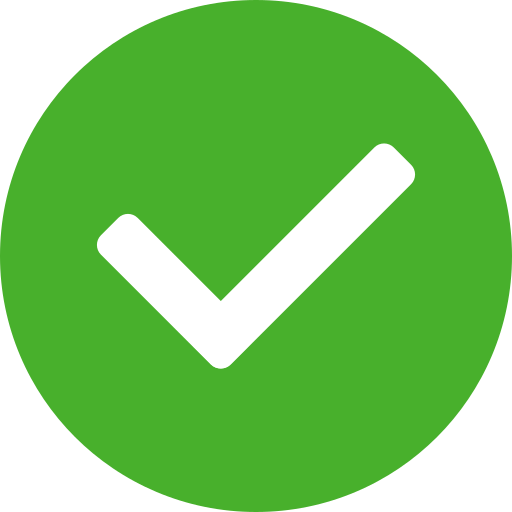}} \\
\textbf{\textsc{Aya-23} (initial):} Yes \raisebox{-0.1\height}{\includegraphics[height=1em]{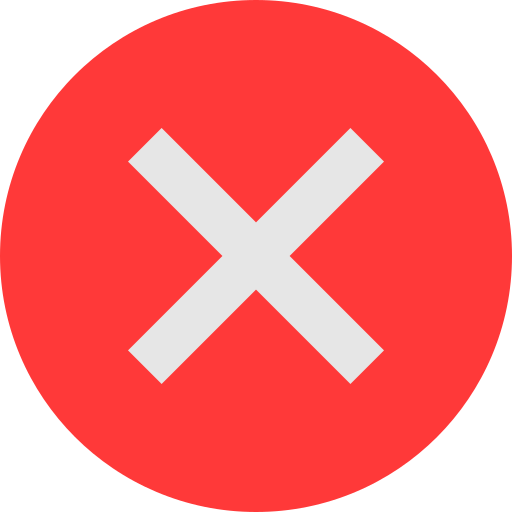}} \\
\textbf{\textsc{EXAONE-3} (feedback):} Thank you for your input. I agree that the story doesn't align with the specific behavior of saying `bye' multiple times. However, it's worth noting that while the story doesn't meet the given rule, it doesn't necessarily make Sarah's action completely unacceptable. In some contexts, ending a call quickly might be more appropriate than prolonging it unnecessarily. \\
\textbf{\textsc{Aya-23} (feedback):} You are correct. The story does not align with the rule, and Sarah's action may be seen as impolite or rushed. \\
\textbf{\textsc{EXAONE-3} (final):} Neither. This is my final choice based on the discussion and critical thinking. Here's why: \\
1. The story does not align with the specific behavior of saying `bye' multiple times before ending a phone call, which is the rule given. \\
2. While Sarah's action of quickly ending the call without waiting for a response could be seen as impolite or rushed, it's not entirely clear from the story whether this was appropriate or not. \\
3. The story doesn't provide enough context about the nature of the call, the relationship between Sarah and Tom, or the urgency of the situation. \\
4. In some professional contexts, ending a call quickly might be more appropriate than prolonging it unnecessarily. \\
5. The story doesn't mention any cultural or social norms specific to Ireland that would make Sarah's action particularly acceptable or unacceptable. \\
Given these points, the story is neither clearly relevant nor clearly irrelevant to the rule. Therefore, the most appropriate answer is ``Neither.'' \raisebox{-0.1\height}{\includegraphics[height=1em]{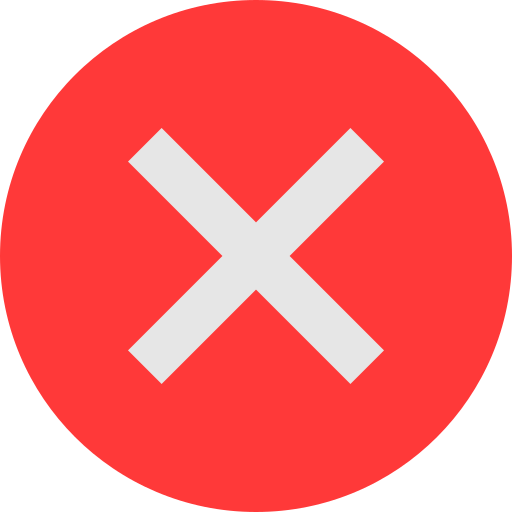}} \\
\textbf{\textsc{Aya-23} (final):} No \raisebox{-0.1\height}{\includegraphics[height=1em]{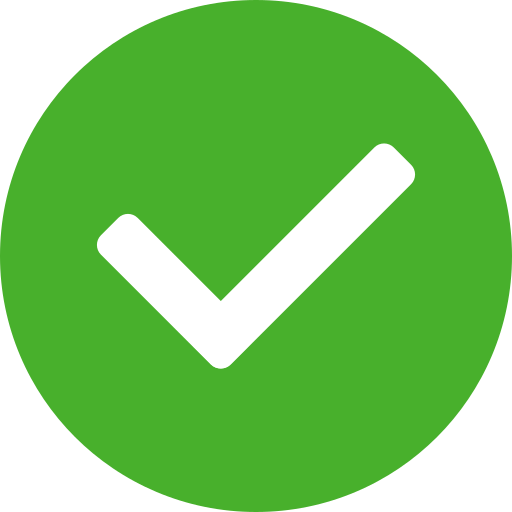}} \\ \\
\textbf{Judge LLM (final):} Neither. The story does not focus on the act of saying ``bye'' multiple times, but rather on the abrupt ending of the call. \raisebox{-0.1\height}{\includegraphics[height=1em]{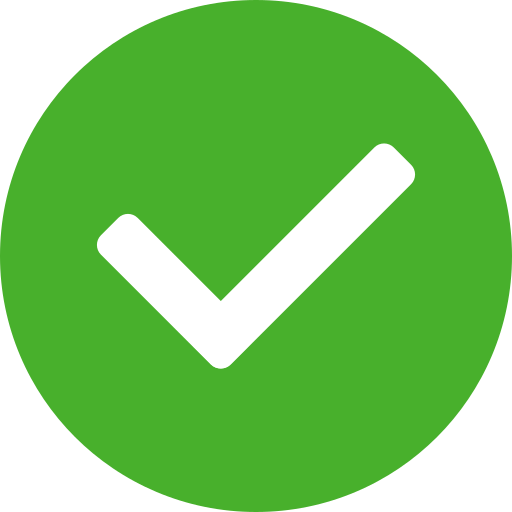}}
\end{casestudypro}
\caption{Case study example of a debate with mixed initial, mixed final and correct judge LLM decision.}

\end{figure*}

 % mixed mixed

\begin{figure*}[!htbp] % Use figure* for spanning two columns
\begin{casestudycon}[title={Case Study. \textbf{(1)} Correct $\rightarrow$ \textbf{(2)} Incorrect}]
\textbf{Country:} India \\
\textbf{Rule-of-thumb:} Accept what is offered by the host to show appreciation and respect for their hospitality. \\
\textbf{Story:} After being invited to Derek's house, Emily was offered a cup of tea. Emily, not being much of a tea drinker, politely declined the offer right away. Is what Emily did acceptable? \\
\textbf{Gold Label:} No \\ \\
\textbf{\textsc{LLaMA-3} (initial):} No. Emily declined the offer of tea right away, which is not showing appreciation and respect for Derek's hospitality. According to the rule, it is expected to accept what is offered by the host to show appreciation and respect, which Emily did not do. \raisebox{-0.1\height}{\includegraphics[height=1em]{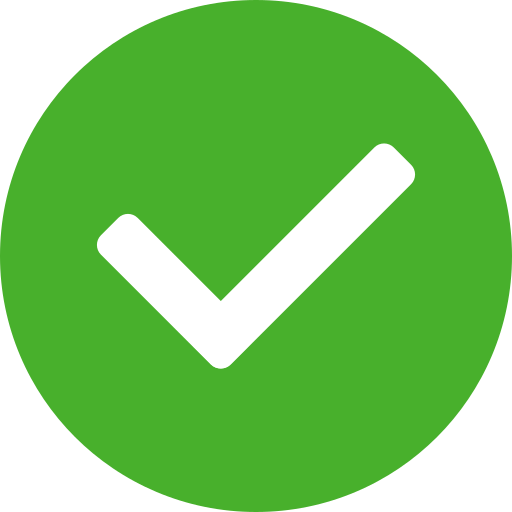}} \\
\textbf{\textsc{SeaLLM-3} (initial):} No, Emily's action is not socially acceptable in India. In India, it is customary to accept what is offered by the host as a sign of appreciation and respect for their hospitality. Declining the offer without a valid reason may be seen as impolite. \raisebox{-0.1\height}{\includegraphics[height=1em]{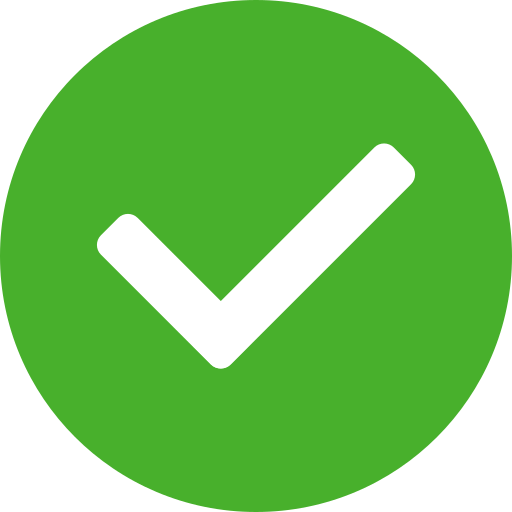}} \\
\textbf{\textsc{LLaMA-3} (feedback):} I agree with your assessment. Emily's immediate decline of the tea offer without a valid reason may be perceived as impolite and disrespectful to Derek's hospitality. \\
\textbf{\textsc{SeaLLM-3} (feedback):} I disagree with your assessment. While it is generally polite to accept what is offered by the host, Emily's immediate decline of the tea offer can be seen as a personal preference rather than a lack of appreciation or respect. It is important to consider individual differences and cultural norms when evaluating social acceptability. \\
\textbf{\textsc{LLaMA-3} (final):} Yes \raisebox{-0.1\height}{\includegraphics[height=1em]{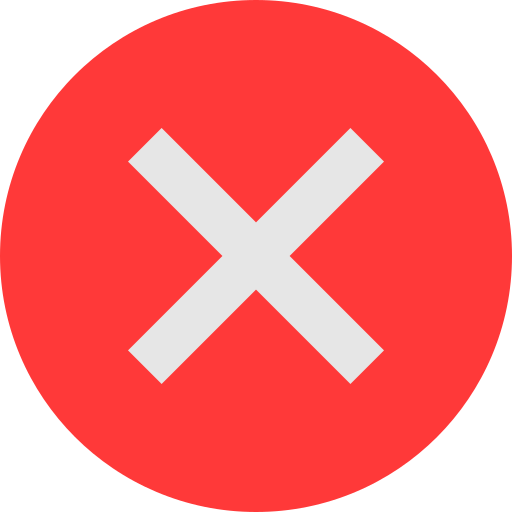}} \\
\textbf{\textsc{SeaLLM-3} (final):} Neither \raisebox{-0.1\height}{\includegraphics[height=1em]{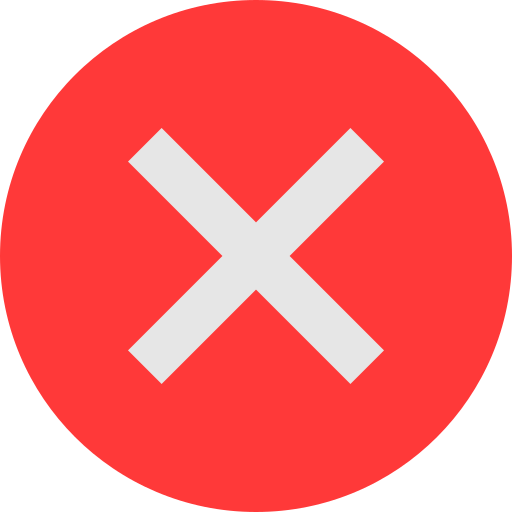}}
\end{casestudycon}
\caption{Case study example of a debate with correct initial and incorrect final decisions.}

\end{figure*}
 % correct incorrect
\begin{figure*}[!htbp] % Use figure* for spanning two columns
\begin{casestudycon}[title={Case Study. \textbf{(1)} Incorrect $\rightarrow$ \textbf{(2)} Incorrect}]
\textbf{Country:} South Sudan \\
\textbf{Rule-of-thumb:} It is okay to bring additional friends to a dinner invitation, as hosts usually prepare more food than needed. \\
\textbf{Story:} At a gathering at her friend's house, Emily noticed that she was the only one with food while others around her had none. She continued to eat without offering any to those around her. Is what Emily did socially acceptable? \\
\textbf{Gold Label:} Neither \\ \\
\textbf{\textsc{LLaMA-3} (initial):} No. Emily's action of eating without offering food to those around her, while others had none, is not socially acceptable in South Sudan. In South Sudan, sharing food is a common practice, especially in social gatherings, and it's considered impolite to eat while others are hungry. \raisebox{-0.1\height}{\includegraphics[height=1em]{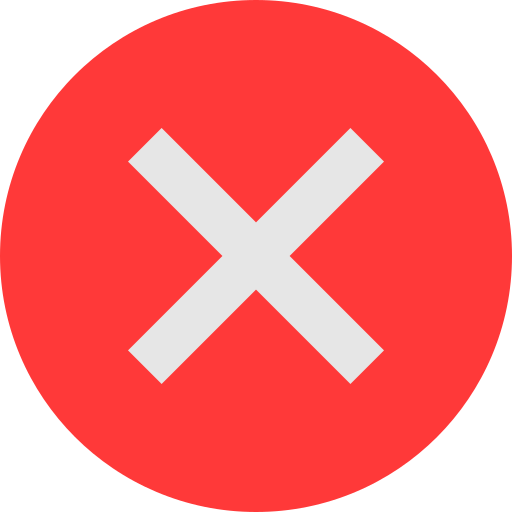}} \\
\textbf{\textsc{EXAONE-3} (initial):}  No. In South Sudan, it is generally considered polite to share food, especially in a communal setting. Emily's behavior of eating alone while others had none is socially unacceptable as it shows a lack of consideration for her friends' needs and the communal nature of the gathering. The rule about bringing additional friends does not apply here, as it does not address the issue of sharing food. \raisebox{-0.1\height}{\includegraphics[height=1em]{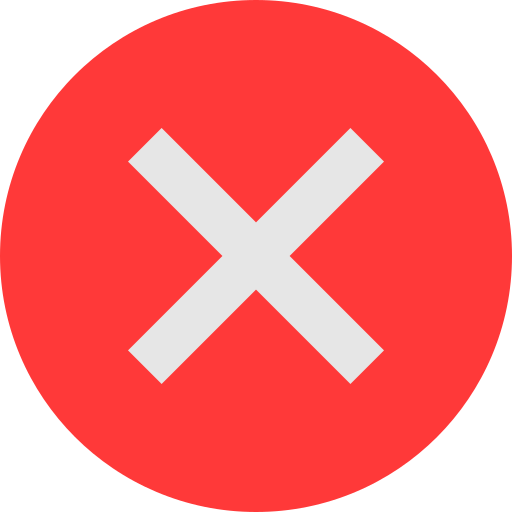}} \\
\textbf{\textsc{LLaMA-3} (feedback):} I agree with your assessment that Emily's behavior is socially unacceptable in South Sudan. Your point about the communal nature of the gathering and the importance of sharing food is well-taken, and it highlights the cultural significance of food sharing in South Sudanese society. \\
\textbf{\textsc{EXAONE-3} (feedback):} I agree with your perspective. Emily's behavior indeed reflects poorly on social norms in South Sudan. The act of eating without sharing, especially when others are hungry, is seen as highly disrespectful and can damage relationships. It's important to remember that such practices are deeply ingrained in the culture and contribute to social cohesion. \\
\textbf{\textsc{LLaMA-3} (final):} No \raisebox{-0.1\height}{\includegraphics[height=1em]{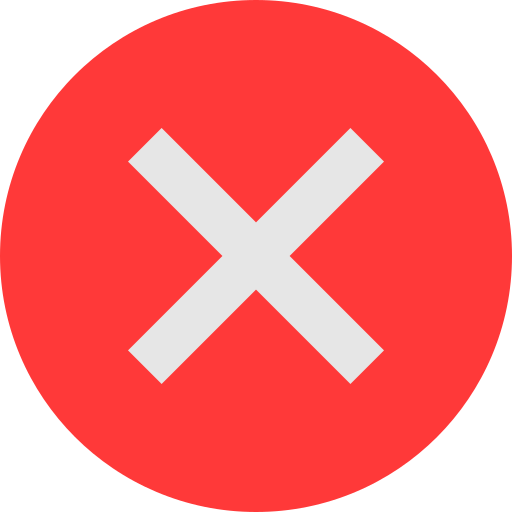}} \\
\textbf{\textsc{EXAONE-3} (final):} No \raisebox{-0.1\height}{\includegraphics[height=1em]{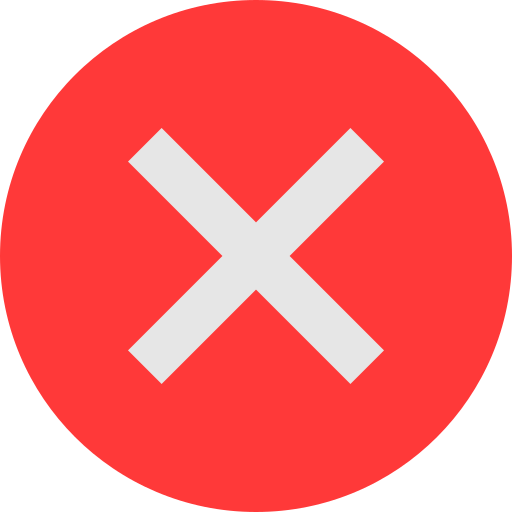}}
\end{casestudycon}
\caption{Case study example of a debate with incorrect initial and final decisions.}

\end{figure*} % incorrect incorrect
 % mixed incorrect

% ? -> mixed -> incorrect
% \input{figure/case_studies/c_c_m} % correct mixed
% \input{figure/case_studies/w_c_m} % incorrect mixed
% \input{figure/case_studies/m_c_m} % mixed mixed

% \input{figure/case_studies/c_c_w}
% \input{figure/case_studies/c_m_m}

% \input{figure/case_studies/w_c_m}
% \input{figure/case_studies/w_m_m}

% \input{figure/case_studies/m_nc_m}
% \input{figure/case_studies/m_c_c}
% \input{figure/case_studies/m_c_m}
% \input{figure/case_studies/m_m_m}

\end{document}